\documentclass[journal]{IEEEtran}

\IEEEoverridecommandlockouts
\def\bibfiles{../../} %

\usepackage{cite}             

\usepackage{amsmath}  
\usepackage{amsfonts,amssymb}
\usepackage{mleftright}       
\mleftright                   

\usepackage{graphicx}         
\usepackage{booktabs}         

\usepackage[lined,boxed,commentsnumbered,linesnumbered,ruled,vlined]{algorithm2e}

\usepackage[caption=false,font=footnotesize]{subfig}

\usepackage{pgfplots}
\usepackage{tikz}
\usetikzlibrary{shapes,patterns,decorations.text,decorations.pathreplacing,calc}
\pgfplotsset{compat=1.8}
\usetikzlibrary{external}
\usepackage{pdfpages}

\usepackage{nicefrac}  
\usepackage{comment}
\usepackage{balance}

\usepackage[capitalize]{cleveref} 
	\crefname{equation}{}{}
	\crefname{theorem}{Theorem}{Theorems}
	\crefname{lemma}{Lemma}{Lemmas}
	\crefname{cor}{Corollary}{Corollaries}
	\crefname{prop}{Proposition}{Propositions}
	\crefname{note}{Note}{Notes}
	\crefname{appsec}{Appendix}{Appendices}
	\crefname{definition}{Definition}{Definitions}
	\crefname{conj}{Conjecture}{Conjectures}
	\crefname{construction}{Construction}{Constructions}
	
\newcommand{\Reals}{\mathbb{R}}      

\newcommand{\Prv}[1]{\Pr\left[#1\right]}

\newcommand{\EE}[2][]{%
	\ifthenelse{\equal{#1}{}}%
	{\mathbb{E} \left[ #2 \right]}
	{\mathbb{E}_{#1} \left[ #2 \right]}
}

\newcommand{\Exp}{\mathbb{E}}

\newcommand{\eE}[2][]{\Exp_{#1}[#2]}

\newcommand{\hatVar}[1]{\widehat{\mathsf{Var}}\left[ #1 \right]}     
\newcommand{\Var}[1]{\mathsf{Var}\left[ #1 \right]}

\newcommand{\NN}[2]{\mathcal{N}\left({#1},{#2}\right)} 

\newcommand{\BC}[3]{\mathcal{BC}\left(#1,#2,#3\right)}

\newcommand{\bignorm}[2][]{\bigl\|#2\bigr\|_{#1}}

\newcommand{\set}[1]{\mathcal{#1}}
\newcommand{\thres}{1}
\newcommand{\numa}{M}

\newcommand{\dd}{\mathop{}\!\mathrm{d}}

\newcommand{\supp}{\operatorname{supp}}

\newcommand{\sigmaDPvar}{\sigma_{2,\mathrm{DP}}}
\newcommand{\sigmaDP}{\sigma_{\mathrm{DP}}}

\newtheorem{theorem}{Theorem}

\newtheorem{lemma}{Lemma}
\newtheorem{definition}{Definition}
\newtheorem{proposition}{Proposition}
\newtheorem{remark}{Remark}

\usepackage{bm}
\usepackage{dsfont}     

\definecolor{deepmagenta}{rgb}{0.80,0.0,0.80}
\definecolor{darkmagenta}{rgb}{0.55,0.0,0.55}
\definecolor{GreenB}{rgb}{0.27,0.67,0.42}
\definecolor{darkgreen}{rgb}{0, 0.5, 0}
\definecolor{darkyellow}{rgb}{0.61,0.53,0.05}
\definecolor{quinacridonemagenta}{rgb}{0.6016,0.0664,0.3086}
\definecolor{lightblue}{rgb}{0.61,0.87,1.0}
\definecolor{darkred}{rgb}{0.55, 0.0, 0.0}
\definecolor{deepcarmine}{rgb}{0.66, 0.13, 0.24}
\definecolor{flamingopink}{rgb}{0.99, 0.56, 0.67}

\usepackage{xr}%
\usepackage{subfiles}
\usepackage{calc}
\usepackage{bibunits}

\newcommand{\vardbtilde}[1]{\tilde{\raisebox{0pt}[0.85\height]{$\tilde{#1}$}}}

\newboolean{THESIS_VERSION}
\setboolean{THESIS_VERSION}{false}

\newboolean{QUICK_PLOTS} %
\setboolean{QUICK_PLOTS}{true} %

\newboolean{MAIN_REF} %
\setboolean{MAIN_REF}{false} %

\newlength{\mywidth}
\newlength{\myheight}
\newlength{\mywidthtwo}
\newlength{\myheighttwo}

\begin{document}

\title{Differentially-Private Collaborative Online Personalized Mean Estimation}

\author{Yauhen~Yakimenka,~\IEEEmembership{Member,~IEEE},
  Chung-Wei~Weng,~%
  Hsuan-Yin~Lin,~\IEEEmembership{Senior Member,~IEEE},
  Eirik~Rosnes,~\IEEEmembership{Senior Member,~IEEE},
  and~J{\"o}rg~Kliewer,~\IEEEmembership{Fellow,~IEEE}
  \thanks{%
  This paper was presented in part at the IEEE International Symposium on Information Theory  (ISIT), Taipei, Taiwan, June 2023 \cite{Yakimenka2023}.}
 \thanks{Y.~Yakimenka and J.~Kliewer are with Helen and John C. Hartmann Department of Electrical and Computer Engineering, New Jersey Institute of Technology, Newark, New Jersey 07102, USA (e-mail: yauhen.yakimenka@njit.edu, jkliewer@njit.edu).}%
 \thanks{C.-W.~Weng, H.-Y.~Lin, and E.~Rosnes are with Simula UiB, N-5006 Bergen, Norway (e-mail: chungwei@simula.no, lin@simula.no, eirikrosnes@simula.no).}}
 
\maketitle

\ifthenelse{\boolean{THESIS_VERSION}}{
\setlength{\mywidth}{0.32\columnwidth}
\setlength{\myheight}{0.32\columnwidth}
\setlength{\mywidthtwo}{0.46\columnwidth}
\setlength{\myheighttwo}{0.44\columnwidth}}{
\setlength{\mywidth}{0.68\columnwidth}
\setlength{\myheight}{0.66\columnwidth}
\setlength{\mywidthtwo}{0.68\columnwidth}
\setlength{\myheighttwo}{0.66\columnwidth}}

\maketitle

\begin{abstract}
  We consider the problem of collaborative personalized mean estimation under a privacy constraint in an environment  of several agents continuously receiving data according to arbitrary unknown agent-specific distributions. In particular, we provide a method based on hypothesis testing coupled with differential privacy and data variance estimation. Two privacy mechanisms and two data variance estimation schemes are proposed, and we provide a theoretical convergence analysis of the proposed algorithm for any bounded unknown distributions on the agents'  data, showing that collaboration provides faster convergence than a fully local approach where agents do not share data. Moreover, we provide analytical performance curves for the case with an oracle class estimator, i.e., the class structure of the agents, where agents receiving data from distributions with the same mean are considered to be in the same class, is known.  The theoretical \emph{faster-than-local} convergence guarantee is backed up by extensive numerical results showing that for a considered scenario the proposed approach indeed converges  much faster than a fully local approach, and performs comparably to ideal performance  where all data is public.   This illustrates the benefit of private collaboration  in an online setting. %
\end{abstract}

\section{Introduction}

Collaborative learning has attracted significant attention lately through popular frameworks such as federated learning (FL) \cite{McMahanMooreRamageHampsonArcas17_1,Konecy-etal16_1,LiSahuTalwalkarSmith20_1} (partially  decentralized) and fully decentralized approaches like swarm learning \cite{Warnat-Herresthal2021}. However, different agents in the learning environment may have different objectives and hence the individually collected data may be heterogeneous and specific for each personalized learning task.  Despite this,  collaboration can significantly accelerate learning among a set of agents sharing a limited set of common objectives. %
 A crucial part of any collaborative algorithm for personalized learning is the identification of agents with data from similar distributions, in particular in an online setting in  which data becomes available continuously over time.

Personalized approaches for distributed learning \cite{Smith2017,Vanhaesebrouck2017,Hanzely2020} have attracted significant interest recently. In \cite{Hanzely2020},  Hanzely and Richt{\'a}rik introduced the concept of personalized FL. In contrast to conventional FL, personalized FL  looks for a trade-off between a global model and local models learned by each agent from its own dataset, as formulated in terms of a correction term to the traditional empirical risk minimization objective  function. As shown in \cite{Hanzely2020}, personalization in general yields reduced communication complexity.

For the online setting, previous work on collaborative learning has mostly focused on the multi-armed bandit (MAB) model, mostly considering a \emph{single} MAB instance (the arm means do not vary across the agents), see, e.g., \cite{Madhushani2021,Tao2019,Hillel2013} and references therein, while some recent works also consider the case where the arm means vary across agents \cite{Karpov2022} and with some amount of personalization by  optimizing a mixture between a global and local objectives \cite{Shi2021}.

In this paper, we consider the problem of collaborative online personalized mean estimation, first introduced in \cite{AsadiBelletMaillardTommasi22_1}, in which each agent continuously receives data according to an unknown agent-specific distribution. The aim of each agent is to calculate an accurate  estimate of the mean of its underlying  distribution as quickly as possible. As in \cite{AsadiBelletMaillardTommasi22_1}, we assume an unknown underlying class structure %
where agents in the same class receive data from distributions with the same mean. %
As noted in \cite{galante2024scalable}, the approach in \cite{AsadiBelletMaillardTommasi22_1} struggles when there is a larger number of agents as the per-agent space and time complexities are both linear in the number of agents.  To address this issue a method that allows each agent to communicate with only a selected number of agents  was  proposed in \cite{galante2024scalable}, combined with both a message-passing scheme and a  consensus-based approach on top for collaborative online mean estimation.

A major limitation of the algorithms proposed in \cite{AsadiBelletMaillardTommasi22_1,galante2024scalable} is that data is directly shared with other agents in the learning environment  without any protection, which  is in contrast to FL where there is no sharing of data amongst the agents. This may  
leak sensitive user information to other agents in the environment. In order to provide some level of user data privacy, we propose to add random noise to the data before it is released to other agents in the environment  according to the principle of differential privacy (DP) \cite{Dwork06_1,DworkMcSherryNissimSmith06_1}. In \cite{Saha2023}, a \emph{local} DP approach is proposed for collaborative mean estimation, but for a centralized setting in which there is a central server coordinating the estimation and for \emph{offline} data, i.e., the data is not continuously received by the agents. On the other hand, the effect of packet losses in agent-to-agent communication is considered.

Our main contributions are as follows.

\begin{itemize}
\item Two  (online) privacy mechanisms inspired by those in \cite{ChanShiSong11_1,Dwork2010} are proposed, using either  Gaussian or  Laplace noise (only Gaussian noise was considered in \cite{Yakimenka2023}). Moreover, as opposed to the initial work in \cite{AsadiBelletMaillardTommasi22_1}, we consider an approach based on  hypothesis testing.

 \item Second, as the variances of the agents' data distributions are typically unknown, they need to be estimated. Hence, we propose two variance estimation schemes %
 and evaluate their effect on the overall performance. Note that in \cite{Yakimenka2023}, the data variances were assumed known, which simplified the derivations.

 \item   %
A theoretical convergence analysis of our proposed method  for any bounded distributions on the data is provided, showing that the proposed method converges \emph{faster} than a local approach for the case of known data variance (see Theorem~\ref{thm:1}). Note that in \cite[Thm.~1]{Yakimenka2023} a much weaker result only showing convergence was provided,  which simplified the proof. %

 \item Finally, numerical results showing that our proposed approach indeed converges  \emph{faster} than a fully local setting where agents do not share data are provided (as predicted by Theorem~\ref{thm:1}). We also compare the simulation results with analytical curves (see Proposition~\ref{prop:2}) for the case with known data variance and with an oracle class estimator, i.e., the class structure of the agents, where agents receiving data from distributions with the same mean are considered to be in the same class, is known. The results show that the best scheme performs comparably to ideal performance  where all data is public (see Proposition~\ref{prop:3}). The effect of estimating the data variance increases the mean squared mean estimation error initially, but in the end it converges to  the mean squared  error  for the case when the data variances are assumed known from the beginning. This illustrates the benefit of collaboration  in an online setting while preserving users' data privacy.

\end{itemize}

Although we base our schemes on the original algorithm from \cite{AsadiBelletMaillardTommasi22_1}, similar approaches can be used for the improved scheme in \cite{galante2024scalable} in order to protect user data under collaboration.

\section{Preliminaries}

\subsection{Notation}
\label{sec:notation}
In general, but with some exceptions, we use uppercase and lowercase letters for random variables (RVs) 
and their realization, %
respectively, and italics for sets, e.g., \(X\), \(x\),  and \(\mathcal{X}\)  represent a RV, its realization, and a set, respectively.  %
The expectation of a RV $X$ is denoted by $\EE{X}$, while its variance is denoted by $\Var{X}$. We define $[n] \triangleq \{1,2,\dotsc, n\}$ and $[i:j] \triangleq \{i,i+1,\dotsc,j\}$, %
while %
$\Reals$ denotes the real numbers. %
$\set{N}(\mu,\sigma^2)$ denotes the Gaussian distribution with mean $\mu$ and variance $\sigma^2$, while $\set{L}(\mu,b)$ denotes the Laplace distribution with mean $\mu$ and scale parameter $b$ (variance is $2b^2$).  $X \sim \set P$ denotes that $X$ is distributed according to the distribution $\set P$, and the probability distribution of an arbitrary RV $X$ with mean $\mu$ and variance $\sigma^2$ is denoted by $\set{P}_X(\mu,\sigma^2)$.  %
Standard order notations $O(\cdot)$ and $o(\cdot)$ are used for asymptotic results, %
while $\Phi(\cdot)$ is the cumulative distribution function of the standard Gaussian distribution. 
$w_{\mathrm{H}}(n)$ denotes the Hamming weight of the binary representation of the nonnegative integer $n$.

\ifthenelse{\boolean{THESIS_VERSION}}{
\begin{figure}[t!]
\begin{center}
\includegraphics[width=0.35\columnwidth]{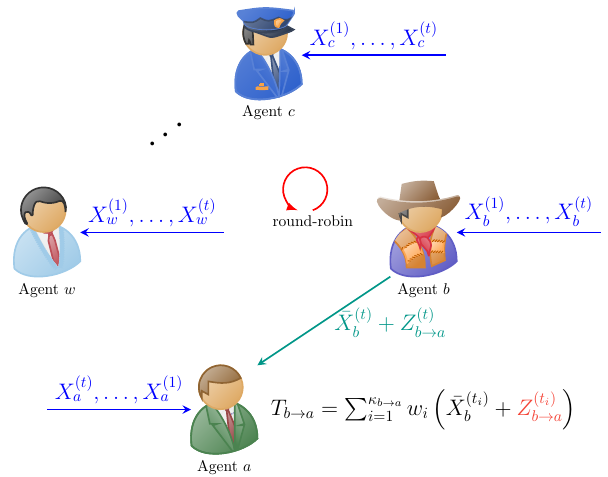}
\caption{System model.} \label{fig:system_model}
\end{center}
\vspace{-4ex}
\end{figure}}
{\begin{figure}[t!]
\begin{center}
\includegraphics[width=0.6\columnwidth]{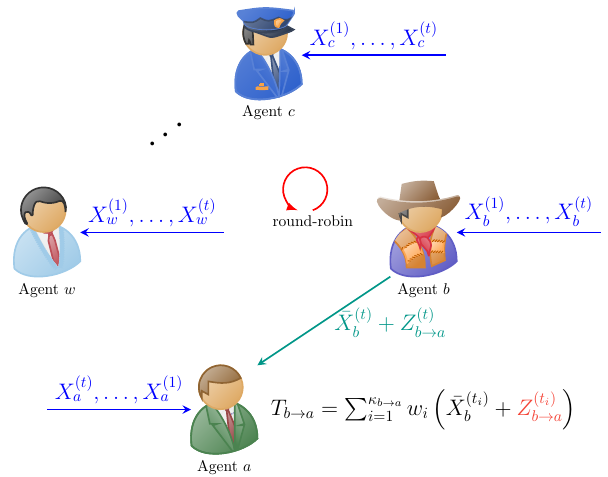}
\caption{System model.} \label{fig:system_model}
\end{center}
\vspace{-4ex}
\end{figure}}

\subsection{System Model (Problem Formulation)}
There are $\numa$ independent agents. Each agent $a \in [\numa]$ wants to estimate the mean of its sample $X_a^{(1)}, X_a^{(2)}, \dotsc \in \mathcal{X}_a \subset \Reals$, where the sample follows an arbitrary unknown distribution $\set D_a$ over the bounded set $\mathcal{X}_a$ with an (unknown) mean $\mu_a$ and known/unknown standard deviation $\sigma_a < \infty$, where $X_a^{(t)}$ arrives to the agent $a$ at time $t$. We assume the time is discrete and synchronized between the agents. For simplicity, we  assume that $\mathcal{X}_a$ is the same for all agents $a$. The agents have limited memory and thus decide to keep only the current mean of the sample: $\bar X_a^{(t)} = \frac{1}{t} \sum_{i=1}^{t} X_a^{(i)}$. 

It is known that some agents $a \neq b$ might have samples from the same distribution $\set D_a = \set D_b$, and they want to exploit this. However, there is no preliminary information on the agents' distributions and also, the agents would like to keep their particular sample values private.

The collaborative algorithm consists of the agents exchanging their current sample means. In order to preserve privacy, a DP mechanism \cite{Dwork06_1,DworkMcSherryNissimSmith06_1} is applied before releasing the current sample mean  to other agents. More precisely, at each time step $t$, the agent $a$ receives $X_a^{(t)}$, updates its sample mean $\bar X_a^{(t)}$, and also chooses another agent $b \in [\numa] \setminus \{a\}$ to query. The agent $b$ then sends its current sample mean to $a$, but privatized: $\bar X_b^{(t)} + Z_{b \to a}^{(t)}$, where $Z_{b \to a}^{(t)}$ is a Gaussian or Laplace RV (``noise'') with zero mean and variance depending on the DP mechanism employed (see \cref{sec:privacy}). While $Z_{b \to a}^{(t)}$ is independent from the sample, it in general depends on the noise generated by agent $b$ at different times. 
We call a particular construction of the noise $Z_{b \to a}^{(t)}$ a \emph{private release mechanism}. The system model is illustrated in \cref{fig:system_model}. Moreover, the agent $a$  (at time $t$) potentially receives a noise-corrupted estimate $V_b^{(t)} + U_{b \to a}^{(t)}$ of the variance $\sigma_b^2$ of the data distribution of agent $b$, where $V_b^{(t)}$ denotes a variance estimate and $U_{b \to a}^{(t)}$ the corresponding Gaussian (or Laplace) noise. Alternatively, agent $a$ can estimate $\sigma_b^2$ based on its previously received noise-corrupted sample means  $\bar X_b^{(\tilde{t})} + Z_{b \to a}^{(\tilde{t})}$ from agent $b$ for times $\tilde{t}$ up to and including $t$.

Based on the content of its memory at time $t$, agent $a$ calculates its current estimate of $\mu_a$, which we denote by $\mu_a^{(t)}$. The goal is to construct a collaborative protocol that allows for faster convergence of $\mu_a^{(t)}$ to $\mu_a$. In this work, we measure the speed of convergence in terms of the average expected squared deviation from the mean, i.e., by
	$\nicefrac{\sum_{a \in [\numa]} \eE{(\mu_a^{(t)} - \mu_a)^2}}{\numa} $
as a function of $t$. 

Also, the agents want regular updates so that at every moment $t$ they have a good estimate of $\mu_a$. Hence, we do not consider the approach where one waits until $t \approx t_{\max}$ and then query every agent's last sample mean.

\subsection{Differential Privacy}

We start by defining the concept of DP. Then, we provide some  key lemmas based on the Gaussian and Laplace mechanisms.

\begin{definition}
  \label{def:DP}
  A randomized function $F\colon\set{X}^n\to\set{Y}$ is $(\epsilon,\delta)$-differentially private if for all subsets $\set{S}\subseteq\set{Y}$ and for all $(x_1,\dotsc,x_n)\in\set{X}^n$ and $(x'_1,\dotsc,x'_n)\in\set{X}^n$ which differ in a single component, i.e., $x_i \neq x_i'$ for exactly one $i \in [n]$,
  \begin{IEEEeqnarray*}{c}
    \Prv{F(x_1,\dotsc,x_n) \in \set{S}} \leq \mathrm{e}^\epsilon \Prv{F(x'_1,\dotsc,x'_n) \in \set{S}} + \delta.
  \end{IEEEeqnarray*}
\end{definition}

\begin{lemma}
  \label{lem:DP}
  Let $(x_1,\dotsc,x_n) \in \mathcal{X}^n$ where $\mathcal{X} = [\mu-L,\mu+L]$ for some finite values $\mu$ and $L$.  %
  Then, the noise-corrupted sample mean 
    $\nicefrac{(x_1+\cdots+x_n)}{n} + \nicefrac Zn$,
  where $Z \sim \mathcal{N}\left(0,\sigmaDP^2 \right)$ and $\sigmaDP^2 \triangleq \nicefrac{8L^2 \ln(1.25/\delta)}{\epsilon^2}$ is 
  $(\epsilon,\delta)$-differentially private for $0 < \epsilon \leq 1$ and $0 < \delta  \leq 1$.
\end{lemma}

\begin{IEEEproof}
See Appendix~\ref{app:proof_lem:DP} in the supplementary material.
\end{IEEEproof}

\begin{lemma}
  \label{lem:DP_Laplace}
  Let $(x_1,\dotsc,x_n) \in \mathcal{X}^n$ where $\mathcal{X} = [\mu-L,\mu+L]$ for some finite values $\mu$ and $L$.  %
  Then, the noise-corrupted sample mean 
    $\nicefrac{(x_1+\cdots+x_n)}{n} + \nicefrac Zn$,
  where $Z \sim \mathcal{L}\left(0,\nicefrac{\sigmaDP}{\sqrt{2}} \right)$ and $\sigmaDP^2 \triangleq \nicefrac{8L^2}{\epsilon^2}$ is 
  $(\epsilon,0)$-differentially private for $\epsilon > 0$.
\end{lemma}

\begin{IEEEproof}
See Appendix~\ref{app:proof_lem:DP_Laplace} in the supplementary material.
\end{IEEEproof}

\begin{lemma}
  \label{lem:DP_squared}
  Let $(x_1,\dotsc,x_n) \in \mathcal{X}^n$ where $\mathcal{X} = [\mu-L,\mu+L]$ for some finite values $\mu$ and $L$.  %
  Then, the noise-corrupted sample variance  
    $\nicefrac{(x_1^2+\cdots+x_n^2)}{(n-1)} -  \nicefrac{(x_1+\cdots+x_n)^2}{n(n-1)}+ \nicefrac{W}{n}$,
  where $W \sim \mathcal{N}\left(0,\sigmaDPvar^2 \right)$ and $\sigmaDPvar^2 \triangleq \nicefrac{32L^4 \ln(1.25/\delta)}{\epsilon^2}$ is 
  $(\epsilon,\delta)$-differentially private for $0 < \epsilon \leq 1$ and $0 < \delta  \leq 1$.
\end{lemma}

\begin{IEEEproof}
See Appendix~\ref{app:proof_lem:DP_squared} in the supplementary material.
\end{IEEEproof}

Note that because of squaring, the standard deviation of the DP noise increases significantly compared to that of Lemma~\ref{lem:DP}. In Lemma~\ref{lem:DP_squared}, it is proportional to $L^4$, while in  Lemma~\ref{lem:DP} it is only proportional to $L^2$.

An analogous result to Lemma~\ref{lem:DP_squared} can be shown for Laplace noise with variance $\sigmaDPvar^2 \triangleq \nicefrac{32L^4}{\epsilon^2}$. Details are omitted for brevity.

\subsection{Mathematical Tools} \label{sec:tools}
\begin{lemma} \label{lem:weighting}
	Assume independent RVs $X_1, X_2, \dotsc, X_n$, $n \ge 2$, have 
	$\Var{X_i} = \sigma_i^2$. If $X=\alpha_1 X_1 + \alpha_2 X_2 + \dotsb + \alpha_n X_n$, where $\sum_{i=1}^n \alpha_i = 1$, then $X$ has the minimum variance when the weights $\alpha_i$ are selected as %
		$\alpha_i^* = \frac{1}{\sigma_i^2 \sum_{j=1}^n \frac{1}{\sigma_j^2}}$,
 and
	\[
		\min_{\alpha_1,\ldots,\alpha_n} \Var{X} = \sum_{i=1}^n \left( \alpha_i^* \right)^2 \sigma_i^2 =\frac{1}{\sum _{i=1}^n \frac{1}{\sigma_i^2}} < \min_{i\in [n]} \sigma_i^2.
	\]
\end{lemma}

Lemma~\ref{lem:weighting} provides an intuition for the whole approach: having access to RVs with the same mean and even very high variances still allows to decrease the total variance of the estimator $X$, provided the weights $\alpha_i$ are properly chosen.

Assume we have two independent Gaussian RVs $X \sim \NN{\mu}{\sigma^2}$ and $Y \sim \NN{\nu}{\tau^2}$, where the variances $\sigma^2$ and $\tau^2$ are known, but the means $\mu$ and $\nu$ are not. We construct a simple hypothesis test for checking if $\mu = \nu$ as follows,
\begin{align*}
	\mathcal H_0:\; \mu = \nu \text{ and } 
	\mathcal H_1:\; \mu \neq \nu.
\end{align*}
First, let 
	$Z = \nicefrac{(X - Y)}{\sqrt{\sigma^2 + \tau^2}} \sim \NN{\nicefrac{(\mu - \nu)}{\sqrt{\sigma^2+\tau^2}}}{1}$.
Then, if we have a predefined confidence level $\theta \in [0,1]$ and $z \triangleq \Phi^{-1}\left( 1 - \nicefrac{\theta}{2}\right)$, we 
\begin{align*}
	&\text{accept $\mathcal H_0$ if $|Z| < z$ and 
	reject  $\mathcal H_0$  otherwise}.
\end{align*}
We call the situation when $\mu = \nu$ but $\mathcal H_0$ is rejected, a \emph{type-I error}. The probability of type-I error is $\theta$. And if $\mu \neq \nu$ but $\mathcal H_0$ is accepted, we call this a \emph{type-II error}.  %

On the other hand, if the  variances $\sigma^2$ and $\tau^2$ are unknown, they need to be estimated and we need to consider sequences of RVs. In particular, assume we have two independent sequences with $n$ independent and identically distributed (i.i.d.) Gaussian RVs $X_i \sim \NN{\mu}{\sigma^2}$ and $Y_i \sim \NN{\nu}{\tau^2}$, $i \in [n]$.  First construct
\[
Z = \frac{\sum_{i=1}^n X_i - \sum_{i=1}^n Y_i}{\sqrt{\frac{S_{\mathrm{x}}^2}{n} + \frac{S_{\mathrm{y}}^2}{n}}},
\]
where $S_{\mathrm{x}}^2$ and $S_{\mathrm{y}}^2$ are the sample variances of $X_1,\ldots,X_n$ and $Y_1,\ldots,Y_n$, respectively, i.e., 
\[
S_{\mathrm{x}}^2 \triangleq  \frac{1}{n-1} \left( \sum_{i=1}^n \left( X_i -  \frac{\sum_{j=1}^n X_j}{n} \right) \right)^2
\]
and similarly for $S_{\mathrm{y}}^2$.
The Welch $t$-test for checking if $\mu = \nu$ is as follows,
\begin{align*}
	&\text{accept $\mathcal H_0$ if $|Z| < z$ and 
	reject  $\mathcal H_0$  otherwise},
\end{align*}
where $z \triangleq \Phi_{\text{t},\nu}^{-1}\left( 1 - \nicefrac{\theta}{2}\right)$, 
\begin{align*} %
\nu = \frac{\left(\frac{S_{\mathrm{x}}^2}{n} + \frac{S_{\mathrm{y}}^2}{n}\right)^2}{\left(\frac{S_{\mathrm{x}}^2}{n}\right)^2 \cdot \frac{1}{n-1}+ \left(\frac{S_{\mathrm{y}}^2}{n}\right)^2 \cdot \frac{1}{n-1}},
\end{align*}
and $\Phi_{\text{t},\nu}$ is the cumulative distribution function of the $t$-distribution with $\nu$ degrees of freedom.

\subsection{Bernstein's Condition}
 Further in the paper, we prove the main convergence result for a wide class of distributions satisfying \emph{Bernstein's condition}.

	\begin{definition}[{\hspace{-0.01cm}\cite[Eq.~(2.15)]{Wainwright19_1}}] \label{def:bernstein}
		We say that a RV $X \in \Reals$ with mean $\mu$ and variance $\sigma^2$ satisfies \emph{Bernstein's condition} with parameter $\beta > 0$, if
		\[
		\left | \EE{(X- \mu)^k} \right| \leq \frac{1}{2} k! \sigma^2 \beta^{k-2} \quad \text{for $k=2,3,\ldots$}.
		\]
	\end{definition}

With some abuse of notation, we write $X \sim \BC{\mu}{\sigma^2}{\beta}$. Note that if $X \sim \BC{\mu}{\sigma^2}{\beta}$, then also $X \sim \BC{\mu}{\sigma^2}{\beta'}$ for any $\beta' \ge \beta$ (monotonicity of the Bernstein parameter).
Examples of  RVs satisfying Bernstein's condition are Gaussian and Laplace RVs, as well as RVs with bounded support.

\section{Our Approach}
There are three ingredients of our approach that should be addressed:
	\begin{itemize}
		\item private release mechanism $Z_{b \to a}^{(t)}$ that adds the minimum amount of noise sufficient for $(\epsilon,\delta)$-DP of the sample by the agent $b$ (see \cref{sec:privacy}),
		\item identification by the agent $a$ of the agents with the same distribution mean (\emph{decision rule}, see \cref{sec:decision_rule}), and
		\item using the information obtained from these agents in order to improve the local mean estimate (statistic $T_{b \to a}$, see \cref{sec:statistic}).
	\end{itemize}

Formally, we define the class of agents having the same distribution mean as $a$ by $\set C_a \triangleq \{ b \in [\numa] : \mu_b = \mu_a \}$. The classes are not known, which makes the problem nontrivial. We denote also by $\set C_a^{(t)} \triangleq \{ b \in [\numa] : \chi_a^{(t)}(b; \theta_t) = 1 \}$ the estimate of the class $\set C_a$ by the agent $a$ at time $t$, where $\chi_a^{(t)}(b; \theta_t)$ is some decision rule at time $t$,  %
i.e., $\chi_a^{(t)}(b; \theta_t) = 1$ if at time $t$ agent $a$ believes that agent $b$ is in $\set C_a$. Here, $\theta_t$ is a prescribed  confidence level that  depends on $t$. %

\subsection{Linear Statistic $T_{b \to a}$} \label{sec:statistic}

Let  $\sum_{i=1}^{\kappa_{b \to a}} w_i = 1$, denote by $t_1,t_2,\dotsc$ the times at which agent $b$ is queried by agent $a$, and let 
\begin{align} \label{eq:Tab}
T_{b \to a} = \sum_{i=1}^{\kappa_{b \to a}} w_i \left(\bar{X}_b^{(t_i)} + Z_{b \to a}^{(t_i)} \right)
\end{align}
be the current statistic of the received noise-corrupted sample means by agent $a$ from agent $b$ after the $\kappa_{b \to a}$-th query, $\kappa_{b \to a} = 1,2,\dotsc$. Then, when the variance of agent $b$'s data distribution $\sigma_b^2$ is known (with  $t_0 \triangleq  0$),
\ifthenelse{\boolean{THESIS_VERSION}}{
\begin{IEEEeqnarray}{rCl}
  \Var{T_{b \to a}}& = &\sigma_b^2 \sum_{i=1}^{\kappa_{b \to a}} (t_i-t_{i-1}) \biggl( \sum_{j=i}^{\kappa_{b \to a}} \frac{w_j}{t_j}\biggr)^2 %
  + \Var{\sum_{i=1}^{\kappa_{b \to a}} w_i Z_{b \to a}^{(t_i)}}.\IEEEeqnarraynumspace\label{eq:var_Tab}
\end{IEEEeqnarray}}
{
\begin{IEEEeqnarray}{rCl}
  \Var{T_{b \to a}}& = &\sigma_b^2 \sum_{i=1}^{\kappa_{b \to a}} (t_i-t_{i-1}) \biggl( \sum_{j=i}^{\kappa_{b \to a}} \frac{w_j}{t_j}\biggr)^2 \nonumber\\
  && +\> \Var{\sum_{i=1}^{\kappa_{b \to a}} w_i Z_{b \to a}^{(t_i)}}.\IEEEeqnarraynumspace\label{eq:var_Tab}
\end{IEEEeqnarray}}

The weights $\{ w_i \}$ depend on $\kappa_{b \to a}$, but we omit this dependency for simpler notation. Picking $w_1=\cdots=w_{\kappa_{b \to a}-1}=0$ and $w_{\kappa_{b \to a}}=1$ corresponds to keeping the last update as in \cite{AsadiBelletMaillardTommasi22_1}, while picking $w_1=\cdots=w_{\kappa_{b \to a}}=\nicefrac{1}{\kappa_{b \to a}}$ corresponds to what we refer to as the mean-of-mean (MoM) statistic. For simplicity, we refer to the former approach as non-MoM.

\subsection{Statistic $V_{b \to a}$} \label{sec:statistic_variance}

Let $V_{b \to a}$ denote the current noise-corrupted estimate of the variance $\sigma_b^2$ by agent $a$. $V_{b \to a}$ can either be an estimate computed based on the received noise-corrupted shared sample means up to the current time (see \cref{sec:privacy_sample_variance_2}), or a directly shared  noise-corrupted unbiased sample variance estimate, i.e., $V_{b \to a} = V_b^{(t_{\kappa_{b \to a}})} + U_{b \to a}^{(t_{\kappa_{b \to a}})}$, where $V_b^{(t_{\kappa_{b \to a}})}$ denotes
the sample variance for the $\kappa_{b \to a}$-th query, while the corresponding noise (Gaussian or Laplace) is denoted by $U_{b \to a}^{(t_{\kappa_{b \to a}})}$. The statistics of the noise $U_{b \to a}^{(t_{\kappa_{b \to a}})}$ (e.g., its variance) depend on the specific privacy mechanism used, and we refer the reader to \cref{sec:privacy_sample_variance_1} for further details.

\subsection{Decision Rule} \label{sec:decision_rule}
The sum $\sum_{i=1}^{\kappa_{b \to a}} w_i \bar{X}_b^{(t_i)}$ is a weighted average of the i.i.d. RVs $X_b^{(1)}, X_b^{(2)}, \dotsc, X_b^{(t_{\kappa_{b \to a}})}$. If this sum is Gaussian distributed, it follows from \eqref{eq:Tab} that $T_{b \to a}$ is also Gaussian (when using a Gaussian privacy mechanism). We now consider separately the cases of known and unknown data variance.

\subsubsection{Known Data Variance}
As $T_{b \to a}$ is Gaussian, we pick a decision rule based on the hypothesis test outlined above in \cref{sec:tools} (known variance case), i.e., we let $\chi_a^{(t)}(b; \theta_t)=1$, for $b \neq a$, if
\ifthenelse{\boolean{THESIS_VERSION}}{
\begin{align*}
\left| \bar{X}_a^{(t)} - T_{b \to a}  \right| &< \Phi^{-1}\left( 1 - \frac{\theta_t}{2}\right) \sqrt{\Var{\bar{X}_a^{(t)}} + \Var{T_{b \to a}}} %
= \Phi^{-1}\left( 1 - \frac{\theta_t}{2}\right) \sqrt{\frac{\sigma_a^2}{t} + \Var{T_{b \to a}}}
\end{align*}}{
\begin{align*}
\left| \bar{X}_a^{(t)} - T_{b \to a}  \right| &< \Phi^{-1}\left( 1 - \frac{\theta_t}{2}\right) \sqrt{\Var{\bar{X}_a^{(t)}} + \Var{T_{b \to a}}}\\
&= \Phi^{-1}\left( 1 - \frac{\theta_t}{2}\right) \sqrt{\frac{\sigma_a^2}{t} + \Var{T_{b \to a}}}
\end{align*}}
and $0$, otherwise, where $\theta_t \in [0,1]$ and $t=t_{\kappa_{b \to a}}$. Additionally, $\chi_a^{(t)}(a;\theta_t)=1$ always, and we set $\chi_a^{(t)}(b;\theta_t)=1$ before agent $a$ receives from agent $b$ for the first time.

If we keep the last update, i.e.,  $w_1=\cdots=w_{\kappa_{b \to a}-1}=0$ and $w_{\kappa_{b \to a}}=1$, the sum above becomes $\bar X_b^{(t_{\kappa_{b \to a}})}$, which is asymptotically Gaussian by the central limit theorem.
But in general, this decision rule leads to an asymptotically correct estimate of the class $\mathcal C_a$, even when asymptotic Gaussness cannot be proved, but more general conditions are satisfied.

\subsubsection{Unknown Data Variance} \label{sec:unknown_data_variance}

We pick a decision rule based on the hypothesis test outlined above in \cref{sec:tools} (unknown variance case), i.e., we let $\chi_a^{(t)}(b; \theta_t)=1$, for $b \neq a$, if
\begin{align*}
\left| \bar{X}_a^{(t)} - T_{b \to a}  \right| 
&< \Phi^{-1}_{\text{t},\nu}\left( 1 - \frac{\theta_t}{2}\right) \sqrt{\frac{V_a^{(t)}}{t} + \hatVar{T_{b \to a}}}
\end{align*} 
and $0$, otherwise, where $\theta_t \in [0,1]$ and $t=t_{\kappa_{b \to a}}$. Additionally, $\chi_a^{(t)}(a;\theta_t)=1$ always, and we set $\chi_a^{(t)}(b;\theta_t)=1$ before agent $a$ receives sufficient data from agent $b$  for the first time, in order to compute an estimate of the data variance $\sigma_b^2$, or directly a shared noise-corrupted estimate of $\sigma_b^2$. Here, 
\ifthenelse{\boolean{THESIS_VERSION}}{
\begin{align*}
V_a^{(t)} &= \frac{1}{t-1} \sum_{j=1}^t \left( X_{a}^{(j)} - \bar{X}_a^{(t)} \right)^2 %
= \frac{1}{t-1} \left( \sum_{j=1}^t \big( X_{a}^{(j)} \big)^2 - t \big( \bar{X}_a^{(t)} \big)^2 \right)
\end{align*}}{
\begin{align*}
V_a^{(t)} &= \frac{1}{t-1} \sum_{j=1}^t \left( X_{a}^{(j)} - \bar{X}_a^{(t)} \right)^2 \\
&= \frac{1}{t-1} \left( \sum_{j=1}^t \big( X_{a}^{(j)} \big)^2 - t \big( \bar{X}_a^{(t)} \big)^2 \right)
\end{align*}}
is the sample variance based on agent $a$'s data, 
\ifthenelse{\boolean{THESIS_VERSION}}{
\begin{IEEEeqnarray}{rCl}
  \hatVar{T_{b \to a}}& = &V_{b \to a} \sum_{i=1}^{\kappa_{b \to a}} (t_i-t_{i-1}) \left( \sum_{j=i}^{\kappa_{b \to a}} \frac{w_j}{t_j}\right)^2 %
  + \Var{\sum_{i=1}^{\kappa_{b \to a}} w_i Z_{b \to a}^{(t_i)}}, \nonumber
\end{IEEEeqnarray}}{
\begin{IEEEeqnarray}{rCl}
  \hatVar{T_{b \to a}}& = &V_{b \to a} \sum_{i=1}^{\kappa_{b \to a}} (t_i-t_{i-1}) \left( \sum_{j=i}^{\kappa_{b \to a}} \frac{w_j}{t_j}\right)^2 \nonumber\\
  && +\> \Var{\sum_{i=1}^{\kappa_{b \to a}} w_i Z_{b \to a}^{(t_i)}}, \nonumber
\end{IEEEeqnarray}}
which is obtained by replacing $\sigma_b^2$ in the expression for the variance of $T_{b \to a}$ in \eqref{eq:var_Tab} by its current estimate $V_{b \to a}$, 
and  the number of degrees is freedom is %
\begin{align*} %
\nu = \frac{\left(\frac{V_a^{(t)}}{t} + \hatVar{T_{b \to a}}\right)^2}{\left(\frac{V_a^{(t)}}{t}\right)^2 \cdot \frac{1}{t-1}+ \left(\hatVar{T_{b \to a}}\right)^2 \cdot \frac{1}{t_{\kappa_{b \to a}}-1}}.
\end{align*}%
\subsection{Algorithm} \label{sec:alg}

We summarize the proposed algorithm in \cref{alg:ss}, which is based on the linear statistic of \cref{sec:statistic}. The crucial step that differentiates \cref{alg:ss} from the \texttt{ColME} algorithm in \cite[Alg.~1]{AsadiBelletMaillardTommasi22_1} is the design of a new decision rule $\chi_a^{(t)}(b; \theta_t)$ in \cref{line:decision_rule}, which is based on %
hypothesis testing as outlined in \cref{sec:decision_rule}. Moreover, we consider a more general linear statistic $T_{b \to a}$ in \cref{line:MoMupdate} (the \texttt{ColME} algorithm corresponds to fixing the last weight $w_{\kappa_{b \to a}}$ equal to one).
The selection of an agent $b$ to query
is done according to some schedule, e.g., round-robin (RR), denoted by \texttt{choose\_agent}. The statistic $V_{b \to a}$ is updated in \cref{line:updateV} based on the specific variance estimation scheme considered (see \cref{sec:variance_estimation} for further details).
Finally, 
both $T_{b \to a}$ and $\mathcal{C}^{(t)}_a$ are updated and the statistics $T_{b \to a}$, for $b \in \mathcal{C}_a^{(t)} \setminus \{ a \}$, are linearly combined in \cref{line:weights} in order to obtain an improved estimate $\mu_a^{(t)}$ of the mean of agent $a$ at time step $t$. The linear combination coefficients in the known variance case %
are optimized based on Lemma~\ref{lem:weighting}, i.e., selected according to %
\begin{align} \label{eq:alpha}
	\alpha_{b \to a}^{(t)} = \begin{cases}
		\frac{t}{\sigma_a^2 \left( \sum_{b' \in \mathcal C_a^{(t)} \setminus \{a\}} \frac{1}{\Var{T_{b' \to a}}}  + \frac{t}{\sigma_a^2}\right)} & \text{ if $b=a$}, \\
		\frac{1}{\Var{T_{b \to a}} \left( \sum_{b' \in \mathcal C_a^{(t)} \setminus \{a\}} \frac{1}{\Var{T_{b' \to a}}}  + \frac{t}{\sigma_a^2}\right)} & \text{ otherwise},
	\end{cases}
\end{align}
while in \cite[Alg.~1]{AsadiBelletMaillardTommasi22_1} several different (heuristic) linear combination schemes are considered.

From \cref{line:weights} of \cref{alg:ss} and Lemma~\ref{lem:weighting} it follows that
\begin{align} \label{eq:var_muat}
\Var{\mu_a^{(t)}} = \frac{1}{\sum_{b' \in \mathcal C_a^{(t)} \setminus \{a\}} \frac{1}{\Var{T_{b' \to a}}}  + \frac{t}{\sigma_a^2}}.
\end{align}
If there has been no values received from an agent $b$, we assume as a convention that $T_{b \to a} = 0$ and $\Var{T_{b \to a}} = +\infty$.

In the unknown data variance case, again building on  Lemma~\ref{lem:weighting}, the linear combination coefficients (which are now RVs) are selected according to
\begin{align} \label{eq:alpha_unknown_var}
	\alpha_{b \to a}^{(t)} = \begin{cases}
		\frac{t}{V_a^{(t)} \left( \sum_{b' \in \mathcal C_a^{(t)} \setminus \{a\}} \frac{1}{\hatVar{T_{b' \to a}}}  + \frac{t}{V_a^{(t)}}\right)} & \text{if $b=a$}, \\
		\frac{1}{\hatVar{T_{b \to a}} \left( \sum_{b' \in \mathcal C_a^{(t)} \setminus \{a\}} \frac{1}{\hatVar{T_{b' \to a}}}  + \frac{t}{V_a^{(t)}}\right)} & \text{otherwise}.
	\end{cases}
\end{align}%
Again, if agent $a$ has not received sufficient data from agent $b$ in order to compute an estimate of the data variance $\sigma_b^2$ or directly a shared noise-corrupted estimate of $\sigma_b^2$, as a convention we set $\hatVar{T_{b \to a}} = +\infty$.

\begin{remark} \label{rem:1}
Note that in the case of known data variance, it is straightforward to show that $\eE{\mu_a^{(t)}}=\mu_a$ when $\mathcal{C}_a^{(t)} = \mathcal{C}_a$. However, in the unknown data variance case, the $\alpha$'s are RVs and then it is not apparent that this result holds. %
The difficulty arises from the fact that the expectation of a product of RVs is in general not equal to the product of the individual expectations. 
\end{remark}

\begin{algorithm}[t]
	\KwIn{agent $a$%
		}	
		\KwOut{$\mu_a^{(t_{\max})}$}
		$\forall\, b \in [\numa] \setminus \{a\}: T_{b \to a} \gets 0, \kappa_{b \to a} \gets 0$\\
		$\mathcal C_a^{(0)} \gets [\numa]$\\
		\For{$t=1,2,\dotsc,t_{\max}$}{
			\tcp{Receive}
			Receive sample $X_a^{(t)} \sim \set D_a$ \\
			$\bar X_a^{(t)} \gets \bar X_a^{(t-1)} \times \frac{t-1}{t} + X_a^{(t)} \times \frac{1}{t}$ \\
				\tcp{Query}
				$b \leftarrow \mathtt{choose\_agent}\left(\mathcal C_a^{(t-1)},[\numa]
				\right)$ \\
					$\kappa_{b \to a} \gets \kappa_{b \to a} + 1$ \\
					$T_{b \to a} \gets \sum_{i=1}^{\kappa_{b \to a}} w_i \left(\bar{X}_b^{(t_i)} + Z_{b \to a}^{(t_i)} \right)$\label{line:MoMupdate}\\
					Update $V_{b \to a}$ \label{line:updateV}\\
				\tcp{Estimate}
				$\set C_a^{(t)} \gets \{ b \in [\numa] : \chi_a^{(t)}(b; \theta_t) = 1 \}$\label{line:decision_rule}\\
				$\mu_a^{(t)} \gets \alpha_{a \to a}^{(t)} \bar X_a^{(t)} + \sum_{b \in \mathcal C_a^{(t)} \setminus \{a\}} \alpha_{b \to a}^{(t)} T_{b \to a}$ \label{line:weights}\\
		}
		\KwRet{$\mu_a^{(t_{\max})}$}
		\caption{Private-ColME}
		\label{alg:ss}
\end{algorithm}

\subsection{Schedules}

As mentioned above in \cref{sec:alg}, the agents are queried according to some schedule. In this work, we study a simple RR schedule in which agents are queried in order of their indices (but skipping the agent $a$ itself). Additionally, we consider a \emph{restricted} RR  (rRR) schedule in which the agents are queried in the same order as in RR, but at any current time step $t$ the agents not in $\mathcal C_a^{(t-1)}$ are skipped.

\section{Privacy} \label{sec:privacy}

Below, we present two private release mechanisms giving different trade-offs between the variance of the linear statistic $T_{b \to a}$ (see \cref{line:MoMupdate} of \cref{alg:ss}) and the overall privacy level for each individual sample $X_b^{(t)}$ of agent $b$ when releasing noise-corrupted sample means to agent $a$. %

Both mechanisms are based on the following idea, which can be seen as a generalization of Lemma~\ref{lem:DP}. To construct a privatized version of $\bar X_b^{(t)}$ for release to agent $a$,  agent $b$ splits the corresponding sum of values with indices $[1:t]$ into $k$ subsums (called \emph{p-sums} in \cite{ChanShiSong11_1}) with indices $[1:\tau_1], [\tau_1+1:\tau_2], \dotsc, [\tau_{k-1}+1:t]$:
\ifthenelse{\boolean{THESIS_VERSION}}{
  \begin{IEEEeqnarray*}{rCl}
      \bar{X}_b^{(t)}=\frac{X_b^{(1)} + \dotsb + X_b^{(t)}}{t} %
    & = &\frac{\sum_{i=1}^{\tau_1}X_b^{(i)}+\sum_{i=\tau_1+1}^{\tau_2}X_b^{(i)}+\cdots+\sum_{i=\tau_{k-1}+1}^{t}X_b^{(i)}}{t}\IEEEeqnarraynumspace\label{eq:split-into_subsums}
  \end{IEEEeqnarray*}}{
  \begin{IEEEeqnarray*}{rCl}
    \IEEEeqnarraymulticol{3}{l}{%
      \bar{X}_b^{(t)}=\frac{X_b^{(1)} + \dotsb + X_b^{(t)}}{t}}\nonumber\\*\,\,%
    & = &\frac{\sum_{i=1}^{\tau_1}X_b^{(i)}+\sum_{i=\tau_1+1}^{\tau_2}X_b^{(i)}+\cdots+\sum_{i=\tau_{k-1}+1}^{t}X_b^{(i)}}{t}\IEEEeqnarraynumspace\label{eq:split-into_subsums}
  \end{IEEEeqnarray*}}  
  and adds independent noise with the same variance %
  $\sigmaDP^2$ (whose value is determined by the underlying DP noise distribution and the privacy parameters $(\epsilon,\delta)$) to each of the subsums:
 \ifthenelse{\boolean{THESIS_VERSION}}{ \begin{IEEEeqnarray*}{rCl}
    \frac{\sum_{i=1}^{\tau_1}X_b^{(i)}+ Z_{b \to a}^{(1:\tau_1)}}{t} 
    + 
    \frac{\sum_{i=\tau_1+1}^{\tau_2}X_b^{(i)}+Z_{b \to a}^{(\tau_1+1:\tau_2)}}{t}
    + \cdots+\frac{\sum_{i=\tau_{k-1}+1}^{t}X_b^{(i)}+ Z_{b \to a}^{(\tau_{k-1}+1:t)}}{t}
    =\bar{X}_b^{(t)} + Z_{b \to a}^{(t)},\IEEEeqnarraynumspace
  \end{IEEEeqnarray*}}{
  \begin{IEEEeqnarray*}{rCl}
    &&\frac{\sum_{i=1}^{\tau_1}X_b^{(i)}+ Z_{b \to a}^{(1:\tau_1)}}{t} 
    + 
    \frac{\sum_{i=\tau_1+1}^{\tau_2}X_b^{(i)}+Z_{b \to a}^{(\tau_1+1:\tau_2)}}{t}
    \nonumber\\
    && +\>\cdots+\frac{\sum_{i=\tau_{k-1}+1}^{t}X_b^{(i)}+ Z_{b \to a}^{(\tau_{k-1}+1:t)}}{t}
    =\bar{X}_b^{(t)} + Z_{b \to a}^{(t)},\IEEEeqnarraynumspace
  \end{IEEEeqnarray*}}
 where all $Z_{b \to a}^{(1:\tau_1)}, Z_{b \to a}^{(\tau_1+1:\tau_2)}, \dots, Z_{b \to a}^{(\tau_{k-1}+1:t)}$ are i.i.d. according to the distribution $\set{P}_{Z_{b \to a}^{(\tau_{i-1}+1:\tau_i)}}(0,\sigmaDP^2)$, and
  \begin{IEEEeqnarray*}{c}
    Z_{b \to a}^{(t)}\triangleq\frac{1}{t} \sum_{i=1}^{k}Z_{b \to a}^{(\tau_{i-1}+1:\tau_i)} \sim \set{P}_{Z_{b \to a}^{(t)}\left(0,\frac{k\sigmaDP^2}{t^2}\right)},\IEEEeqnarraynumspace
  \end{IEEEeqnarray*}
where $\tau_0=0$ and $\tau_k=t$. Note that the variance of the noise $Z_{b \to a}^{(t)}$ depends only on the time of release, $t$, the number of subsums in the split, $k$, and the desired $(\epsilon,\delta)$.
	
Similar to Lemma~\ref{lem:DP}, this release mechanism provides $(\epsilon,\delta)$-DP for each $X_b^{(1)}$, $X_b^{(2)}$, \dots, $X_b^{(t)}$. However, a subsum with the corresponding noise can be re-used by the agent $b$ further for constructing privatized versions of the means $\bar X_b^{(t')}$ for $t' > t$, thus reducing the amount of ``fresh'' noise added. If for example 
{$X_b^{(1)} + \dotsb + X_b^{(\tau_1)} + Z_{b \to a}^{(1:\tau_1)}$ 
is released several times (with the same value of $Z_{b \to a}^{(1:\tau_1)}$) as a subsum of different sums, the privacy of each of the $X_b^{(1)},\dotsc, X_b^{(\tau_1)}$ stays the same.

The only difference between the two mechanisms below is how we split into the subsums. In both mechanisms, if the same subsum (i.e., with the same interval of indices) needs to be used for the calculation of several privatized means by agent $b$ for release to agent $a$, we actually \emph{require} that the exact same value of noise is used for this subsum.

We stress here that in both mechanisms  below, the RVs $Z_{b \to a}^{(t)}$ are \emph{dependent} for $t=t_1, t_2, \dotsc$, and the variance of the linear statistic $T_{b \to a}$, needed for the implementation of the decision rule $\chi_a^{(t)}(b; \theta_t)$ and the computation of the coefficients $\alpha_{b \to a}^{(t)}$ in \eqref{eq:alpha},  depends on it  and the weights $\{ w_i \}$ used. An analytical expression for the variance (specific to the weights and the privacy mechanism) can be derived from \eqref{eq:var_Tab}, and is given in   Appendix~\ref{app:Tba_variances}. %

\subsection{Privacy Mechanism I (PM-I)} \label{sec:schemeII}

This mechanism is inspired by the Simple Counting Mechanism II in \cite{ChanShiSong11_1}. The split of sums into $k=\kappa$  subsums as above now exactly corresponds to the times $t_1, t_2, \dotsc$ when agent $b$ is queried by agent $a$, i.e., $[1:t_\kappa]$ is split into $[1:t_1], [t_1+1:t_2], \dotsc, [t_{\kappa-1}+1:t_\kappa]$. Hence, $\bar X_b^{(t_\kappa)} + Z_{b \to a}^{(t_\kappa)} = \bar X_b^{(t_\kappa)} + \nicefrac{1}{t_\kappa}\sum_{i=1}^\kappa Z_{b \to a}^{(t_{i-1}+1:t_i)}$, and $Z_{b \to a}^{(t_\kappa)} \sim \set{P}_{Z_{b \to a}^{(t_\kappa)}}(0,\nicefrac{\kappa\sigmaDP^2}{t_\kappa^2})$.

Note that PM-I allows for efficient implementation by the agent $b$. Indeed, it can keep only the current value of 
  $\sum_{i=1}^\kappa Z^{(t_{i-1}+1:t_i)}_{b \to a}$.
At the next release time $t = t_{\kappa+1}$, it updates this cumulative noise by adding ``fresh'' noise $Z_{b \to a}^{(t_\kappa+1:t_{\kappa+1})}$ and releases $\bar X_b^{(t_{\kappa+1})}$ privatized with this updated noise (divided by $t_{\kappa+1}$). In particular, agent $b$ does not need to keep $t_1, \dotsc, t_\kappa$. Hence, agent $b$ needs $O(1)$ memory to implement PM-I.

Since every $X_b^{(t)}$, $1 \le t \le t_{\kappa}$, participates in an exactly one subsum (defined by the aforementioned split), PM-I allows for a constant  privacy level as we query agent $b$ from agent $a$, rather than getting weaker and weaker over time, while at the same time having a decreasing DP noise variance due to the factor $\nicefrac{\kappa}{t_\kappa^2}$.

\begin{lemma} \label{lem:keeping_last_optimal_schemeII}
Consider  RR or rRR scheduling and an oracle class estimator, i.e., $\mathcal{C}_a^{(t)} = \mathcal{C}_a$ for all $t$. Next, let agent $b$ be the last agent  queried by agent $a$ in a single round. Then, selecting $w_1=\cdots=w_{\kappa_{b \to a}-1}=0$ and $w_{\kappa_{b \to a}}=1$ minimizes the variance of $T_{a \to b}$.
\end{lemma}

\begin{IEEEproof}
See Appendix~\ref{sec:proof_keeping_last_optimal_schemeII} in the supplementary material.
\end{IEEEproof}

Hence, based on Lemma~\ref{lem:keeping_last_optimal_schemeII}, keeping the last update (as proposed in \cite{AsadiBelletMaillardTommasi22_1}) is a good strategy for PM-I.

\subsection{Privacy Mechanism II (PM-II)}
\label{sec:PM-II}
This mechanism is inspired by  the Binary Counting Mechanism in \cite{ChanShiSong11_1}.
When an agent $b$ releases data to agent $a$ for the $\kappa$-th time (at time instant $t_\kappa$), we construct the split into $k=w_{\mathrm{H}}(\kappa)$ subsums in two steps as follows. First, consider the same split of $[1:t_\kappa]$ into  $[1:t_1], [t_1+1:t_2], \dotsc, [t_{\kappa-1}+1:t_\kappa]$ as for PM-I. Second, we now join the corresponding subsums into larger subsums according to the binary representation of $\kappa$. More precisely, let $\kappa = 2^{s_1} + 2^{s_2} + \dotsb + 2^{s_{w_{\mathrm{H}}(\kappa)}}$, $s_i > s_{i+1}$, be the unique representation of $\kappa$ based on positions of ones in the binary representation of $\kappa$.\footnote{E.g., if $\kappa=13=1101$, we represent it as $\kappa = 2^3 + 2^2 + 2^0$.} Then, we join the first $2^{s_1}$ aforementioned subsums into the first larger subsum, the next $2^{s_2}$ subsums into the second larger subsum, etc. In total, these two steps result in splitting $[1:t_\kappa]$ as follows: %
\begin{center}
	\begin{tabular}{ll}
	$[1:t_{2^{s_1}}]$ & (first subsum),\\
	$[t_{2^{s_1}}+1:t_{2^{s_1} + 2^{s_2}}]$ & (second subsum), \\
	$[t_{2^{s_1} + 2^{s_2}}+1:t_{2^{s_1} + 2^{s_2}+2^{s_3}}]$ & (third subsum),\\
        \qquad\vdots
        \\
	$[t_{2^{s_1} + 2^{s_2}+\dotsb+2^{s_{w_{\mathrm{H}}(\kappa)-1}}}+1:t_\kappa]$ & ($w_{\mathrm{H}}(\kappa)$-th subsum).
	\end{tabular}
\end{center}
Finally, we add independent noise with variance $\sigmaDP^2$ to each of the corresponding subsums and construct the noisy mean $\bar X_b^{(t_\kappa)} + Z_{b \to a}^{(t_\kappa)}$ as previously. This mechanism gives
$Z_{b \to a}^{(t_\kappa)} \sim \set{P}_{Z_{b \to a}^{(t_\kappa)}}(0,\nicefrac{w_{\mathrm{H}}(\kappa) \sigmaDP^2}{t_\kappa^2})$.
The noise term $Z_{b \to a}^{(t_\kappa)}$ has variance at most $\nicefrac{(\left \lfloor \log_2 \kappa \right \rfloor+1)}{t_\kappa^2}$ times $\sigmaDP^2$ as $w_{\mathrm{H}}(\kappa) \leq \left \lfloor \log_2\kappa \right \rfloor+1$. The variance is a factor of at most $\nicefrac{(\left \lfloor \log_2\kappa \right \rfloor+1)}{\kappa}$ of the variance of PM-I and this factor approaches zero in $\kappa$.
On the other hand, in contrast to PM-I, a value $X_b^{(t)}$ for $1 \le t \le t_{\kappa}$ can be used in up to $\lfloor \log_2 \kappa \rfloor + 1$ different subsums.
Therefore, from the composition theorem of DP (see, e.g.,  \cite[Thm.~3.1]{KairouzOhViswanath17_1}), PM-II 
will give $((\left \lfloor \log_2 \kappa \right \rfloor +1)\epsilon,(\left \lfloor \log_2 \kappa \right \rfloor+1)\delta)$-DP with respect to each individual sample $X_b^{(1)},\dotsc,X_b^{(t_{\kappa})}$. %

Compared to PM-I,  the privacy parameters of this mechanism grow with $\kappa$, and hence the privacy level gets weaker over time. On the other hand, the DP noise variance decreases faster with $\kappa$. As shown below in \cref{sec:numerical_results}, this may result in reaching a given target average mean squared error for a given \emph{fixed} overall privacy level faster in some scenarios.  %
On the other hand, in contrast to PM-I, agent $b$ needs $O(\lfloor \log_2\kappa\rfloor)$ memory to implement PM-II.

Interestingly, keeping only the last update does not minimize $\Var{T_{b \to a}}$ as for PM-I (cf.\ Lemma~\ref{lem:keeping_last_optimal_schemeII}). For this particular mechanism we will illustrate in \cref{sec:numerical_results} below that a \emph{windowed} MoM (wMoM) approach where  $w_1=\cdots=w_{2^{\lfloor \log_2 \kappa_{b \to a} \rfloor}-1}=0$ and $w_{2^{\lfloor \log_2 \kappa_{b \to a} \rfloor}}=\cdots=w_{\kappa_{b \to a}}=\nicefrac{1}{(\kappa_{b \to a}-2^{\lfloor \log_2 \kappa_{b \to a} \rfloor}+1)}$ performs better.  %

\section{Data Variance Estimation} \label{sec:variance_estimation}

In this section, we propose two data variance estimation methods. For the first method (inspired by the release mechanisms of \cref{sec:privacy}), each agent (when queried) releases its most recent sample variance computed from $k$  \emph{partial} sample variances, analogue to the $k$ \emph{p-sums} of \cref{sec:privacy}.  %
In order to preserve privacy, the releases are protected by additive DP noise.  For the second method there is no additional release of data, but the data variances are instead estimated from the released noise-corrupted sample means. As shown below, both methods have their advantages and drawbacks. For brevity, we will write in this section $\kappa$ instead of $\kappa_{b \to a}$.

\subsection{Releasing the Sample Variance} \label{sec:privacy_sample_variance_1}

We can run either PM-I or PM-II considering sample variances based on parts of the individual data entries in addition to partial sums of data entries, but with noise variance $\sigmaDPvar^2$ selected according to Lemma~\ref{lem:DP_squared}, i.e., agent $b$ computes
 \ifthenelse{\boolean{THESIS_VERSION}}{
\begin{align}
\tilde{V}_{b}^{(i)}&=\sum_{j=\tau_{i-1}+1}^{\tau_i} \big( X_b^{(j)} \big)^2 - \frac{\left(\sum_{j=\tau_{i-1}+1}^{\tau_i} X_b^{(j)} \right)^2}{\tau_i-\tau_{i-1}} %
+ \frac{\tau_i-\tau_{i-1}-1}{\tau_i-\tau_{i-1}}W_{b \to a}^{(\tau_{i-1}+1:\tau_i)}  \label{eq:Vbi} %
\end{align}}{
\begin{align}
\tilde{V}_{b}^{(i)}&=\sum_{j=\tau_{i-1}+1}^{\tau_i} \big( X_b^{(j)} \big)^2 - \frac{\left(\sum_{j=\tau_{i-1}+1}^{\tau_i} X_b^{(j)} \right)^2}{\tau_i-\tau_{i-1}} \nonumber \\
&\quad + \frac{\tau_i-\tau_{i-1}-1}{\tau_i-\tau_{i-1}}W_{b \to a}^{(\tau_{i-1}+1:\tau_i)}  \label{eq:Vbi} %
\end{align}}
locally for $i \in [k]$ for the $\kappa$-th query from agent $a$, 
where the  RVs $W_{b \to a}^{(1:\tau_1)},\ldots,W_{b \to a}^{(\tau_{k-1}+1:t)}$ are i.i.d. according to $\set{P}_{W_{b \to a}^{(\tau_{i-1}+1:\tau_i)}}(0,\sigmaDPvar^2)$. 
Now, let
\begin{align}
\vardbtilde{V}_b^{(i)} &= \tilde{V}_b^{(i)}  + \frac{\left(\sum_{j=\tau_{i-1}+1}^{\tau_i} X_b^{(j)} + Z_{b \to a}^{(\tau_{i-1}+1:\tau_i)}\right)^2}{\tau_i-\tau_{i-1}}.  \label{eq:expression1} %
\end{align}
$\vardbtilde{V}_b^{(i)}$ can be computed locally at agent $b$ since the term added to $\tilde{V}_b^{(i)}$ is available locally at agent $b$ (see \cref{sec:privacy}). %
Next, let
 \ifthenelse{\boolean{THESIS_VERSION}}{
\begin{align} \label{eq:Vba}
V_{b \to a}&= \frac{1}{t-1}\sum_{i=1}^{k} \vardbtilde{V}_b^{(i)} - \frac{t}{t-1} \big( \bar{X}_b^{(t)} + Z_{b \to a}^{(t)} \big)^2 %
- \frac{\sigmaDP^2}{t-1} \left( \sum_{i=1}^{k} \frac{1}{\tau_i-\tau_{i-1}} - \frac{{k}}{t} \right). 
\end{align}}{
\begin{align} \label{eq:Vba}
V_{b \to a}&= \frac{1}{t-1}\sum_{i=1}^{k} \vardbtilde{V}_b^{(i)} - \frac{t}{t-1} \big( \bar{X}_b^{(t)} + Z_{b \to a}^{(t)} \big)^2 \nonumber \\
&\quad - \frac{\sigmaDP^2}{t-1} \left( \sum_{i=1}^{k} \frac{1}{\tau_i-\tau_{i-1}} - \frac{{k}}{t} \right). 
\end{align}}
Again, $V_{b \to a}$ can be computed locally at agent $b$ as the first term subtracted is available due to \cref{line:MoMupdate} of \cref{alg:ss} (see also \cref{sec:privacy}) and the second term subtracted is a constant. $V_{b \to a}$ is the desired  variance estimate that can be released to agent $a$ in response to the $\kappa$-th query to agent $b$ from agent $a$. 
\begin{lemma} \label{lem:Sch_I_bias}
The variance estimate $V_{b \to a}$ in \eqref{eq:Vba} is an unbiased estimator for $\sigma_b^2$.
\end{lemma}
\begin{IEEEproof}
$V_{b \to a}$ can be re-written  using \eqref{eq:Vbi} and \eqref{eq:expression1}   as
\begin{align}
V_{b \to a}&=\frac{1}{t-1} \sum_{j=1}^{t} \big( X_b^{(j)} \big)^2 -
\frac{t}{t-1} \big( \bar{X}_b^{(t)} \big)^2 \label{eq:sample_variance}\\
 &\quad+\frac{1}{t-1} \sum_{i=1}^{k} \frac{\big (Z_{b \to a}^{(\tau_{i-1}+1:\tau_i)} \big)^2}{\tau_i-\tau_{i-1}} \label{eq:sample_variance_noise-0}\\
&\quad+ \frac{1}{t-1} \sum_{i=1}^{k} \frac{\tau_i-\tau_{i-1}-1}{\tau_i-\tau_{i-1}}W_{b \to a}^{(\tau_{i-1}+1:\tau_i)} \label{eq:sample_variance_noise-1} \\
&\quad+  \frac{2}{t-1} \sum_{i=1}^{k} \frac{Z_{b \to a}^{(\tau_{i-1}+1:\tau_i)} \sum_{j=\tau_{i-1}+1}^{\tau_i} X_b^{(j)}}{\tau_i-\tau_{i-1}}  \label{eq:sample_variance_noise-2}\\
&\quad- \frac{2}{t-1} \bar{X}_b^{(t)} \sum_{i=1}^{k} Z_{b \to a}^{(\tau_{i-1}+1:\tau_i)} \label{eq:sample_variance_noise-3}\\
&\quad-  \frac{1}{t(t-1)} \left( \sum_{i=1}^{k} Z_{b \to a}^{(\tau_{i-1}+1:\tau_i)} \right)^2 \label{eq:sample_variance_noise-3a} \\
&\quad- \frac{\sigmaDP^2}{t-1} \left( \sum_{i=1}^{k} \frac{1}{\tau_i-\tau_{i-1}} - \frac{{k}}{t} \right),\label{eq:sample_variance_noise-4}
\end{align}
where the expression in \eqref{eq:sample_variance} is the sample variance, denoted in the following by $V_b^{(t)}$, with mean $\sigma_b^2$. All noise terms in \eqref{eq:sample_variance_noise-1} to \eqref{eq:sample_variance_noise-3} have zero mean, while the remaining squared noise terms in \eqref{eq:sample_variance_noise-0} and \eqref{eq:sample_variance_noise-3a} have mean equal to the term  that is subtracted in \eqref{eq:sample_variance_noise-4}. This is due to the fact that the data and the noise are independent and that the square of a zero-mean RV %
has mean equal to the variance of the RV being squared. Hence, the overall mean of the noise terms is equal to zero and we have an unbiased estimator for $\sigma_b^2$.
\end{IEEEproof}

To summarize, we can write
$V_{b \to a} = V_b^{(t)} + U_{b \to a}^{(t)}$,
where $V_b^{(t)}$ is the ``$X$-based'' sample variance in \eqref{eq:sample_variance} and $U_{b \to a}^{(t)}$ denotes  the overall noise.

Because of the additive noise when releasing partial sample variances, $V_{b \to a}$ might be negative in which case we set it to infinity.

\begin{remark}
Note that replacing  $X_b^{(i)}$ by its square $\big(X_b^{(i)}\big)^2$ in both PM-I and PM-II would not work as the sensitivity of a squared bounded value depends on the mean of the value being squared, which in our case is unknown. %
\end{remark}

\subsection{Using Current Releases} \label{sec:privacy_sample_variance_2}

Define
\begin{align*}
Y_b^{(i)} & \triangleq \frac{1}{\sqrt{\tau_i-\tau_{i-1}}}  \sum_{j=\tau_{i-1}+1}^{\tau_i}  X_b^{(j)}  \text{ and } \bar{Y}_b^{({k})} \triangleq \frac{\sum_{i=1}^{k} Y_b^{(i)}}{{k}}.
\end{align*}
Then,
$\Var{Y_b^{(i)}} = \sigma_b^2$. %
Moreover, define
\begin{align*} 
\tilde{Z}_{b \to a}^{(\tau_{i-1}+1:\tau_i)} \triangleq \frac{1}{\sqrt{\tau_i-\tau_{i-1}}} Z_{b \to a}^{(\tau_{i-1}+1:\tau_i)}
\end{align*}
with variance $\Var{\tilde{Z}_{b \to a}^{(\tau_{i-1}+1:\tau_i)}} = \frac{\sigmaDP^2}{\tau_i-\tau_{i-1}}$. 
Now, let %
 \ifthenelse{\boolean{THESIS_VERSION}}{\begin{align}
V_{b \to a}&=\frac{1}{{k}-1}  \sum_{i=1}^{{k}} \Big(Y_b^{(i)}+\tilde{Z}_{b \to a}^{(\tau_{i-1}+1:\tau_i)} \Big)^2  %
- \frac{\Big(\sum_{i=1}^{k} Y_b^{(i)} + \sum_{i=1}^{k} \tilde{Z}_{b \to a}^{(\tau_{i-1}+1:\tau_i)}\Big)^2}{{k} ({k}-1)} %
-\frac{\sigmaDP^2}{{k}}  \sum_{i=1}^{k} \frac{1}{\tau_i-\tau_{i-1}}.   \label{eq:sample_variance_noise_Y-1} 
\end{align}}{
\begin{align}
V_{b \to a}&=\frac{1}{{k}-1}  \sum_{i=1}^{{k}} \Big(Y_b^{(i)}+\tilde{Z}_{b \to a}^{(\tau_{i-1}+1:\tau_i)} \Big)^2  \nonumber \\
& \quad - \frac{\Big(\sum_{i=1}^{k} Y_b^{(i)} + \sum_{i=1}^{k} \tilde{Z}_{b \to a}^{(\tau_{i-1}+1:\tau_i)}\Big)^2}{{k} ({k}-1)} \nonumber \\
&\quad  -\frac{\sigmaDP^2}{{k}}  \sum_{i=1}^{k} \frac{1}{\tau_i-\tau_{i-1}}.   \label{eq:sample_variance_noise_Y-1} 
\end{align}}
Note that $V_{b \to a}$ (as defined above) can be computed locally at agent $b$ and can be considered as a  noise-corrupted ``$Y$-based'' sample variance.
\begin{lemma} \label{lem:Sch_II_bias}
The variance estimate $V_{b \to a}$ in \eqref{eq:sample_variance_noise_Y-1}  is an unbiased estimator for $\sigma_b^2$. 
\end{lemma}
\begin{IEEEproof}
\begin{align}
V_{b \to a}&=  \frac{1}{{k}-1}  \sum_{i=1}^{{k}} \Big(Y_b^{(i)} \Big)^2 -   \frac{{k}}{{k}-1} \Big( \bar{Y}_b^{({k})} \Big)^2 \label{eq:sample_variance_noise_Y-2}\\
 &\quad+\frac{1}{{k}-1} \sum_{i=1}^{k} \frac{\big (Z_{b \to a}^{(\tau_{i-1}+1:\tau_i)} \big)^2}{\tau_i-\tau_{i-1}} \notag \\ %
&\quad+  \frac{2}{{k}-1} \sum_{i=1}^{k} \frac{Z_{b \to a}^{(\tau_{i-1}+1:\tau_i)} \sum_{j=\tau_{i-1}+1}^{\tau_i} X_b^{(j)}}{\tau_i-\tau_{i-1}} \notag \\ %
&\quad- \frac{2}{{k}-1} \bar{Y}_b^{({k})} \sum_{i=1}^{k} \frac{Z_{b \to a}^{(\tau_{i-1}+1:\tau_i)}}{\sqrt{\tau_i-\tau_{i-1}}} \notag \\ %
&\quad-  \frac{1}{{k}({k}-1)} \left( \sum_{i=1}^{k} \frac{Z_{b \to a}^{(\tau_{i-1}+1:\tau_i)}}{\sqrt{\tau_i-\tau_{i-1}}} \right)^2 \notag \\ %
&\quad-\frac{\sigmaDP^2}{{k}}  \sum_{i=1}^{k} \frac{1}{\tau_i-\tau_{i-1}}, \label{eq:sample_variance_noise_Y-3}
\end{align}
from which it can be seen that $V_{b \to a}$ is an unbiased estimator for $\sigma_b^2$ as the mean of the squared noise terms equals the expression that is subtracted in \eqref{eq:sample_variance_noise_Y-3}. %
\end{IEEEproof} %

To summarize, we can write
$V_{b \to a} = V_b^{(t)} + U_{b \to a}^{(t)}$,
where $V_b^{(t)}$ is now the ``$Y$-based'' sample variance in \eqref{eq:sample_variance_noise_Y-2} and $U_{b \to a}^{(t)}$ denotes  the noise term.

\subsubsection{Improved Estimator (Bayesian Approach)} \label{sec:Bayesian_app}

Because of the additive noise,  $V_{b \to a}$ might be negative. In order to fix this issue we can apply a Bayesian approach with  a priori distribution on $\sigma_b^2$ in order to enforce the corresponding estimator to be nonnegative. We assume PM-I and RR scheduling, which implies that $k = \kappa$, $\tau_i = t_i$, and $\tau_i - \tau_{i-1} = M-1$  for all $i \in [k]$.  We proceed as follows.

First, re-write $V_{b \to a}$ from  \eqref{eq:sample_variance_noise_Y-1} as
\begin{align*}
V_{b \to a}&= V'_{b \to a}  -\frac{\sigmaDP^2}{{\kappa}}  \sum_{i=1}^{\kappa} \frac{1}{\tau_i-\tau_{i-1}} = V'_{b \to a} - K^{({\kappa})} \cdot \sigmaDP^2,
\end{align*}
where
 \ifthenelse{\boolean{THESIS_VERSION}}{
\begin{align*}
V'_{b \to a} &\triangleq \frac{1}{{\kappa}-1}  \sum_{i=1}^{{\kappa}} \Big(Y_b^{(i)}+\tilde{Z}_{b \to a}^{(\tau_{i-1}+1:\tau_i)} \Big)^2  %
- \frac{\Big(\sum_{i=1}^{\kappa} Y_b^{(i)} + \sum_{i=1}^{\kappa} \tilde{Z}_{b \to a}^{(\tau_{i-1}+1:\tau_i)}\Big)^2}{{\kappa} ({\kappa}-1)}
\end{align*}}{
\begin{align*}
V'_{b \to a} &\triangleq \frac{1}{{\kappa}-1}  \sum_{i=1}^{{\kappa}} \Big(Y_b^{(i)}+\tilde{Z}_{b \to a}^{(\tau_{i-1}+1:\tau_i)} \Big)^2   \\
& \quad - \frac{\Big(\sum_{i=1}^{\kappa} Y_b^{(i)} + \sum_{i=1}^{\kappa} \tilde{Z}_{b \to a}^{(\tau_{i-1}+1:\tau_i)}\Big)^2}{{\kappa} ({\kappa}-1)}
\end{align*}}
and
\begin{align*}
K^{({\kappa})} &\triangleq \frac{1}{{\kappa}}  \sum_{i=1}^{\kappa} \frac{1}{\tau_i-\tau_{i-1}} = \frac{1}{M-1} = \frac{\Var{\tilde{Z}_{b \to a}^{(\tau_{i-1}+1:\tau_i)}}}{\sigmaDP^2}.
\end{align*}

Moreover,  assume that $\sigmaDP^2 > 0$ and that $\sigma_b^2$ is a RV with an \emph{uninformative} prior  proportional to $\nicefrac{1}{\left(\sigma_b^2 + K^{({\kappa})} \sigmaDP^2 \right)^2}$ ($K^{({\kappa})} \sigmaDP^2$ is the variance of  $\tilde{Z}_{b \to a}^{(\tau_{i-1}+1:\tau_i)}$, $i \in [\kappa]$)  for nonnegative values of $\sigma_b^2$ and zero elsewhere, i.e.,\footnote{This prior distribution is chosen since then $\nicefrac{\sigma_b^2}{\left(\sigma_b^2 + K^{({\kappa})} \sigmaDP^2 \right)}$ is uniformly distributed over the interval $[0,1)$.}

\begin{align*}
{\rm Pr}\left( \sigma_b^2\right) \propto \begin{cases}
0 & \text{if $\sigma_b^2 < 0$},\\
\frac{1}{\left(\sigma_b^2 + K^{({\kappa})} {\sigmaDP^2}\right)^2} & \text{otherwise}.
\end{cases}
\end{align*}
Now, assuming that the $Y_b^{(i)}$'s are Gaussian RVs (which by the central limit theorem they are when the number of agents is sufficiently large) and that we use Gaussian DP noise, the posteriori probability for $\sigma_b^2$ equals
 \ifthenelse{\boolean{THESIS_VERSION}}{
 \begin{align*} 
&{\rm Pr}\left(\sigma_b^2 \Bigl\vert \{Y_b^{(i)}+\tilde{Z}_{b \to a}^{(\tau_{i-1}+1:\tau_i)} \}_{i=1}^{\kappa} \right) %
\propto \left(\sigma_b^2 + K^{({\kappa})} \sigmaDP^2  \right)^{-\frac{{\kappa}+2}{2}-1} {\mathrm e}^{-\frac{({\kappa}-1) V'_{b \to a}}{2\left(\sigma_b^2 +  K^{({\kappa})} \sigmaDP^2 \right)} }
\end{align*}}{
\begin{align*} 
&{\rm Pr}\left(\sigma_b^2 \Bigl\vert \{Y_b^{(i)}+\tilde{Z}_{b \to a}^{(\tau_{i-1}+1:\tau_i)} \}_{i=1}^{\kappa} \right) \\
&\quad \propto \left(\sigma_b^2 + K^{({\kappa})} \sigmaDP^2  \right)^{-\frac{{\kappa}+2}{2}-1} {\mathrm e}^{-\frac{({\kappa}-1) V'_{b \to a}}{2\left(\sigma_b^2 +  K^{({\kappa})} \sigmaDP^2 \right)} }
\end{align*}}
for $\sigma_b^2 \geq 0$ and zero elsewhere, which is a \emph{truncated}  inverse gamma distribution with shape parameter $\nicefrac{({\kappa}+2)}{2}$ and scale parameter $\nicefrac{({\kappa}-1) V'_{b \to a}}{2}$. The distribution is considered \emph{truncated} as $\sigma_b^2 \geq 0$ in our case, while for the \emph{standard} inverse gamma distribution we would have $\sigma_b^2 + K^{({\kappa})} \sigmaDP^2  \geq 0$. 
From the distribution above the  posterior expectation for $\sigma_b^2$ can be shown to be equal to
\begin{align*}
V'_{b \to a} \frac{{\kappa}-1}{2} \frac{f(-1)}{f(0)} -  K^{({\kappa})}\sigmaDP^2, 
\end{align*}
where 
$f(s) \triangleq  \gamma\left(\frac{{\kappa}+2}{2}+s,\frac{({\kappa}-1) V'_{b \to a}}{2 K^{({\kappa})} \sigmaDP^2 } \right)$ 
and $\gamma(s,x) \triangleq  \int_{0}^{x} t^{s-1} \mathrm{e}^{-t} \dd{t}$ %
 is the lower incomplete gamma function.\footnote{Note that changing the shape parameter $\frac{\kappa+2}{2}$ used in $f(\cdot)$ to $\frac{\kappa+1}{2}$ would correspond to selecting the so-called Jeffreys prior \cite{Jeffrey1961,Yang1998}, while changing it to $\frac{\kappa-2}{2}$  would correspond to a uniform, but improper prior \cite{Yang1998}.} %
Hence, we can define an improved estimator by 
\begin{align} \label{eq:improved_V}
V^{\mathrm{imp}}_{b \to a} &\triangleq \begin{cases}
V_{b \to a} & \text{if $V_{b \to a} \geq 0$}, \\
V'_{b \to a} \frac{{\kappa}-1}{2} \frac{f(-1)}{f(0)} -  K^{({\kappa})} \sigmaDP^2   & \text{otherwise}.
\end{cases}
\end{align}
Note that when using Laplace DP noise, we can still apply the same improved estimator in \eqref{eq:improved_V} as $\tilde{Z}_{b \to a}^{(\tau_{i-1}+1:\tau_i)}$ is approximately Gaussian by the central limit theorem for a large number of agents.

\section{Performance Analysis}

We first present the performance of a pure local approach in which each agent $a \in [\numa]$ estimates its mean solely based on its own data (see Proposition~\ref{prop:1} in \cref{sec:local}), then the ideal performance  where all data is public (see Proposition~\ref{prop:3} in \cref{sec:ideal}), and then that of \cref{alg:ss} with an oracle class estimator under RR scheduling (see Proposition~\ref{prop:2} in \cref{sec:oracle}).  For the local approach, the privacy is perfect as there is no sharing of data. Finally, we show that   the average mean squared error from  \cref{alg:ss} converges to zero faster than a pure local approach as $t \to \infty$ (see Theorem~\ref{thm:1} in \cref{sec:convergence}).

\subsection{Local Approach} \label{sec:local}
\begin{proposition}[Local] \label{prop:1}
The average mean squared error  of a pure local approach is
\begin{align*} %
\frac{1}{\numa} \sum_{a \in [\numa]}\EE{\left(\mu_a^{(t)} - \mu_a\right)^2} =  \frac{1}{\numa t}\sum_{a \in [\numa]}\sigma_a^2.
\end{align*}
\end{proposition}

\subsection{Ideal Performance}  \label{sec:ideal}
It is good to understand the limits of what can be achieved. If privacy is ignored, and agent $a$ knows $\mathcal C_a$ and has access to \emph{all} the data of all the agents in $\mathcal C_a$ at time $t$, it virtually has one large sample of size $|\mathcal C_a| t$ (as opposed to the sample of size $t$ for the pure local approach). With this, we have the following proposition.
\begin{proposition}[Ideal] \label{prop:3}
The average mean squared error of an \emph{ideal} scheme is
\begin{align*}
\frac{1}{\numa} \sum_{a \in [\numa]}\EE{\left(\mu_a^{(t)} - \mu_a\right)^2} = \frac{1}{\numa t}\sum_{a \in [\numa]}\frac{\sigma_a^2}{|\mathcal C_a|},
\end{align*}
and no approach can perform better than this ideal  performance.
\end{proposition}

\subsection{Oracle Performance}  \label{sec:oracle}

In this subsection, we present an analytical formula for the average mean squared error of \cref{alg:ss} for the case of an oracle class estimator (i.e., for the case when $\mathcal{C}_a^{(t)} = \mathcal{C}_a$) with  RR scheduling and under the assumption that the data variances are known and equal. In particular, we consider the case where all $\numa-1$ agents besides agent $a$ are assigned independently at random to a set of  $\Gamma$ classes with a probability of $p=\nicefrac{1}{\Gamma}$ of being assigned to each specific class. Then,  the number of agents in the same class as agent $a$, excluding agent $a$, is a RV following a binomial distribution with parameters $\numa-1$ (number of trials) and $p$ (success probability). 
   For this setup, we can prove the following theorem.
   
\begin{proposition}[Oracle With RR Sch.] \label{prop:2}
Assume $\sigma_a = \sigma_b = \sigma$ for all agents $a \neq b$ is known. %
Then, the average mean squared error of \cref{alg:ss} for the case of an oracle class estimator with RR scheduling is
\ifthenelse{\boolean{THESIS_VERSION}}{
\begin{align*} %
	&\frac{1}{\numa} \sum_{a \in [\numa]} \EE{\left( \mu_a^{(t)} - \mu_a \right)^2} = %
	\sum_{n=1}^{\numa} \sum_{\substack{1 \leq \ell_1 < \cdots < \ell_{n-1} < M}} %
	 p^{n-1} (1-p)^{\numa-n} / e_{n,t},
\end{align*}}{
\begin{align*} %
	&\frac{1}{\numa} \sum_{a \in [\numa]} \EE{\left( \mu_a^{(t)} - \mu_a \right)^2} = \\
	&  \sum_{n=1}^{\numa} \sum_{\substack{1 \leq \ell_1 < \cdots < \ell_{n-1} < M}} %
	 p^{n-1} (1-p)^{\numa-n} / e_{n,t},
\end{align*}}
where $e_{1,t} = \nicefrac{t}{\sigma^2}$ and for $2 \leq n \leq \numa$,
\ifthenelse{\boolean{THESIS_VERSION}}{
\begin{align*}
e_{n,t} &=  \frac{t}{\sigma^2} + 
\sum_{j=1}^{n-1}  \left( \sigma^2  \sum_{i=1}^{\kappa^{(t)}_{\ell_j \to a}}  
 \left(t^{(\ell_j)}_i-t^{(\ell_j)}_{i-1}\right) \left( \sum_{l=i}^{\kappa^{(t)}_{\ell_j \to a}} \frac{w_l}{t^{(\ell_j)}_l}\right)^2 %
  + \Var{\sum_{i=1}^{\kappa^{(t)}_{\ell_j \to a}} w_i Z_{\ell_j \to a}^{\left(t^{(\ell_j)}_i\right)}} \right)^{-1},
\end{align*}}{
\begin{align*}
e_{n,t} &=  \frac{t}{\sigma^2} + 
\sum_{j=1}^{n-1}  \left( \sigma^2  \sum_{i=1}^{\kappa^{(t)}_{\ell_j \to a}}  
 \left(t^{(\ell_j)}_i-t^{(\ell_j)}_{i-1}\right) \left( \sum_{l=i}^{\kappa^{(t)}_{\ell_j \to a}} \frac{w_l}{t^{(\ell_j)}_l}\right)^2 \right. \\
 &\quad + \left.
  \Var{\sum_{i=1}^{\kappa^{(t)}_{\ell_j \to a}} w_i Z_{\ell_j \to a}^{\left(t^{(\ell_j)}_i\right)}} \right)^{-1},
\end{align*}}
where $t^{(\ell_j)}_i = \thres+(i-1)(M-1)+\ell_j-1$ and 
\ifthenelse{\boolean{THESIS_VERSION}}{
\begin{align*}
\kappa^{(t)}_{\ell_j \to a} = \begin{cases}
1+ \left\lfloor \frac{t-\thres}{M-1} \right\rfloor  & %
\text{if $1 \leq \ell_j \leq (t-\thres) \bmod (M-1)+1$},\\
 \left\lfloor \frac{t-\thres}{M-1} \right\rfloor  & %
\text{if $(t-\thres) \bmod (M-1)+2 \leq \ell_j \leq M-1$}.
\end{cases}
\end{align*}}{
\begin{align*}
\kappa^{(t)}_{\ell_j \to a} = \begin{cases}
1+ \left\lfloor \frac{t-\thres}{M-1} \right\rfloor  & \\
&\hspace{-1.5cm}\text{if $1 \leq \ell_j \leq (t-\thres) \bmod (M-1)+1$},\\
 \left\lfloor \frac{t-\thres}{M-1} \right\rfloor  & \\
&\hspace{-1.5cm}\text{if $(t-\thres) \bmod (M-1)+2 \leq \ell_j \leq M-1$}.
\end{cases}
\end{align*}}
\end{proposition}

\begin{IEEEproof}
See Appendix~\ref{app:prop2} in the supplementary material. %
\end{IEEEproof}

By using the expressions for the noise variance in \eqref{eq:var_noise_term_schemeI} and \eqref{eq:var_noise_term_schemeII} (in Appendix~\ref{app:Tba_variances}) for PM-I and PM-II, respectively, the average mean squared error can be plotted as a function of time.  In order to reduce the computational complexity we restrict the range of $n$ from $\max(pM-15,1)$ (below) to  $\min(pM+15,M)$ (above), i.e., to a range around the average $pM$. Moreover, we also restrict the averaging over $\ell_1,\ldots,\ell_{n-1}$.

\begin{remark}
Note that a similar expression can be derived for rRR scheduling. In particular, by restricting $\ell_j=j$, removing the inner summation over $\ell_1,\ldots,\ell_{n-1}$, multiply by the binomial coefficient $\binom{M-1}{n-1}$, and replacing $M$ by $n$ in the expressions for $t^{(\ell_j)}_i$ and $\kappa^{(t)}_{\ell_j \to a}$ gives the corresponding expression for rRR scheduling. Further details are omitted for brevity.
\end{remark}

\ifthenelse{\boolean{THESIS_VERSION}}{}{
\ifthenelse{\boolean{QUICK_PLOTS}}
{\begin{figure*}[h!]
		\centering
		\subfloat{




\begin{tikzpicture}
\begin{axis}[%
width=\mywidth,
height=\myheight,
xmin=1.0,
xmax=30000,
xtick={1,5000,10000,15000,20000,25000,30000},
xlabel={Time $t$},
xticklabel style = {/pgf/number format/fixed, /pgf/number format/precision=2},
xlabel style={
	yshift=0.5ex,
	name=label,
	font=\small},
grid style={gray,opacity=0.5,dotted},
xmajorgrids,
ymajorgrids,
yminorgrids,
ymode=log,
ymin=1e-7,
ymax=1e-3,
ytick={1e-7,1e-6,1e-5,1e-4,1e-3},
ylabel={Average mean squared error},
legend style={font=\tiny},
ylabel style={
	yshift=-1.0ex,
	name=label,
	font=\small},
axis background/.style={fill=white},
legend cell align=left,
legend style={at={(axis cs: 30000,0.001)},anchor=north east,nodes={scale=0.75, transform shape}},
]


\addplot [color=darkgreen,dashed,line width=1pt, mark options={solid, line width = 0.5pt, fill=white}]table[x=time,y=acc_local] {data_final_version/local_curve.txt.100};
\addlegendentry{Local};




\addplot [color=red,solid,line width=1pt, mark options={solid, line width = 0.5pt, fill=white}]table[x=time,y=MSE] {data_final_version/Mark_pm1_MoM_RR_Hypo_eps1.000000_M200_20runs_together.csv.100};
\addlegendentry{PM-I (MoM, RR)};

\addplot [color=red,dotted,line width=1pt, mark options={solid, line width = 0.5pt, fill=white}]table[x=time,y=MSE] {data_final_version/Mark_pm1_MoM_restRR_Hypo_eps1.000000_M200_20runs_together.csv.100};
\addlegendentry{PM-I (MoM, rRR)}; 




\addplot [color=blue,solid,line width=1pt, mark options={solid, line width = 0.5pt, fill=white}]table[x=time,y=MSE] 
{data_final_version/Mark_pm1_nonMoM_RR_Hypo_eps1.000000_M200_20runs_together.csv.100};
\addlegendentry{PM-I (non-MoM, RR)};

\addplot [color=blue,dotted,line width=1pt, mark options={solid, line width = 0.5pt, fill=white}]table[x=time,y=MSE]
{data_final_version/Mark_pm1_nonMoM_restRR_Hypo_eps1.000000_M200_20runs_together.csv.100}; 
\addlegendentry{PM-I (non-MoM, rRR)}; 




\addplot [color=lightblue,solid,line width=1pt, mark options={solid, line width = 0.5pt, fill=white}]table[x=time,y=MSE] 
{data_final_version/Mark_pm2_nonMoM_RR_Hypo_eps1.000000_M200_20runs_together.csv.100};
\addlegendentry{PM-II (non-MoM, RR)};

\addplot [color=lightblue,dotted,line width=1pt, mark options={solid, line width = 0.5pt, fill=white}]table[x=time,y=MSE] 
{data_final_version/Mark_pm2_nonMoM_restRR_Hypo_eps1.000000_M200_20runs_together.csv.100};
\addlegendentry{PM-II (non-MoM, rRR)};


\addplot [color=darkyellow,dashed,line width=1pt, mark options={solid, line width = 0.5pt, fill=white}]table[x=time,y=error] {data_final_version/ideal_curve_M200.txt.100};
\addlegendentry{Ideal};

\end{axis}
\end{tikzpicture}%

		}
		\hfill
		\subfloat{




\begin{tikzpicture}
\begin{axis}[%
width=\mywidth,
height=\myheight,
xmin=1.0,
xmax=30000,
xtick={1,5000,10000,15000,20000,25000,30000},
xlabel={Time $t$},
xticklabel style = {/pgf/number format/fixed, /pgf/number format/precision=2},
xlabel style={
	yshift=0.5ex,
	name=label,
	font=\small},
grid style={gray,opacity=0.5,dotted},
xmajorgrids,
ymajorgrids,
yminorgrids,
ymode=log,
ymin=1e-7,
ymax=1e-3,
ytick={1e-7,1e-6,1e-5,1e-4,1e-3},
legend style={font=\tiny},
ylabel style={
	yshift=-1.0ex,
	name=label,
	font=\small},
axis background/.style={fill=white},
legend cell align=left,
legend style={at={(axis cs: 30000,0.001)},anchor=north east,nodes={scale=0.75, transform shape}},
]


\addplot [color=darkgreen,dashed,line width=1pt, mark options={solid, line width = 0.5pt, fill=white}]table[x=time,y=acc_local] {data_final_version/local_curve.txt.100};
\addlegendentry{Local};


\addplot [color=red,solid,line width=1pt, mark options={solid, line width = 0.5pt, fill=white}]table[x=time,y=mean_acc_simple_mean_schII] {data_final_version/mean_acc_uniformdata_M200_epsilon1.000000_RR.txt.100};
\addlegendentry{PM-I (MoM, RR)};

\addplot [color=blue,solid,line width=1pt, mark options={solid, line width = 0.5pt, fill=white}]table[x=time,y=mean_acc_last_schII] {data_final_version/mean_acc_uniformdata_M200_epsilon1.000000_RR.txt.100};
\addlegendentry{PM-I (non-MoM, RR)};


\addplot [color=flamingopink,solid,line width=1pt, mark options={solid, line width = 0.5pt, fill=white}]table[x=time,y=mean_acc_simple_mean_schIV-Hybrid] {data_final_version/mean_acc_uniformdata_M200_epsilon1.000000_RR.txt.100};
\addlegendentry{PM-II (MoM, RR)};

\addplot [color=lightblue,solid,line width=1pt, mark options={solid, line width = 0.5pt, fill=white}]table[x=time,y=mean_acc_last_schIV-Hybrid] {data_final_version/mean_acc_uniformdata_M200_epsilon1.000000_RR.txt.100};
\addlegendentry{PM-II (non-MoM, RR)};

\addplot [color=black,solid,line width=1pt, mark options={solid, line width = 0.5pt, fill=white}]table[x=time,y=mean_acc_simple_mean_schIV-Hybrid] {data_final_version/mean_acc_uniformdata_sliding_window_MOM_M200_epsilon1.000000_RR.txt.100};
\addlegendentry{PM-II (wMoM, RR)};


\addplot [color=darkyellow,dashed,line width=1pt, mark options={solid, line width = 0.5pt, fill=white}]table[x=time,y=error] {data_final_version/ideal_curve_M200.txt.100};
\addlegendentry{Ideal};

\end{axis}
\end{tikzpicture}%

		}
		\hfill
		\subfloat{




\begin{tikzpicture}
\begin{axis}[%
width=\mywidth,
height=\myheight,
xmin=1.0,
xmax=30000,
xtick={1,5000,10000,15000,20000,25000,30000},
xlabel={Time $t$},
xticklabel style = {/pgf/number format/fixed, /pgf/number format/precision=2},
xlabel style={
	yshift=0.5ex,
	name=label,
	font=\small},
grid style={gray,opacity=0.5,dotted},
xmajorgrids,
ymajorgrids,
yminorgrids,
ymode=log,
ymin=1e-6,
ymax=1e-3,
ytick={1e-7,1e-6,1e-5,1e-4,1e-3},
legend style={font=\tiny},
ylabel style={
	yshift=-1.0ex,
	name=label,
	font=\small},
axis background/.style={fill=white},
legend cell align=left,
legend style={at={(axis cs: 30000,0.001)},anchor=north east,nodes={scale=0.75, transform shape}},
]


\addplot [color=darkgreen,dashed,line width=1pt, mark options={solid, line width = 0.5pt, fill=white}]table[x=time,y=acc_local] {data_final_version/local_curve.txt.100};
\addlegendentry{Local};


\addplot [color=red,solid,line width=1pt, mark options={solid, line width = 0.5pt, fill=white}]table[x=time,y=mean_acc_simple_mean_schII] {data_final_version/mean_acc_uniformdata_M30_epsilon1.000000_RR.txt.100};
\addlegendentry{PM-I (MoM, RR)};

\addplot [color=blue,solid,line width=1pt, mark options={solid, line width = 0.5pt, fill=white}]table[x=time,y=mean_acc_last_schII] {data_final_version/mean_acc_uniformdata_M30_epsilon1.000000_RR.txt.100};
\addlegendentry{PM-I (non-MoM, RR)};


\addplot [color=flamingopink,solid,line width=1pt, mark options={solid, line width = 0.5pt, fill=white}]table[x=time,y=mean_acc_simple_mean_schIV-Hybrid] {data_final_version/mean_acc_uniformdata_M30_epsilon1.000000_RR.txt.100};
\addlegendentry{PM-II (MoM, RR)};

\addplot [color=lightblue,solid,line width=1pt, mark options={solid, line width = 0.5pt, fill=white}]table[x=time,y=mean_acc_last_schIV-Hybrid] {data_final_version/mean_acc_uniformdata_M30_epsilon1.000000_RR.txt.100};
\addlegendentry{PM-II (non-MoM, RR)};

\addplot [color=black,solid,line width=1pt, mark options={solid, line width = 0.5pt, fill=white}]table[x=time,y=mean_acc_simple_mean_schIV-Hybrid] {data_final_version/mean_acc_uniformdata_sliding_window_MOM_M30_epsilon1.000000_RR.txt.100};
\addlegendentry{PM-II (wMoM, RR)};


\addplot [color=darkyellow,dashed,line width=1pt, mark options={solid, line width = 0.5pt, fill=white}]table[x=time,y=error] {data_final_version/ideal_curve_M30.txt.100};
\addlegendentry{Ideal};

\end{axis}
\end{tikzpicture}%

		}
	\vspace{-2ex}
		\caption{Average mean squared error of \cref{alg:ss} with $\sigma=\nicefrac{1}{2}$. The left plot shows simulation results with RR and rRR scheduling for the case of $\numa=200$  agents forming  three classes, with an \emph{overall} privacy level of $\epsilon=1$  with $\delta=10^{-6}$. The average of $20$ simulation runs is presented. The middle and right plots show the corresponding  (analytical; see Proposition~\ref{prop:2})  performance with an oracle class estimator and with RR scheduling for  $\numa=200$ and $30$ agents,  respectively. %
			The curves are for uniform data and $L = \sigma \sqrt{3}$.}
		\label{fig:1}
		\vspace{-3ex}
\end{figure*}}{\begin{figure*}[h!]
        \centering
 		\subfloat{
			\input{figures_final_version/MSE_Simulation_M200_epsilon1.000000.tikz}
		}
		\hfill
		\subfloat{
			\input{figures_final_version/MSE_Oracle_RR_M200_epsilon1.000000_noylabel.tikz}
		}
		\hfill
		\subfloat{
			\input{figures_final_version/MSE_Oracle_RR_M30_epsilon5.000000_noylabel.tikz}        }
 \caption{Average mean squared error of \cref{alg:ss} with $\sigma=\nicefrac{1}{2}$. Plots (from left): 1) simulation results with RR and rRR scheduling for the case of $\numa=200$  agents forming  three classes using $\epsilon=1$  and $\delta=10^{-6}$ {\bf TODO: Not updated}. 2) The corresponding oracle performance  with RR scheduling for the case of $\numa=200$ and $\numa=30$,  respectively, for the middle and right plot. %
 The curves are for uniform data and $L = \sigma \sqrt{3}$.}
	 \label{fig:1}
	\vspace{-3ex}
\end{figure*}}}

\subsection{Convergence of \cref{alg:ss} }  \label{sec:convergence}

The following theorem shows that the average mean squared error from  \cref{alg:ss} converges to zero faster than a pure local approach as $t \to \infty$, for the case of known data variance.

\begin{theorem} \label{thm:1}
Let $\nicefrac{1}{\theta_t} = \mathrm e^{o(t)}$ and $\nicefrac{1}{\theta_t} \to \infty$ as $t \to \infty$, and assume that $|\mathcal C_a| \ge 2$ for every $a$. 
For any  distributions $\mathcal{D}_a$, $a \in [\numa]$, with a known variance (and bounded support), for both PM-I and PM-II and with both non-MoM and MoM weights, \cref{alg:ss} with RR scheduling and either Gaussian or Laplace DP noise 
produces $\mu_a^{(t)}$ that is asymptotically unbiased and strictly better than the local approach: %
\begin{align*}
\frac{1}{\numa} \sum_{a \in [\numa]} \EE{\left(\mu_a^{(t)} - \mu_a\right)^2} < \frac{1}{\numa t}\sum_{a \in [\numa]}\sigma_a^2,
\end{align*}
for all large enough $t$.
\end{theorem}

\begin{IEEEproof}
See Appendix~\ref{app:proof-main-theorem}.
\end{IEEEproof}

Note that Theorem~\ref{thm:1} is applicable to a wider class of  distributions $\mathcal{D}_a$, beyond those with a bounded support, satisfying Bernstein's condition. However, a finite DP  guarantee requires a distribution $\mathcal{D}_a$ with bounded support as otherwise the additive DP noise would have infinite variance.

\subsection{Unknown Data Variance Case}
There are two main difficulties in extending Theorem~\ref{thm:1} to the case of unknown data variance. The first being that the $\alpha$'s in \eqref{eq:alpha_unknown_var} become RVs and hence it is unclear whether $\mu_a^{(t)}$ is an unbiased estimator for $\mu_a$, even asymptotically for large $t$, as noted in Remark~\ref{rem:1}. The second difficulty arises from the test statistic  
$\nicefrac{\left| \bar{X}_a^{(t)} - T_{b \to a}  \right|}{\sqrt{\frac{V_a^{(t)}}{t} + \hatVar{T_{b \to a}}}}$ used in \cref{sec:unknown_data_variance} 
as it is hard to say much about its distribution. In particular, when the data distributions and the DP noise satisfy Bernstein's condition, it is unclear if the test statistic also satisfies the condition due to the division.

\ifthenelse{\boolean{THESIS_VERSION}}{
\ifthenelse{\boolean{QUICK_PLOTS}}
{\begin{figure*}[h!]
		\centering
		\subfloat{




\begin{tikzpicture}
\begin{axis}[%
width=\mywidth,
height=\myheight,
xmin=1.0,
xmax=30000,
xtick={1,5000,10000,15000,20000,25000,30000},
xlabel={Time $t$},
xticklabel style = {/pgf/number format/fixed, /pgf/number format/precision=2},
xlabel style={
	yshift=0.5ex,
	name=label,
	font=\small},
grid style={gray,opacity=0.5,dotted},
xmajorgrids,
ymajorgrids,
yminorgrids,
ymode=log,
ymin=1e-7,
ymax=1e-3,
ytick={1e-7,1e-6,1e-5,1e-4,1e-3},
ylabel={Average mean squared error},
legend style={font=\tiny},
ylabel style={
	yshift=-1.0ex,
	name=label,
	font=\small},
axis background/.style={fill=white},
legend cell align=left,
legend style={at={(axis cs: 30000,0.001)},anchor=north east,nodes={scale=0.75, transform shape}},
]


\addplot [color=darkgreen,dashed,line width=1pt, mark options={solid, line width = 0.5pt, fill=white}]table[x=time,y=acc_local] {data_final_version/local_curve.txt.100};
\addlegendentry{Local};




\addplot [color=red,solid,line width=1pt, mark options={solid, line width = 0.5pt, fill=white}]table[x=time,y=MSE] {data_final_version/Mark_pm1_MoM_RR_Hypo_eps1.000000_M200_20runs_together.csv.100};
\addlegendentry{PM-I (MoM, RR)};

\addplot [color=red,dotted,line width=1pt, mark options={solid, line width = 0.5pt, fill=white}]table[x=time,y=MSE] {data_final_version/Mark_pm1_MoM_restRR_Hypo_eps1.000000_M200_20runs_together.csv.100};
\addlegendentry{PM-I (MoM, rRR)}; 




\addplot [color=blue,solid,line width=1pt, mark options={solid, line width = 0.5pt, fill=white}]table[x=time,y=MSE] 
{data_final_version/Mark_pm1_nonMoM_RR_Hypo_eps1.000000_M200_20runs_together.csv.100};
\addlegendentry{PM-I (non-MoM, RR)};

\addplot [color=blue,dotted,line width=1pt, mark options={solid, line width = 0.5pt, fill=white}]table[x=time,y=MSE]
{data_final_version/Mark_pm1_nonMoM_restRR_Hypo_eps1.000000_M200_20runs_together.csv.100}; 
\addlegendentry{PM-I (non-MoM, rRR)}; 




\addplot [color=lightblue,solid,line width=1pt, mark options={solid, line width = 0.5pt, fill=white}]table[x=time,y=MSE] 
{data_final_version/Mark_pm2_nonMoM_RR_Hypo_eps1.000000_M200_20runs_together.csv.100};
\addlegendentry{PM-II (non-MoM, RR)};

\addplot [color=lightblue,dotted,line width=1pt, mark options={solid, line width = 0.5pt, fill=white}]table[x=time,y=MSE] 
{data_final_version/Mark_pm2_nonMoM_restRR_Hypo_eps1.000000_M200_20runs_together.csv.100};
\addlegendentry{PM-II (non-MoM, rRR)};


\addplot [color=darkyellow,dashed,line width=1pt, mark options={solid, line width = 0.5pt, fill=white}]table[x=time,y=error] {data_final_version/ideal_curve_M200.txt.100};
\addlegendentry{Ideal};

\end{axis}
\end{tikzpicture}%

		}
		\hfill
		\subfloat{




\begin{tikzpicture}
\begin{axis}[%
width=\mywidth,
height=\myheight,
xmin=1.0,
xmax=30000,
xtick={1,5000,10000,15000,20000,25000,30000},
xlabel={Time $t$},
xticklabel style = {/pgf/number format/fixed, /pgf/number format/precision=2},
xlabel style={
	yshift=0.5ex,
	name=label,
	font=\small},
grid style={gray,opacity=0.5,dotted},
xmajorgrids,
ymajorgrids,
yminorgrids,
ymode=log,
ymin=1e-7,
ymax=1e-3,
ytick={1e-7,1e-6,1e-5,1e-4,1e-3},
legend style={font=\tiny},
ylabel style={
	yshift=-1.0ex,
	name=label,
	font=\small},
axis background/.style={fill=white},
legend cell align=left,
legend style={at={(axis cs: 30000,0.001)},anchor=north east,nodes={scale=0.75, transform shape}},
]


\addplot [color=darkgreen,dashed,line width=1pt, mark options={solid, line width = 0.5pt, fill=white}]table[x=time,y=acc_local] {data_final_version/local_curve.txt.100};
\addlegendentry{Local};


\addplot [color=red,solid,line width=1pt, mark options={solid, line width = 0.5pt, fill=white}]table[x=time,y=mean_acc_simple_mean_schII] {data_final_version/mean_acc_uniformdata_M200_epsilon1.000000_RR.txt.100};
\addlegendentry{PM-I (MoM, RR)};

\addplot [color=blue,solid,line width=1pt, mark options={solid, line width = 0.5pt, fill=white}]table[x=time,y=mean_acc_last_schII] {data_final_version/mean_acc_uniformdata_M200_epsilon1.000000_RR.txt.100};
\addlegendentry{PM-I (non-MoM, RR)};


\addplot [color=flamingopink,solid,line width=1pt, mark options={solid, line width = 0.5pt, fill=white}]table[x=time,y=mean_acc_simple_mean_schIV-Hybrid] {data_final_version/mean_acc_uniformdata_M200_epsilon1.000000_RR.txt.100};
\addlegendentry{PM-II (MoM, RR)};

\addplot [color=lightblue,solid,line width=1pt, mark options={solid, line width = 0.5pt, fill=white}]table[x=time,y=mean_acc_last_schIV-Hybrid] {data_final_version/mean_acc_uniformdata_M200_epsilon1.000000_RR.txt.100};
\addlegendentry{PM-II (non-MoM, RR)};

\addplot [color=black,solid,line width=1pt, mark options={solid, line width = 0.5pt, fill=white}]table[x=time,y=mean_acc_simple_mean_schIV-Hybrid] {data_final_version/mean_acc_uniformdata_sliding_window_MOM_M200_epsilon1.000000_RR.txt.100};
\addlegendentry{PM-II (wMoM, RR)};


\addplot [color=darkyellow,dashed,line width=1pt, mark options={solid, line width = 0.5pt, fill=white}]table[x=time,y=error] {data_final_version/ideal_curve_M200.txt.100};
\addlegendentry{Ideal};

\end{axis}
\end{tikzpicture}%

		}
		\hfill
		\subfloat{




\begin{tikzpicture}
\begin{axis}[%
width=\mywidth,
height=\myheight,
xmin=1.0,
xmax=30000,
xtick={1,5000,10000,15000,20000,25000,30000},
xlabel={Time $t$},
xticklabel style = {/pgf/number format/fixed, /pgf/number format/precision=2},
xlabel style={
	yshift=0.5ex,
	name=label,
	font=\small},
grid style={gray,opacity=0.5,dotted},
xmajorgrids,
ymajorgrids,
yminorgrids,
ymode=log,
ymin=1e-6,
ymax=1e-3,
ytick={1e-7,1e-6,1e-5,1e-4,1e-3},
legend style={font=\tiny},
ylabel style={
	yshift=-1.0ex,
	name=label,
	font=\small},
axis background/.style={fill=white},
legend cell align=left,
legend style={at={(axis cs: 30000,0.001)},anchor=north east,nodes={scale=0.75, transform shape}},
]


\addplot [color=darkgreen,dashed,line width=1pt, mark options={solid, line width = 0.5pt, fill=white}]table[x=time,y=acc_local] {data_final_version/local_curve.txt.100};
\addlegendentry{Local};


\addplot [color=red,solid,line width=1pt, mark options={solid, line width = 0.5pt, fill=white}]table[x=time,y=mean_acc_simple_mean_schII] {data_final_version/mean_acc_uniformdata_M30_epsilon1.000000_RR.txt.100};
\addlegendentry{PM-I (MoM, RR)};

\addplot [color=blue,solid,line width=1pt, mark options={solid, line width = 0.5pt, fill=white}]table[x=time,y=mean_acc_last_schII] {data_final_version/mean_acc_uniformdata_M30_epsilon1.000000_RR.txt.100};
\addlegendentry{PM-I (non-MoM, RR)};


\addplot [color=flamingopink,solid,line width=1pt, mark options={solid, line width = 0.5pt, fill=white}]table[x=time,y=mean_acc_simple_mean_schIV-Hybrid] {data_final_version/mean_acc_uniformdata_M30_epsilon1.000000_RR.txt.100};
\addlegendentry{PM-II (MoM, RR)};

\addplot [color=lightblue,solid,line width=1pt, mark options={solid, line width = 0.5pt, fill=white}]table[x=time,y=mean_acc_last_schIV-Hybrid] {data_final_version/mean_acc_uniformdata_M30_epsilon1.000000_RR.txt.100};
\addlegendentry{PM-II (non-MoM, RR)};

\addplot [color=black,solid,line width=1pt, mark options={solid, line width = 0.5pt, fill=white}]table[x=time,y=mean_acc_simple_mean_schIV-Hybrid] {data_final_version/mean_acc_uniformdata_sliding_window_MOM_M30_epsilon1.000000_RR.txt.100};
\addlegendentry{PM-II (wMoM, RR)};


\addplot [color=darkyellow,dashed,line width=1pt, mark options={solid, line width = 0.5pt, fill=white}]table[x=time,y=error] {data_final_version/ideal_curve_M30.txt.100};
\addlegendentry{Ideal};

\end{axis}
\end{tikzpicture}%

		}
	\vspace{-2ex}
		\caption{Average mean squared error of \cref{alg:ss} with $\sigma=\nicefrac{1}{2}$. The left plot shows simulation results with RR and rRR scheduling for the case of $\numa=200$  agents forming  three classes, with an \emph{overall} privacy level of $\epsilon=1$  with $\delta=10^{-6}$. The average of $20$ simulation runs is presented. The middle and right plots show the corresponding  (analytical; see Proposition~\ref{prop:2})  performance with an oracle class estimator and with RR scheduling for  $\numa=200$ and $30$ agents,  respectively. %
			The curves are for uniform data and $L = \sigma \sqrt{3}$.}
		\label{fig:1}
		\vspace{-3ex}
\end{figure*}}{\begin{figure*}[h!]
        \centering
 		\subfloat{
			\input{figures_final_version/MSE_Simulation_M200_epsilon1.000000.tikz}
		}
		\hfill
		\subfloat{
			\input{figures_final_version/MSE_Oracle_RR_M200_epsilon1.000000_noylabel.tikz}
		}
		\hfill
		\subfloat{
			\input{figures_final_version/MSE_Oracle_RR_M30_epsilon5.000000_noylabel.tikz}        }
 \caption{Average mean squared error of \cref{alg:ss} with $\sigma=\nicefrac{1}{2}$. Plots (from left): 1) simulation results with RR and rRR scheduling for the case of $\numa=200$  agents forming  three classes using $\epsilon=1$  and $\delta=10^{-6}$ {\bf TODO: Not updated}. 2) The corresponding oracle performance  with RR scheduling for the case of $\numa=200$ and $\numa=30$,  respectively, for the middle and right plot. %
 The curves are for uniform data and $L = \sigma \sqrt{3}$.}
	 \label{fig:1}
	\vspace{-3ex}
\end{figure*}}}{}

\section{Numerical Results} \label{sec:numerical_results}

We consider the case of $\numa$  agents forming  three classes. The  agents are placed uniformly at random within the  classes, giving roughly balanced class sizes.  The agents'  data distributions are uniform (to model tabular data) on a range of size $2L=2 \sigma  \sqrt{3}$ (giving a standard deviation of $\sigma$), with $\sigma=\nicefrac{1}{2}$, but with different class-dependent means;  $\nicefrac{1}{5}$, $\nicefrac{2}{5}$, and $\nicefrac{4}{5}$, respectively, for the three classes. In order to have a fair comparison between PM-I and PM-II,  $(\epsilon,\delta)$ of PM-II is scaled by $\lfloor \log_2(t_{\max}) \rfloor +1$ so that both mechanisms provide the same  privacy level. For the decision rule, we use $\theta_t = \nicefrac{0.05}{\ln(t+1)}$. 

\ifthenelse{\boolean{THESIS_VERSION}}{}{
\begin{figure*}[h!]
		\centering
		\subfloat{




\begin{tikzpicture}

\begin{axis}[%
width=\mywidthtwo,
height=\myheighttwo,
xmin=1.0,
xmax=30000,
xtick={1,5000,10000,15000,20000,25000,30000},
xlabel={Time $t$},
xticklabel style = {/pgf/number format/fixed, /pgf/number format/precision=2},
xlabel style={
	yshift=0.5ex,
	name=label,
	font=\small},
grid style={gray,opacity=0.5,dotted},
xmajorgrids,
ymajorgrids,
yminorgrids,
ymode=log,
ymin=1e-7,
ymax=1e-3,
ytick={1e-7,1e-6,1e-5,1e-4,1e-3},
ylabel={Average mean squared error},
legend style={font=\tiny},
ylabel style={
	yshift=-1.0ex,
	name=label,
	font=\small},
axis background/.style={fill=white},
legend cell align=left,
legend style={at={(axis cs: 30000,0.001)},anchor=north east,nodes={scale=0.75, transform shape}},
]


\addplot [color=darkgreen,dashed,line width=1pt, mark options={solid, line width = 0.5pt, fill=white}]table[x=time,y=acc_local] {data_final_version/local_curve.txt.100};
\addlegendentry{Local};




\addplot [color=red,solid,line width=1pt, mark options={solid, line width = 0.5pt, fill=white}]table[x=time,y=MSE] {data_final_version/Mark_pm1_MoM_RR_Hypo_eps1.000000_M200_20runs_together.csv.100};
\addlegendentry{PM-I (MoM)};





\addplot [color=blue,solid,line width=1pt, mark options={solid, line width = 0.5pt, fill=white}]table[x=time,y=MSE] 
{data_final_version/Mark_pm1_nonMoM_RR_Hypo_eps1.000000_M200_20runs_together.csv.100};
\addlegendentry{PM-I (non-MoM)};



\addplot [color=blue,dotted,line width=1pt, mark options={solid, line width = 0.5pt, fill=white}]table[x=time,y=Scheme1esti]
{data_tifs/Mark_pm1_estiVar1_nonMoM_RR_M200_20runs.csv.100};
\addlegendentry{PM-I (non-MoM, Var-Est-1) }; 



\addplot [color=blue,dashed,line width=1pt, mark options={solid, line width = 0.5pt, fill=white}]table[x=time,y=Scheme2estiWscale]
{data_tifs/Mark_pm1_estiVar2ScaleCorrected_nonMoM_RR_M200_20runs.csv.100};
\addlegendentry{PM-I (non-MoM, Var-Est-2, Bay.)}; 

\addplot [color=orange,dashed,line width=1pt, mark options={solid, line width = 0.5pt, fill=white}]table[x=time,y=Scheme2estiWOscale]
{data_tifs/Mark_pm1_estiVar2Scale_nonMoM_RR_M200_20runs.csv.100};
\addlegendentry{PM-I (non-MoM, Var-Est-2)};







\addplot [color=darkyellow,dashed,line width=1pt, mark options={solid, line width = 0.5pt, fill=white}]table[x=time,y=error] {data_final_version/ideal_curve_M200.txt.100};
\addlegendentry{Ideal};

\end{axis}


\end{tikzpicture}%

		}
		\subfloat{




\begin{tikzpicture}

\begin{axis}[%
width=\mywidthtwo,
height=\myheighttwo,
xmin=1.0,
xmax=2000,
xtick={1,200,500,1000,1500,2000},
xlabel={Time $t$},
xticklabel style = {/pgf/number format/fixed, /pgf/number format/precision=2},
xlabel style={
	yshift=0.5ex,
	name=label,
	font=\small},
grid style={gray,opacity=0.5,dotted},
xmajorgrids,
ymajorgrids,
yminorgrids,
ymode=log,
ymin=1e-5,
ymax=1e-1,
ytick={1e-5,1e-4,1e-3,1e-2,1e-1},
legend style={font=\tiny},
ylabel style={
	yshift=-1.0ex,
	name=label,
	font=\small},
axis background/.style={fill=white},
legend cell align=left,
legend style={at={(axis cs: 30000,0.001)},anchor=north east,nodes={scale=0.75, transform shape}},
]


\addplot [color=darkgreen,dashed,line width=1pt, mark options={solid, line width = 0.5pt, fill=white}]table[x=time,y=acc_local] {data_final_version/local_curve.txt.inset};




\addplot [color=red,solid,line width=1pt, mark options={solid, line width = 0.5pt, fill=white}]table[x=time,y=MSE] {data_final_version/Mark_pm1_MoM_RR_Hypo_eps1.000000_M200_20runs_together.csv.inset};





\addplot [color=blue,solid,line width=1pt, mark options={solid, line width = 0.5pt, fill=white}]table[x=time,y=MSE] 
{data_final_version/Mark_pm1_nonMoM_RR_Hypo_eps1.000000_M200_20runs_together.csv.inset};



\addplot [color=blue,dotted,line width=1pt, mark options={solid, line width = 0.5pt, fill=white}]table[x=time,y=Scheme1esti]
{data_tifs/Mark_pm1_estiVar1_nonMoM_RR_M200_20runs.csv.inset};



\addplot [color=blue,dashed,line width=1pt, mark options={solid, line width = 0.5pt, fill=white}]table[x=time,y=Scheme2estiWscale]
{data_tifs/Mark_pm1_estiVar2ScaleCorrected_nonMoM_RR_M200_20runs.csv.inset};

\addplot [color=orange,dashed,line width=1pt, mark options={solid, line width = 0.5pt, fill=white}]table[x=time,y=Scheme2estiWOscale]
{data_tifs/Mark_pm1_estiVar2Scale_nonMoM_RR_M200_20runs.csv.inset};







\addplot [color=darkyellow,dashed,line width=1pt, mark options={solid, line width = 0.5pt, fill=white}]table[x=time,y=error] {data_final_version/ideal_curve_M200.txt.inset};

\end{axis}


\end{tikzpicture}%

		}
	\vspace{-2ex}
		\caption{(Left plot): Comparisons of the average mean squared error of \cref{alg:ss} with and without data variance estimation for the same scenario as in \cref{fig:1} (left and middle plot) for PM-I with non-MoM weights and RR scheduling. For completeness, the performance of a fully local approach, PM-I with MoM weights, and an ideal scheme is also depicted. The average of $20$ simulation runs is presented. Right plot: The initial part of the performance curves from the left plot up to time $t = 2000$.}
		\label{fig:2}
		\vspace{-3ex}
\end{figure*}}

\subsection{Known Data Variance Case}

We first consider the case when the variances of data are known. 
In \cref{fig:1} (left plot), we show  simulation results for  RR and  rRR scheduling in \cref{alg:ss} (with $t_{\max}=3 \cdot 10^4$) for both PM-I and PM-II with the MoM and non-MoM approach for $\numa=200$ agents. The \emph{overall} privacy level is $\epsilon=1$ with $\delta=10^{-6}$ using the Gaussian DP mechansim. As expected, for PM-I the non-MoM approach outperforms the MoM approach (cf.\ Lemma~\ref{lem:keeping_last_optimal_schemeII}), while there is still some gap  to ideal performance. Moreover, PM-I with non-MoM weights outperforms  PM-II for the range of squared errors shown in the plot. Note that RR  performs better than rRR, which was not the case when privacy is ignored (see \cite[Fig.~1]{AsadiBelletMaillardTommasi22_1}). This can be explained by the fact that sample means are released less often with RR than with rRR (the gaps $t_i-t_{i-1}$ are larger for RR).
The collaborative schemes (ultimately) perform significantly better than a pure local approach, while PM-II with non-MoM weights and rRR scheduling being a notable exception; the reason being that the  decision rule type-I error  increases over time rather than converging to a close-to-zero value. %
In the middle plot, the corresponding (analytical)   performance  with an oracle class estimator and with RR scheduling  given by Proposition~\ref{prop:2} %
is provided which shows qualitatively the same behavior as in the left plot. %
While PM-I performs better than PM-II for the range of squared errors shown, asymptotically (for very low squared errors) the curves will cross (not shown). %
Note that PM-II with wMoM weights smoothens the corresponding curve with non-MoM weights (the oscillations are due to the factor $w_{\mathrm{H}}(\kappa)$ in the variance of $Z_{b \to a}^{(t_\kappa)}$) and shows that having non-MoM weights for PM-II is not optimal. Moreover, PM-II with wMoM weights performs significantly better than a pure MoM approach. In the right plot, we show the corresponding oracle performance for $\numa=30$. %
In contrast to the middle plot, PM-II with wMoM weights shows the best performance for a low squared error. On the other hand, compared to $M=200$, the performance gap to ideal performance is far larger.

\ifthenelse{\boolean{THESIS_VERSION}}{
\begin{figure*}[h!]
		\centering
		\subfloat{




\begin{tikzpicture}

\begin{axis}[%
width=\mywidthtwo,
height=\myheighttwo,
xmin=1.0,
xmax=30000,
xtick={1,5000,10000,15000,20000,25000,30000},
xlabel={Time $t$},
xticklabel style = {/pgf/number format/fixed, /pgf/number format/precision=2},
xlabel style={
	yshift=0.5ex,
	name=label,
	font=\small},
grid style={gray,opacity=0.5,dotted},
xmajorgrids,
ymajorgrids,
yminorgrids,
ymode=log,
ymin=1e-7,
ymax=1e-3,
ytick={1e-7,1e-6,1e-5,1e-4,1e-3},
ylabel={Average mean squared error},
legend style={font=\tiny},
ylabel style={
	yshift=-1.0ex,
	name=label,
	font=\small},
axis background/.style={fill=white},
legend cell align=left,
legend style={at={(axis cs: 30000,0.001)},anchor=north east,nodes={scale=0.75, transform shape}},
]


\addplot [color=darkgreen,dashed,line width=1pt, mark options={solid, line width = 0.5pt, fill=white}]table[x=time,y=acc_local] {data_final_version/local_curve.txt.100};
\addlegendentry{Local};




\addplot [color=red,solid,line width=1pt, mark options={solid, line width = 0.5pt, fill=white}]table[x=time,y=MSE] {data_final_version/Mark_pm1_MoM_RR_Hypo_eps1.000000_M200_20runs_together.csv.100};
\addlegendentry{PM-I (MoM)};





\addplot [color=blue,solid,line width=1pt, mark options={solid, line width = 0.5pt, fill=white}]table[x=time,y=MSE] 
{data_final_version/Mark_pm1_nonMoM_RR_Hypo_eps1.000000_M200_20runs_together.csv.100};
\addlegendentry{PM-I (non-MoM)};



\addplot [color=blue,dotted,line width=1pt, mark options={solid, line width = 0.5pt, fill=white}]table[x=time,y=Scheme1esti]
{data_tifs/Mark_pm1_estiVar1_nonMoM_RR_M200_20runs.csv.100};
\addlegendentry{PM-I (non-MoM, Var-Est-1) }; 



\addplot [color=blue,dashed,line width=1pt, mark options={solid, line width = 0.5pt, fill=white}]table[x=time,y=Scheme2estiWscale]
{data_tifs/Mark_pm1_estiVar2ScaleCorrected_nonMoM_RR_M200_20runs.csv.100};
\addlegendentry{PM-I (non-MoM, Var-Est-2, Bay.)}; 

\addplot [color=orange,dashed,line width=1pt, mark options={solid, line width = 0.5pt, fill=white}]table[x=time,y=Scheme2estiWOscale]
{data_tifs/Mark_pm1_estiVar2Scale_nonMoM_RR_M200_20runs.csv.100};
\addlegendentry{PM-I (non-MoM, Var-Est-2)};







\addplot [color=darkyellow,dashed,line width=1pt, mark options={solid, line width = 0.5pt, fill=white}]table[x=time,y=error] {data_final_version/ideal_curve_M200.txt.100};
\addlegendentry{Ideal};

\end{axis}


\end{tikzpicture}%

		}
		\subfloat{




\begin{tikzpicture}

\begin{axis}[%
width=\mywidthtwo,
height=\myheighttwo,
xmin=1.0,
xmax=2000,
xtick={1,200,500,1000,1500,2000},
xlabel={Time $t$},
xticklabel style = {/pgf/number format/fixed, /pgf/number format/precision=2},
xlabel style={
	yshift=0.5ex,
	name=label,
	font=\small},
grid style={gray,opacity=0.5,dotted},
xmajorgrids,
ymajorgrids,
yminorgrids,
ymode=log,
ymin=1e-5,
ymax=1e-1,
ytick={1e-5,1e-4,1e-3,1e-2,1e-1},
legend style={font=\tiny},
ylabel style={
	yshift=-1.0ex,
	name=label,
	font=\small},
axis background/.style={fill=white},
legend cell align=left,
legend style={at={(axis cs: 30000,0.001)},anchor=north east,nodes={scale=0.75, transform shape}},
]


\addplot [color=darkgreen,dashed,line width=1pt, mark options={solid, line width = 0.5pt, fill=white}]table[x=time,y=acc_local] {data_final_version/local_curve.txt.inset};




\addplot [color=red,solid,line width=1pt, mark options={solid, line width = 0.5pt, fill=white}]table[x=time,y=MSE] {data_final_version/Mark_pm1_MoM_RR_Hypo_eps1.000000_M200_20runs_together.csv.inset};





\addplot [color=blue,solid,line width=1pt, mark options={solid, line width = 0.5pt, fill=white}]table[x=time,y=MSE] 
{data_final_version/Mark_pm1_nonMoM_RR_Hypo_eps1.000000_M200_20runs_together.csv.inset};



\addplot [color=blue,dotted,line width=1pt, mark options={solid, line width = 0.5pt, fill=white}]table[x=time,y=Scheme1esti]
{data_tifs/Mark_pm1_estiVar1_nonMoM_RR_M200_20runs.csv.inset};



\addplot [color=blue,dashed,line width=1pt, mark options={solid, line width = 0.5pt, fill=white}]table[x=time,y=Scheme2estiWscale]
{data_tifs/Mark_pm1_estiVar2ScaleCorrected_nonMoM_RR_M200_20runs.csv.inset};

\addplot [color=orange,dashed,line width=1pt, mark options={solid, line width = 0.5pt, fill=white}]table[x=time,y=Scheme2estiWOscale]
{data_tifs/Mark_pm1_estiVar2Scale_nonMoM_RR_M200_20runs.csv.inset};







\addplot [color=darkyellow,dashed,line width=1pt, mark options={solid, line width = 0.5pt, fill=white}]table[x=time,y=error] {data_final_version/ideal_curve_M200.txt.inset};

\end{axis}


\end{tikzpicture}%

		}
	\vspace{-2ex}
		\caption{(Left plot): Comparisons of the average mean squared error of \cref{alg:ss} with and without data variance estimation for the same scenario as in \cref{fig:1} (left and middle plot) for PM-I with non-MoM weights and RR scheduling. For completeness, the performance of a fully local approach, PM-I with MoM weights, and an ideal scheme is also depicted. The average of $20$ simulation runs is presented. Right plot: The initial part of the performance curves from the left plot up to time $t = 2000$.}
		\label{fig:2}
		\vspace{-3ex}
\end{figure*}}{}

\ifthenelse{\boolean{THESIS_VERSION}}{}{
\begin{figure*}[h!]
		\centering
		\subfloat{




\begin{tikzpicture}

\begin{axis}[%
width=\mywidthtwo,
height=\myheighttwo,
xmin=1.0,
xmax=30000,
xtick={1,5000,10000,15000,20000,25000,30000},
xlabel={Time $t$},
xticklabel style = {/pgf/number format/fixed, /pgf/number format/precision=2},
xlabel style={
	yshift=0.5ex,
	name=label,
	font=\small},
grid style={gray,opacity=0.5,dotted},
xmajorgrids,
ymajorgrids,
yminorgrids,
ymode=log,
ymin=1e-6,
ymax=1e-3,
ytick={1e-7,1e-6,1e-5,1e-4,1e-3},
ylabel={Average mean squared error},
legend style={font=\tiny},
ylabel style={
	yshift=-1.0ex,
	name=label,
	font=\small},
axis background/.style={fill=white},
legend cell align=left,
legend style={at={(axis cs: 30000,0.001)},anchor=north east,nodes={scale=0.75, transform shape}},
]


\addplot [color=darkgreen,dashed,line width=1pt, mark options={solid, line width = 0.5pt, fill=white}]table[x=time,y=acc_local] {data_final_version/local_curve.txt.100};
\addlegendentry{Local};









\addplot [color=blue,solid,line width=1pt, mark options={solid, line width = 0.5pt, fill=white}]table[x=time,y=MSE] 
{data_final_version/Mark_pm1_nonMoM_Laplace_eps1.000000_M15_100runs_together.csv.100};
\addlegendentry{PM-I (non-MoM)};

\addplot [color=lightblue,solid,line width=1pt, mark options={solid, line width = 0.5pt, fill=white}]table[x=time,y=MSE] 
{data_final_version/Mark_pm2_nonMoM_Laplace_eps1.000000_M15_100runs_together.csv.100};
\addlegendentry{PM-II (non-MoM)};



\addplot [color=darkyellow,dashed,line width=1pt, mark options={solid, line width = 0.5pt, fill=white}]table[x=time,y=error] {data_final_version/ideal_curve_M15.txt.100};
\addlegendentry{Ideal};

\addplot [color=blue,solid,line width=1pt, mark options={solid, line width = 0.5pt, fill=white}]table[x=time,y=MSE] 
{data_final_version/Mark_pm1_nonMoM_Gauss_eps1.000000_M15_100runs_together.csv.100};

\addplot [color=lightblue,solid,line width=1pt, mark options={solid, line width = 0.5pt, fill=white}]table[x=time,y=MSE] 
{data_final_version/Mark_pm2_nonMoM_Gauss_eps1.000000_M15_100runs_together.csv.100};

\end{axis}

\ifthenelse{\boolean{THESIS_VERSION}}{
\begin{scope}
\node[ellipse,dashed,minimum height=2.5pt,minimum width=5pt,draw=black, shift={(0.0,0.0)},line width=0.25mm] (l1) at (3.1,2.6) {\tiny \phantom{am am}};
\draw[->,black] (l1.east) to ($ (l1) + (1.0,0.25) $);
\node[] at ($ (l1) + (1.0,0.4) $) {\scriptsize {Gaussian}};
\end{scope}

\begin{scope}[shift={(1.25,-0.85)}]
\node[ellipse,dashed,minimum height=2.5pt,minimum width=5pt,draw=black, shift={(0.0,0.0)},line width=0.25mm] (l1) at (2.95,2.45) {\tiny \phantom{am am}};
\draw[->,black] (l1.north) to ($ (l1) + (0.75,0.75) $);
\node[] at ($ (l1) + (0.5,0.90) $) {\scriptsize {Laplace}};
\end{scope}}{
\begin{scope}
\node[ellipse,dashed,minimum height=2.5pt,minimum width=5pt,draw=black, shift={(0.0,0.0)},line width=0.25mm] (l1) at (2.25,1.75) {\tiny \phantom{am am}};
\draw[->,black] (l1.east) to ($ (l1) + (1.0,0.25) $);
\node[] at ($ (l1) + (1.0,0.4) $) {\scriptsize {Gaussian}};
\end{scope}

\begin{scope}[shift={(1.25,-0.85)}]
\node[ellipse,dashed,minimum height=2.5pt,minimum width=5pt,draw=black, shift={(0.0,0.0)},line width=0.25mm] (l1) at (2.25,1.75) {\tiny \phantom{am am}};
\draw[->,black] (l1.north) to ($ (l1) + (0.75,0.75) $);
\node[] at ($ (l1) + (0.5,0.90) $) {\scriptsize {Laplace}};
\end{scope}}


\end{tikzpicture}%

		}
		\subfloat{




\begin{tikzpicture}
\begin{axis}[%
width=\mywidthtwo,
height=\myheighttwo,
xmin=1.0,
xmax=30000,
xtick={1,5000,10000,15000,20000,25000,30000},
xlabel={Time $t$},
xticklabel style = {/pgf/number format/fixed, /pgf/number format/precision=2},
xlabel style={
	yshift=0.5ex,
	name=label,
	font=\small},
grid style={gray,opacity=0.5,dotted},
xmajorgrids,
ymajorgrids,
yminorgrids,
ymode=log,
ymin=1e-6,
ymax=1e-3,
ytick={1e-7,1e-6,1e-5,1e-4,1e-3},
legend style={font=\tiny},
ylabel style={
	yshift=-1.0ex,
	name=label,
	font=\small},
axis background/.style={fill=white},
legend cell align=left,
legend style={at={(axis cs: 30000,0.001)},anchor=north east,nodes={scale=0.75, transform shape}},
]


\addplot [color=darkgreen,dashed,line width=1pt, mark options={solid, line width = 0.5pt, fill=white}]table[x=time,y=acc_local] {data_final_version/local_curve.txt.100};
\addlegendentry{Local};



\addplot [color=blue,solid,line width=1pt, mark options={solid, line width = 0.5pt, fill=white}]table[x=time,y=mean_acc_last_schII] {data_final_version/mean_acc_uniformdata_M15_Laplace_epsilon1.000000_RR.txt.100};
\addlegendentry{PM-I (non-MoM)};



\addplot [color=lightblue,solid,line width=1pt, mark options={solid, line width = 0.5pt, fill=white}]table[x=time,y=mean_acc_last_schIV-Hybrid] {data_final_version/mean_acc_uniformdata_M15_LaplaceNew_epsilon1.000000_RR.txt.100};
\addlegendentry{PM-II (non-MoM)};

\addplot [color=black,solid,line width=1pt, mark options={solid, line width = 0.5pt, fill=white}]table[x=time,y=mean_acc_simple_mean_schIV-Hybrid] {data_final_version/mean_acc_uniformdata_sliding_window_MOM_M15_LaplaceNew_epsilon1.000000_RR.txt.100};
\addlegendentry{PM-II (wMoM)};


\addplot [color=darkyellow,dashed,line width=1pt, mark options={solid, line width = 0.5pt, fill=white}]table[x=time,y=error] {data_final_version/ideal_curve_M15.txt.100};
\addlegendentry{Ideal};

\end{axis}
\end{tikzpicture}%

		}
	\vspace{-2ex}
		\caption{(Left plot): Comparisons of the average mean squared error of \cref{alg:ss} with Gaussian and Laplace DP noise for the same scenario as in \cref{fig:1}, but for $\numa=15$ agents, for PM-I and PM-II with non-MoM weights and RR scheduling. The average of $100$ simulation runs is presented. For completeness, the performance of a fully local approach and an ideal scheme is also depicted. (Right plot): The corresponding (analytical; see Proposition~\ref{prop:2})  performance with Laplace DP noise and with an oracle class estimator.}
		\label{fig:3}
		\vspace{-3ex}
\end{figure*}}

\subsection{Unknown Data Variance Case}

As illustrated in \cref{fig:1} for $\numa=200$ agents (left and middle plot), PM-I with non-MoM weights and RR scheduling shows the best performance for the considered scenario. In \cref{fig:2} (left plot), we compare the performance of this scheme with and without data variance estimation using the two proposed variance estimation schemes (blue dotted and blue and orange dashed curves). The middle plot shows the initial part of the curves from the left plot until time $t = 2000$. As can be seen from the figure, the variance estimation scheme from \cref{sec:privacy_sample_variance_2}, referred to as Sch-Var-$2$ (blue and orange dashed curves; blue for the Bayesian approach), performs the best and the performance converges to the performance curve for the case when the data variance is known (blue solid curve). Note that the Bayesian approach of \cref{sec:Bayesian_app} performs worse than the non-Bayesian approach in the beginning, while there is a crossing at around $t=600$ (choosing the  Jeffreys prior gives  worse results with a crossing at around $t=850$; not shown). This can be explained by the fact that initially the class estimation is far from perfect and then it turns out that it  advantageous to simply skip some agents $b$ from the computation of the current estimate $\mu_a^{(t)} $ (see \cref{line:weights} in \cref{alg:ss}) by setting the estimate of the corresponding data variance $V_{b \to a}$ to infinity (which gives $\alpha^{(t)}_{b \to a}=0$) rather than estimating the data variance more accurately (which gives  $\alpha^{(t)}_{b \to a}> 0$) and include that agent $b$. Note that the first variance estimation scheme from \cref{sec:privacy_sample_variance_1}, referred to as Sch-Var-$1$, displays an asymptotic  performance  loss which can be explained by the fact that  the privacy budget of $\epsilon=1$ has to be split equally between the releases of the current sample mean and the sample variance in order to  meet an overall privacy budget of $\epsilon=1$. On the other hand, data variance estimation of other agents can start immediately at time $t=2$, while for Sch-Var-$2$ data variance estimation of other agents can only start at time $t=M$.

\ifthenelse{\boolean{THESIS_VERSION}}{
\begin{figure*}[h!]
		\centering
		\subfloat{




\begin{tikzpicture}

\begin{axis}[%
width=\mywidthtwo,
height=\myheighttwo,
xmin=1.0,
xmax=30000,
xtick={1,5000,10000,15000,20000,25000,30000},
xlabel={Time $t$},
xticklabel style = {/pgf/number format/fixed, /pgf/number format/precision=2},
xlabel style={
	yshift=0.5ex,
	name=label,
	font=\small},
grid style={gray,opacity=0.5,dotted},
xmajorgrids,
ymajorgrids,
yminorgrids,
ymode=log,
ymin=1e-6,
ymax=1e-3,
ytick={1e-7,1e-6,1e-5,1e-4,1e-3},
ylabel={Average mean squared error},
legend style={font=\tiny},
ylabel style={
	yshift=-1.0ex,
	name=label,
	font=\small},
axis background/.style={fill=white},
legend cell align=left,
legend style={at={(axis cs: 30000,0.001)},anchor=north east,nodes={scale=0.75, transform shape}},
]


\addplot [color=darkgreen,dashed,line width=1pt, mark options={solid, line width = 0.5pt, fill=white}]table[x=time,y=acc_local] {data_final_version/local_curve.txt.100};
\addlegendentry{Local};









\addplot [color=blue,solid,line width=1pt, mark options={solid, line width = 0.5pt, fill=white}]table[x=time,y=MSE] 
{data_final_version/Mark_pm1_nonMoM_Laplace_eps1.000000_M15_100runs_together.csv.100};
\addlegendentry{PM-I (non-MoM)};

\addplot [color=lightblue,solid,line width=1pt, mark options={solid, line width = 0.5pt, fill=white}]table[x=time,y=MSE] 
{data_final_version/Mark_pm2_nonMoM_Laplace_eps1.000000_M15_100runs_together.csv.100};
\addlegendentry{PM-II (non-MoM)};



\addplot [color=darkyellow,dashed,line width=1pt, mark options={solid, line width = 0.5pt, fill=white}]table[x=time,y=error] {data_final_version/ideal_curve_M15.txt.100};
\addlegendentry{Ideal};

\addplot [color=blue,solid,line width=1pt, mark options={solid, line width = 0.5pt, fill=white}]table[x=time,y=MSE] 
{data_final_version/Mark_pm1_nonMoM_Gauss_eps1.000000_M15_100runs_together.csv.100};

\addplot [color=lightblue,solid,line width=1pt, mark options={solid, line width = 0.5pt, fill=white}]table[x=time,y=MSE] 
{data_final_version/Mark_pm2_nonMoM_Gauss_eps1.000000_M15_100runs_together.csv.100};

\end{axis}

\ifthenelse{\boolean{THESIS_VERSION}}{
\begin{scope}
\node[ellipse,dashed,minimum height=2.5pt,minimum width=5pt,draw=black, shift={(0.0,0.0)},line width=0.25mm] (l1) at (3.1,2.6) {\tiny \phantom{am am}};
\draw[->,black] (l1.east) to ($ (l1) + (1.0,0.25) $);
\node[] at ($ (l1) + (1.0,0.4) $) {\scriptsize {Gaussian}};
\end{scope}

\begin{scope}[shift={(1.25,-0.85)}]
\node[ellipse,dashed,minimum height=2.5pt,minimum width=5pt,draw=black, shift={(0.0,0.0)},line width=0.25mm] (l1) at (2.95,2.45) {\tiny \phantom{am am}};
\draw[->,black] (l1.north) to ($ (l1) + (0.75,0.75) $);
\node[] at ($ (l1) + (0.5,0.90) $) {\scriptsize {Laplace}};
\end{scope}}{
\begin{scope}
\node[ellipse,dashed,minimum height=2.5pt,minimum width=5pt,draw=black, shift={(0.0,0.0)},line width=0.25mm] (l1) at (2.25,1.75) {\tiny \phantom{am am}};
\draw[->,black] (l1.east) to ($ (l1) + (1.0,0.25) $);
\node[] at ($ (l1) + (1.0,0.4) $) {\scriptsize {Gaussian}};
\end{scope}

\begin{scope}[shift={(1.25,-0.85)}]
\node[ellipse,dashed,minimum height=2.5pt,minimum width=5pt,draw=black, shift={(0.0,0.0)},line width=0.25mm] (l1) at (2.25,1.75) {\tiny \phantom{am am}};
\draw[->,black] (l1.north) to ($ (l1) + (0.75,0.75) $);
\node[] at ($ (l1) + (0.5,0.90) $) {\scriptsize {Laplace}};
\end{scope}}


\end{tikzpicture}%

		}
		\subfloat{




\begin{tikzpicture}
\begin{axis}[%
width=\mywidthtwo,
height=\myheighttwo,
xmin=1.0,
xmax=30000,
xtick={1,5000,10000,15000,20000,25000,30000},
xlabel={Time $t$},
xticklabel style = {/pgf/number format/fixed, /pgf/number format/precision=2},
xlabel style={
	yshift=0.5ex,
	name=label,
	font=\small},
grid style={gray,opacity=0.5,dotted},
xmajorgrids,
ymajorgrids,
yminorgrids,
ymode=log,
ymin=1e-6,
ymax=1e-3,
ytick={1e-7,1e-6,1e-5,1e-4,1e-3},
legend style={font=\tiny},
ylabel style={
	yshift=-1.0ex,
	name=label,
	font=\small},
axis background/.style={fill=white},
legend cell align=left,
legend style={at={(axis cs: 30000,0.001)},anchor=north east,nodes={scale=0.75, transform shape}},
]


\addplot [color=darkgreen,dashed,line width=1pt, mark options={solid, line width = 0.5pt, fill=white}]table[x=time,y=acc_local] {data_final_version/local_curve.txt.100};
\addlegendentry{Local};



\addplot [color=blue,solid,line width=1pt, mark options={solid, line width = 0.5pt, fill=white}]table[x=time,y=mean_acc_last_schII] {data_final_version/mean_acc_uniformdata_M15_Laplace_epsilon1.000000_RR.txt.100};
\addlegendentry{PM-I (non-MoM)};



\addplot [color=lightblue,solid,line width=1pt, mark options={solid, line width = 0.5pt, fill=white}]table[x=time,y=mean_acc_last_schIV-Hybrid] {data_final_version/mean_acc_uniformdata_M15_LaplaceNew_epsilon1.000000_RR.txt.100};
\addlegendentry{PM-II (non-MoM)};

\addplot [color=black,solid,line width=1pt, mark options={solid, line width = 0.5pt, fill=white}]table[x=time,y=mean_acc_simple_mean_schIV-Hybrid] {data_final_version/mean_acc_uniformdata_sliding_window_MOM_M15_LaplaceNew_epsilon1.000000_RR.txt.100};
\addlegendentry{PM-II (wMoM)};


\addplot [color=darkyellow,dashed,line width=1pt, mark options={solid, line width = 0.5pt, fill=white}]table[x=time,y=error] {data_final_version/ideal_curve_M15.txt.100};
\addlegendentry{Ideal};

\end{axis}
\end{tikzpicture}%

		}
	\vspace{-2ex}
		\caption{(Left plot): Comparisons of the average mean squared error of \cref{alg:ss} with Gaussian and Laplace DP noise for the same scenario as in \cref{fig:1}, but for $\numa=15$ agents, for PM-I and PM-II with non-MoM weights and RR scheduling. The average of $100$ simulation runs is presented. For completeness, the performance of a fully local approach and an ideal scheme is also depicted. (Right plot): The corresponding (analytical; see Proposition~\ref{prop:2})  performance with Laplace DP noise and with an oracle class estimator.}
		\label{fig:3}
		\vspace{-3ex}
\end{figure*}}{}
\subsection{Laplace Noise}

Finally, we compare the performance with Gaussian  and Laplace DP noise. Note that the Laplace mechanism allows for a privacy level $\epsilon > 1$ and also adds noise with lower variance when $\ln(1.25/ \delta) \geq 1$, i.e., when $\delta \leq 0.4598$ (see Lemmas~\ref{lem:DP} and \ref{lem:DP_Laplace}), which explains the lower mean squared error that we observe with Laplace noise in \cref{fig:3} (left plot). In more details, in  \cref{fig:3} (left plot), we  compare the average mean squared error of \cref{alg:ss} with Gaussian (upper set of light blue and blue curves) and Laplace (lower set of light blue and blue curves)  DP noise for the same scenario as in \cref{fig:1}, but for $\numa=15$ agents, for both PM-I and PM-II with non-MoM weights (as shown in \cref{fig:1}, non-MoM weights gave better performance than MoM  weights) and RR scheduling. Interestingly, we can observe that for $\numa=15$ agents, PM-II outperforms PM-I for low error values. The corresponding oracle performance (with an oracle class estimator) based on   Proposition~\ref{prop:2}  with Laplace DP noise is shown in the right plot. As the mean squared error decreases, the simulation results from the left plot align almost perfectly with the corresponding curves in the right plot.

\section{Conclusion}
We presented a private collaborative algorithm for personalized  mean estimation in an online setting where agents continuously receive  samples according to arbitrary unknown distributions. %
An approach based on hypothesis testing and DP was proposed and proven analytically, under general conditions, to converge faster compared to a pure local approach where the agents do not share their data. Moreover, analytical performance curves for the case with an oracle class estimator, i.e., the class structure of the agents, where agents receiving data from distributions with the same mean are considered to be in the same class, is known were provided. In the case that the variances of the agents' data distributions are  unknown, two variance estimation schemes were proposed. Extensive numerical results were presented showing that for a considered scenario, the best scheme (with and without variance estimation), performs comparably to an ideal scheme where all data is public. %
The communication cost was not considered in this work and its implication on the accuracy/privacy will be studied next as future work.

\appendices

\section{Variances of $T_{b \to a}$}\label{app:Tba_variances}
Here, we derive explicit formulas for the variance of $T_{b \to a}$ for  PM-I and PM-II, both for MoM and non-MoM weights.

First, recall  the general expression for the variance from \eqref{eq:var_Tab}:
\ifthenelse{\boolean{THESIS_VERSION}}{%
\begin{IEEEeqnarray*}{rCl}
	\Var{T_{b \to a}}& = &\sigma_b^2 \sum_{i=1}^{\kappa_{b \to a}} (t_i-t_{i-1}) \biggl( \sum_{j=i}^{\kappa_{b \to a}} \frac{w_j}{t_j}\biggr)^2 %
	+ \Var{\sum_{i=1}^{\kappa_{b \to a}} w_i Z_{b \to a}^{(t_i)}}.
\end{IEEEeqnarray*}}{%
\begin{IEEEeqnarray*}{rCl}
	\Var{T_{b \to a}}& = &\sigma_b^2 \sum_{i=1}^{\kappa_{b \to a}} (t_i-t_{i-1}) \biggl( \sum_{j=i}^{\kappa_{b \to a}} \frac{w_j}{t_j}\biggr)^2 \nonumber\\
	&& +\> \Var{\sum_{i=1}^{\kappa_{b \to a}} w_i Z_{b \to a}^{(t_i)}}.
\end{IEEEeqnarray*}}

The first sum (the noise coming from the data) does not depend on the privacy mechanism. In the non-MoM case, the weights are $w_1 = w_2 = \dotsb = w_{\kappa_{b \to a}-1} = 0$, $w_{\kappa_{b \to a}}=1$, and we have 
\[
\sigma_b^2 \sum_{i=1}^{\kappa_{b \to a}} (t_i-t_{i-1}) \biggl( \sum_{j=i}^{\kappa_{b \to a}} \frac{w_j}{t_j}\biggr)^2 = \frac{\sigma_b^2}{t_{\kappa_{b \to a}}}.
\]
In the MoM case, the weights are $w_1 = w_2 = \dotsb = w_{\kappa_{b \to a}} = \nicefrac{1}{\kappa_{b \to a}}$, and we obtain
\begin{align*}
	&\frac{\sigma_b^2}{\kappa_{b \to a}^2} \sum_{i=1}^{\kappa_{b \to a}} (t_i-t_{i-1}) \left( \sum_{j=i}^{\kappa_{b \to a}} \frac{1}{t_j}\right)^2 \nonumber = \frac{\sigma_b^2}{\kappa_{b \to a}^2} \sum_{i=1}^{\kappa_{b \to a}} \frac{2i-1}{t_i}.
\end{align*}
Note that for RR scheduling, $\kappa_{b \to a} = O(t_{\kappa_{b \to a}})$.
\subsection{Privacy Mechanism I}

For PM-I, we have
\ifthenelse{\boolean{THESIS_VERSION}}{
\begin{align*}
	\sum_{j=1}^{\kappa_{b \to a}}  w_j Z_{b \to a}^{(t_j)} &=  \sum_{j=1}^{\kappa_{b \to a}}  \frac{w_j}{t_j}\sum_{i=1}^j Z^{(t_{i-1}+1:t_i)}_{b \to a} %
	= \sum_{i=1}^{\kappa_{b \to a}}  \left( \sum_{j=i}^{\kappa_{b \to a}} \frac{w_j}{t_j} \right) Z^{(t_{i-1}+1:t_i)}_{b \to a},
\end{align*}}{
\begin{align*}
	\sum_{j=1}^{\kappa_{b \to a}}  w_j Z_{b \to a}^{(t_j)} &=  \sum_{j=1}^{\kappa_{b \to a}}  \frac{w_j}{t_j}\sum_{i=1}^j Z^{(t_{i-1}+1:t_i)}_{b \to a}\\
	&= \sum_{i=1}^{\kappa_{b \to a}}  \left( \sum_{j=i}^{\kappa_{b \to a}} \frac{w_j}{t_j} \right) Z^{(t_{i-1}+1:t_i)}_{b \to a},
\end{align*}}
from which it follows that
\begin{align} \label{eq:var_noise_term_schemeI}
	\Var{\sum_{j=1}^{\kappa_{b \to a}}  w_j Z_{b \to a}^{(t_j)}} = \sigmaDP^2 \sum_{i=1}^{\kappa_{b \to a}} \left(\sum_{j=i}^{\kappa_{b \to a}} \frac{w_j}{t_j} \right)^2.
\end{align}
Therefore, in the non-MoM case with PM-I, we get
\[
\Var{T_{b \to a}} = \frac{\sigma_b^2}{t_{\kappa_{b \to a}}} + \frac{\kappa_{b \to a} \sigmaDP^2}{t_{\kappa_{b \to a}}^2}, 
\]
while the MoM case with PM-I is as follows:
\ifthenelse{\boolean{THESIS_VERSION}}{
\begin{align*}
	\Var{T_{b \to a}} &= \frac{\sigma_b^2}{\kappa_{b \to a}^2} \sum_{i=1}^{\kappa_{b \to a}} \frac{2i-1}{t_i} + \frac{\sigmaDP^2}{\kappa_{b \to a}^2} \sum_{i=1}^{\kappa_{b \to a}} \left(\sum_{j=i}^{\kappa_{b \to a}} \frac{1}{t_j} \right)^2\\
	& \le \frac{\sigma_b^2}{\kappa_{b \to a}^2} \sum_{i=1}^{\kappa_{b \to a}} \frac{2i-1}{i}+ \frac{\sigmaDP^2}{\kappa_{b \to a}^2} \sum_{i=1}^{\kappa_{b \to a}} \left(\sum_{j=i}^{\kappa_{b \to a}} \frac{1}{j} \right)^2 \\
	&= \frac{\sigma_b^2}{\kappa_{b \to a}^2} \left( 2\kappa_{b \to a} - H_{\kappa_{b \to a}} \right) + \frac{\sigmaDP^2}{\kappa_{b \to a}^2} \left( 2\kappa_{b \to a} - H_{\kappa_{b \to a}} \right)  \\
	&= \frac{\sigma_b^2 + \sigmaDP^2}{\kappa_{b \to a}^2} \left( 2\kappa_{b \to a} - H_{\kappa_{b \to a}} \right) \le 2 \frac{\sigma_b^2 + \sigmaDP^2}{\kappa_{b \to a}},
\end{align*}}{
\begin{align*}
	&\Var{T_{b \to a}} = \frac{\sigma_b^2}{\kappa_{b \to a}^2} \sum_{i=1}^{\kappa_{b \to a}} \frac{2i-1}{t_i} + \frac{\sigmaDP^2}{\kappa_{b \to a}^2} \sum_{i=1}^{\kappa_{b \to a}} \left(\sum_{j=i}^{\kappa_{b \to a}} \frac{1}{t_j} \right)^2\\
	& \le \frac{\sigma_b^2}{\kappa_{b \to a}^2} \sum_{i=1}^{\kappa_{b \to a}} \frac{2i-1}{i}+ \frac{\sigmaDP^2}{\kappa_{b \to a}^2} \sum_{i=1}^{\kappa_{b \to a}} \left(\sum_{j=i}^{\kappa_{b \to a}} \frac{1}{j} \right)^2 \\
	&= \frac{\sigma_b^2}{\kappa_{b \to a}^2} \left( 2\kappa_{b \to a} - H_{\kappa_{b \to a}} \right) + \frac{\sigmaDP^2}{\kappa_{b \to a}^2} \left( 2\kappa_{b \to a} - H_{\kappa_{b \to a}} \right)  \\
	&= \frac{\sigma_b^2 + \sigmaDP^2}{\kappa_{b \to a}^2} \left( 2\kappa_{b \to a} - H_{\kappa_{b \to a}} \right) \le 2 \frac{\sigma_b^2 + \sigmaDP^2}{\kappa_{b \to a}},
\end{align*}}
where $H_j \triangleq  \sum_{i=1}^j \nicefrac{1}{i} = O(\log(j))$ denotes the $j$-th harmonic number.

\subsection{Privacy Mechanism II}

Let $\mathsf{Bin}_i(j)$ denote the $i$-th significant bit of the binary representation of the integer $j$. Then, for PM-II we have
\ifthenelse{\boolean{THESIS_VERSION}}{
	\begin{align}
		&\Var{\sum_{i=1}^{\kappa_{b \to a}} w_i Z_{b \to a}^{(t_i)}} %
		= \sigmaDP^2  
		\sum_{l=1}^{\lfloor \log_2 \kappa_{b \to a} \rfloor +1} \sum_{i = 0}^{l-1}  \left( \sum_{j=2^{l-1}}^{\min\left(2^l-1,\kappa_{b \to a}\right)}  \mathsf{Bin}_{l-i}(j) \frac{w_j}{t_j}\right)^2. \label{eq:var_noise_term_schemeII}
	\end{align}}{
	\begin{align}
		&\Var{\sum_{i=1}^{\kappa_{b \to a}} w_i Z_{b \to a}^{(t_i)}} \nonumber \\
		&= \sigmaDP^2  
		\sum_{l=1}^{\lfloor \log_2 \kappa_{b \to a} \rfloor +1} \sum_{i = 0}^{l-1}  \left( \sum_{j=2^{l-1}}^{\min\left(2^l-1,\kappa_{b \to a}\right)}  \mathsf{Bin}_{l-i}(j) \frac{w_j}{t_j}\right)^2. \label{eq:var_noise_term_schemeII}
	\end{align}}
	For the non-MoM case with PM-II, we have
	\[
	\Var{T_{b \to a}} = \frac{\sigma_b^2}{t_{\kappa_{b \to a}}} + \frac{w_{\mathrm{H}}(\kappa_{b \to a}) \sigmaDP^2}{t_{\kappa_{b \to a}}^2},
	\]
	while for the MoM case with PM-II,  we have
	\ifthenelse{\boolean{THESIS_VERSION}}{
	\begin{align*}
		\Var{T_{b \to a}} &= \frac{\sigma_b^2}{\kappa_{b \to a}^2} \sum_{i=1}^{\kappa_{b \to a}} \frac{2i-1}{t_i} %
		+ \frac{\sigmaDP^2}{\kappa_{b \to a}^2} \sum_{l=1}^{\lfloor \log_2 \kappa_{b \to a} \rfloor +1} \sum_{i = 0}^{l-1}  \left( \sum_{j=2^{l-1}}^{\min\left(2^l-1,\kappa_{b \to a}\right)}   \frac{\mathsf{Bin}_{l-i}(j)}{t_j}\right)^2 %
		\le \frac{2 \sigma_b^2}{\kappa_{b \to a}} + \frac{2\sigmaDP^2}{\kappa_{b \to a}}.
	\end{align*}}{
	\begin{align*}
		&\Var{T_{b \to a}} = \frac{\sigma_b^2}{\kappa_{b \to a}^2} \sum_{i=1}^{\kappa_{b \to a}} \frac{2i-1}{t_i} \\
		&\quad + \frac{\sigmaDP^2}{\kappa_{b \to a}^2} \sum_{l=1}^{\lfloor \log_2 \kappa_{b \to a} \rfloor +1} \sum_{i = 0}^{l-1}  \left( \sum_{j=2^{l-1}}^{\min\left(2^l-1,\kappa_{b \to a}\right)}   \frac{\mathsf{Bin}_{l-i}(j)}{t_j}\right)^2 \\
		&\le \frac{2 \sigma_b^2}{\kappa_{b \to a}} + \frac{2\sigmaDP^2}{\kappa_{b \to a}}.
	\end{align*}}
	
	As a final note, in all these cases, and with RR scheduling, we have that $\Var{T_{b \to a}} = O(\nicefrac 1{t_{\kappa_{b \to a}}})$. The RR scheduling ensures infinite updates of the statistic $T_{b \to a}$.

\section{Proof of Theorem~\ref{thm:1}} \label{app:proof-main-theorem}
First, let us mention some of the properties of RVs satisfying Bernstein's condition.
\ifthenelse{\boolean{THESIS_VERSION}}{
\begin{proposition} \label{prop:iid_bernstein}
If $X_i \sim \BC{\mu_i}{\sigma_i^2}{\beta_i}$, %
$i \in [n]$, are mutually independent, then $X = \delta_1 X_1 + \cdots + \delta_n X_n \sim \BC{\mu}{\sigma^2}{\beta}$, where $\mu = \delta_1 \mu_1+\cdots+\delta_n \mu_n$, $\sigma^2 = \delta_1^2 \sigma_1^2+\cdots+ \delta_n^2 \sigma_n^2$, and $\beta = \sqrt n\max\{|\delta_1| \beta_1,\ldots,|\delta_n| \beta_n,$ $|\delta_1| \sigma_1,\ldots,|\delta_n| \sigma_n\}$.
\end{proposition}}{
\begin{proposition} \label{prop:iid_bernstein}
If $X_i \sim \BC{\mu_i}{\sigma_i^2}{\beta_i}$, %
$i \in [n]$, are mutually independent, then $X = \delta_1 X_1 + \cdots + \delta_n X_n \sim \BC{\mu}{\sigma^2}{\beta}$, where $\mu = \delta_1 \mu_1+\cdots+\delta_n \mu_n$, $\sigma^2 = \delta_1^2 \sigma_1^2+\cdots+ \delta_n^2 \sigma_n^2$, and $\beta = \sqrt n\max\{|\delta_1| \beta_1,\ldots,|\delta_n| \beta_n, |\delta_1| \sigma_1,\ldots,|\delta_n| \sigma_n\}$.
\end{proposition}}
\begin{IEEEproof}
See Appendix~\ref{app:proof_prop:4} in the supplementary material.
\end{IEEEproof}

\begin{proposition}[\hspace{-0.01cm}{\cite[Prop. 2.10]{Wainwright19_1}}]
\label{prop:BC-tails}
	If $X \sim \BC{\mu}{\sigma^2}{\beta}$, then the tails of its distribution are bounded as follows,
	\[
	\Prv{|X-\mu| \ge x} \le 2 {\mathrm e}^{-\frac{x^2}{2(\sigma^2 + \beta x)}},
	\]
	for all $x \ge 0$. 
\end{proposition}

Let RVs from $\mathcal D_a$ satisfy Bernstein's condition with parameter $\beta_a$ for all $a \in [\numa]$. Then, $\bar{X}_a^{(t)} \sim \BC{\mu_a}{\frac{\sigma_a^2}t}{\frac{\max(\beta_a, \sigma_a)}{\sqrt t}}$. We can also show that $T_{b \to a}$ satisfies Bernstein's condition.

\begin{lemma} \label{lem:Tba_BC}
	Let $Z_{b \to a}^{(\tau_{i-1}+1:\tau_i)} \sim \BC{0}{\sigmaDP^2}{\beta_{\mathrm{DP}}}$ %
	(cf. \cref{sec:privacy}). Then, under RR scheduling there exist $\sigma_T^2$ and $\beta_T>0$ independent of $t$ such that $T_{b \to a} \sim \BC{\mu_b}{\Var{T_{b \to a}}}{\frac{\beta_T \ln t}{\sqrt{t}}}$, where $\Var{T_{b \to a}} \le \frac{\sigma_T^2}{t}$.
\end{lemma}

\begin{IEEEproof}
See Appendix~\ref{app:proof_Tba_BC} in the supplementary material.
\end{IEEEproof}

Next, we show an asymptotic upper bound on the probability of type-II error in the settings of Theorem~\ref{thm:1}.

\begin{lemma}\label{lem:prob-type-II}
For $b \notin \mathcal C_a$, the probability of type-II error under RR scheduling is
\[
	\Prv{\chi_a^{(t)}(b; \theta_t) = 1} = o\left( \frac{1}{t^n}\right)
\]
for any positive integer $n$.
\end{lemma}

\begin{IEEEproof}
See Appendix~\ref{app:proof_lem:prob-type-II} in the supplementary material.
\end{IEEEproof}

Note that in the proof, we only needed the order $\Phi^{-1}\left(1 - \nicefrac{\theta_t}2\right) = o(\sqrt t)$. Thus, the same asymptotic result will hold for any term instead of $\Phi^{-1}\left(1 - \nicefrac{\theta_t}2\right)$ of the same order.

Now, we can prove Theorem~\ref{thm:1} as follows,
\ifthenelse{\boolean{THESIS_VERSION}}{
\begin{align*}
	\EE{\left( \mu_a^{(t)} - \mu_a \right)^2} 
	&= \EE{\left( \mu_a^{(t)} - \mu_a \right)^2 \mid \mathcal C_a^{(t)} \subseteq \mathcal C_a} \Prv{\mathcal C_a^{(t)} \subseteq \mathcal C_a} \\
	&\quad + \EE{\left( \mu_a^{(t)} - \mu_a \right)^2 \mid \mathcal C_a^{(t)} \setminus \mathcal C_a \neq \emptyset} \Prv{\mathcal C_a^{(t)} \setminus \mathcal C_a \neq \emptyset}. 
\end{align*}}{
\begin{align*}
	&\EE{\left( \mu_a^{(t)} - \mu_a \right)^2} \\
	&= \EE{\left( \mu_a^{(t)} - \mu_a \right)^2 \mid \mathcal C_a^{(t)} \subseteq \mathcal C_a} \Prv{\mathcal C_a^{(t)} \subseteq \mathcal C_a} \\
	&\quad + \EE{\left( \mu_a^{(t)} - \mu_a \right)^2 \mid \mathcal C_a^{(t)} \setminus \mathcal C_a \neq \emptyset} \Prv{\mathcal C_a^{(t)} \setminus \mathcal C_a \neq \emptyset}. 
\end{align*}}
The second term is easy to bound asymptotically as follows,
\ifthenelse{\boolean{THESIS_VERSION}}{
\begin{align*}
	\EE{\left( \mu_a^{(t)} - \mu_a \right)^2 \mid \mathcal C_a^{(t)} \setminus \mathcal C_a \neq \emptyset} \Prv{\mathcal C_a^{(t)} \setminus \mathcal C_a \neq \emptyset} 
	&\le V \cdot \Prv{\mathcal C_a^{(t)} \setminus \mathcal C_a \neq \emptyset} \\
	&\le V \sum_{b \notin \mathcal C_a} \Prv{\chi_a^{(t)}(b; \theta_t) = 1} = o\left( \frac{1}{t^n}\right),
\end{align*}}{
\begin{align*}
	&\EE{\left( \mu_a^{(t)} - \mu_a \right)^2 \mid \mathcal C_a^{(t)} \setminus \mathcal C_a \neq \emptyset} \Prv{\mathcal C_a^{(t)} \setminus \mathcal C_a \neq \emptyset} \\
	&\le V \cdot \Prv{\mathcal C_a^{(t)} \setminus \mathcal C_a \neq \emptyset} \\
	&\le V \sum_{b \notin \mathcal C_a} \Prv{\chi_a^{(t)}(b; \theta_t) = 1} = o\left( \frac{1}{t^n}\right),
\end{align*}}
for any positive integer $n$ by Lemma~\ref{lem:prob-type-II}. Here, $V$ is some positive constant (expectation is bounded).

If $\mathcal C_a^{(t)} \subseteq \mathcal C_a$, then $\EE{T_{b \to a}} = \mu_b = \mu_a$ for all $b \in \mathcal C_a^{(t)}$ and $\mu_a^{(t)}$ is an unbiased estimate of $\mu_a$. Then,
\ifthenelse{\boolean{THESIS_VERSION}}{
\begin{align*}
	\EE{\left( \mu_a^{(t)} - \mu_a \right)^2 \mid \mathcal C_a^{(t)} \subseteq \mathcal C_a} = \Var{\mu_a^{(t)}} 
	& = 
	\sum_{b \in \mathcal C_a} \left( \alpha_{b \to a}^{(t)} \right)^2 \Var{T_{b \to a}}
	\le (1-\alpha) \frac{\sigma_a^2}t,
\end{align*}}{
\begin{align*}
	&\EE{\left( \mu_a^{(t)} - \mu_a \right)^2 \mid \mathcal C_a^{(t)} \subseteq \mathcal C_a} = \Var{\mu_a^{(t)}} \\
	& = 
	\sum_{b \in \mathcal C_a} \left( \alpha_{b \to a}^{(t)} \right)^2 \Var{T_{b \to a}}
	\le (1-\alpha) \frac{\sigma_a^2}t,
\end{align*}}
where the inequality is from Lemma~\ref{lem:weighting}, and $\alpha \in (0,1)$ is some constant independent of $t$. (Can be derived from constant upper bounds on $t\Var{T_{b \to a}} = O(1)$.)

Note that here we implicitly used the fact that $\Prv{\mathcal C_a^{(t)} = \{a\}} < 1$, which is true for $|\mathcal C_a|\ge 2$, because our decision rule is correct with nonzero probability (the intuition being that any interval decision rule gives correct answer with nonzero probability for RVs with support $(-\infty,+\infty)$).

Altogether,
\[
\EE{\left( \mu_a^{(t)} - \mu_a \right)^2} \le (1-\alpha) \frac{\sigma_a^2}t + o\left( \frac 1{t^n}\right) \le \left( 1 - \frac{\alpha}2 \right) \frac{\sigma_a^2}t,
\]
for all large enough $t$, which concludes the proof.

\ifthenelse{\boolean{THESIS_VERSION}}{}{\IEEEtriggeratref{2}}
\bibliographystyle{IEEEtran}
\bibliography{\bibfiles/defshort1,\bibfiles/biblioHY,\bibfiles/refs}

\ifCLASSOPTIONcaptionsoff
  \newpage
\fi

\ifthenelse{\boolean{MAIN_REF}}{}{\clearpage
\documentclass[perslearn_arxiv.tex]{subfiles}

\ifSubfilesClassLoaded{%
  \setcounter{page}{1}
  \setcounter{theorem}{1}
  \setcounter{equation}{21}
  \setcounter{figure}{3}
}{}%

\newenvironment{Figure}
  {\par\medskip\noindent\minipage{\linewidth}}
  {\endminipage\par\medskip}

\title{Differentially-Private Collaborative Online Personalized Mean Estimation: Supplementary Material}

\begin{document}

\ifthenelse{\boolean{THESIS_VERSION}}{
\onecolumn %
\hrule height 4pt
\vskip 0.25in
\vskip -\parskip%
{\centering
{\LARGE\bf Differentially-Private Collaborative Online Personalized\\[1mm] Mean Estimation: Supplementary Material\par}}
\vskip 0.29in
\vskip -\parskip
\hrule height 1pt
\vskip 0.09in%
\vspace{2ex}
}{
\twocolumn[%
\hrule height 4pt
\vskip 0.25in
\vskip -\parskip%
{\centering
{\LARGE\bf Differentially-Private Collaborative Online Personalized\\[1mm] Mean Estimation: Supplementary Material\par}}
\vskip 0.29in
\vskip -\parskip
\hrule height 1pt
\vskip 0.09in%
\vspace{2ex}
]}

   \begin{bibunit}[IEEEtranSA]

\appendices

\setcounter{section}{2}

\section{Proofs} \label{app:DP}

\subsection{Proof of  Lemma~\ref{lem:DP}} \label{app:proof_lem:DP}
  The  $\ell_2$ \emph{sensitivity} 
  $\Delta_2$ 
  for the sample mean function 
  $\nicefrac{(x_1+\cdots+x_n)}{n}$ %
  is 
  $\Delta_2  \triangleq 
  \max_{x_i \neq x'_i} \bignorm[2]{\nicefrac{(x_1+\dotsb+x_i+\dotsb+x_n)}{n} - \nicefrac{(x_1+\dotsb+x'_i+\dotsb+x_n)}{n}} = \nicefrac{2L}{n}$, and then according to \cite[Thm.~A.1]{DworkRoth14_1} the addition of Gaussian noise with variance at least $\nicefrac{(2\ln(\nicefrac{1.25}{\delta}) \Delta_2^2)}{\epsilon^2} = \nicefrac{\sigmaDP^2}{n^2}$ provides $(\epsilon,\delta)$-DP, which proves Lemma~\ref{lem:DP}. %

\subsection{Proof of  Lemma~\ref{lem:DP_Laplace}} \label{app:proof_lem:DP_Laplace}
\ifthenelse{\boolean{THESIS_VERSION}}{
For Laplace noise, the $\ell_1$ sensitivity $\Delta_1$ is used. As $\Delta_1  \triangleq 
  \max_{x_i \neq x'_i} \bignorm[1]{\nicefrac{(x_1+\dotsb+x_i+\dotsb+x_n)}{n} - \nicefrac{(x_1+\dotsb+x'_i+\dotsb+x_n)}{n}}$ $= \nicefrac{2L}{n} = \Delta_2$ for the sample mean function, then according to  \cite[Thm.~3.6]{DworkRoth14_1} the addition of Laplace noise with variance at least $\nicefrac{(2 \Delta_1^2)}{\epsilon^2} = \nicefrac{\sigmaDP^2}{n^2}$ provides $(\epsilon,0)$-DP, which proves Lemma~\ref{lem:DP_Laplace}.}{
For Laplace noise, the $\ell_1$ sensitivity $\Delta_1$ is used. As $\Delta_1  \triangleq 
  \max_{x_i \neq x'_i} \bignorm[1]{\nicefrac{(x_1+\dotsb+x_i+\dotsb+x_n)}{n} - \nicefrac{(x_1+\dotsb+x'_i+\dotsb+x_n)}{n}} = \nicefrac{2L}{n} = \Delta_2$ for the sample mean function, then according to  \cite[Thm.~3.6]{DworkRoth14_1} the addition of Laplace noise with variance at least $\nicefrac{(2 \Delta_1^2)}{\epsilon^2} = \nicefrac{\sigmaDP^2}{n^2}$ provides $(\epsilon,0)$-DP, which proves Lemma~\ref{lem:DP_Laplace}.}  

\subsection{Proof of  Lemma~\ref{lem:DP_squared}}  \label{app:proof_lem:DP_squared}
To prove Lemma~\ref{lem:DP_squared} we first need to compute the $\ell_2$ sensitivity $\Delta_2$ for the sample variance function $F(x_1,\ldots,x_n) \triangleq \nicefrac{(x_1^2+\cdots+x_n^2)}{(n-1)} -  \nicefrac{(x_1+\cdots+x_n)^2}{n(n-1)}$. In the following, for simplicity, define $\tilde{F}(x_1,\ldots,x_n) \triangleq (n-1) F(x_1,\ldots,x_n)$. The function is nonnegative and convex in $x_1,\ldots,x_n$. Hence, we proceed as follows.

Without loss of generality, take 
 the partial derivative of $F(x_1,\ldots,x_n)$ with respect to $x_1$ and set it equal to zero, which leads to $x_1 = \nicefrac{(x_2+\cdots+x_n)}{(n-1)}$. Note that the range of  $\nicefrac{(x_2+\cdots+x_n)}{(n-1)}$ is from $\mu-L$ to $\mu+L$. Thus, the minimum value of $F(x_1,\ldots,x_n)$ (for fixed $x_2,\ldots,x_n$) is always within the valid range of $x_1$. Therefore,  the $\ell_2$ sensitivity $\Delta_2$ of $F(x_1,\ldots,x_n)$ is equal to
 \ifthenelse{\boolean{THESIS_VERSION}}{
 \begin{align*}
 \frac{1}{n-1} \cdot \max & \left[\tilde{F}(\mu-L,x_2,\ldots,x_n)- %
 \tilde{F}\left(\frac{x_2+\cdots+x_n}{n-1},x_2,\ldots,x_n \right), \right. \\
 & \left. \;\tilde{F}(\mu+L,x_2,\ldots,x_n)-\tilde{F}\left(\frac{x_2+\cdots+x_n}{n-1},x_2,\ldots,x_n \right) \right],
 \end{align*}}{
 \begin{align*}
& \frac{1}{n-1} \cdot \max \left[\tilde{F}(\mu-L,x_2,\ldots,x_n)-\right. \\
 & \left. \tilde{F}\left(\frac{x_2+\cdots+x_n}{n-1},x_2,\ldots,x_n \right), \right. \\
 & \left. \;\tilde{F}(\mu+L,x_2,\ldots,x_n)-\tilde{F}\left(\frac{x_2+\cdots+x_n}{n-1},x_2,\ldots,x_n \right) \right],
 \end{align*}}
 where the maximization is both over the two arguments inside the $\max$ and over $x_2,\ldots,x_n$.  
 Evaluating $\tilde{F}(x_1,\ldots,x_n)$ at $x_1=\nicefrac{(x_2+\cdots+x_n)}{(n-1)}$ gives
 \ifthenelse{\boolean{THESIS_VERSION}}{
\begin{align}
&\left( \frac{x_2+\cdots+x_n}{n-1} \right)^2 + x_2^2 + \cdots + x_n^2 - \frac{1}{n}\left( \frac{x_2+\cdots+x_n}{n-1} \right)^2 %
- 2 \frac{(x_2+\cdots+x_n)^2}{n(n-1)} -  \frac{(x_2+\cdots+x_n)^2}{n} \nonumber \\
&=  -\frac{1}{n-1}\left( x_2+\cdots+x_n \right)^2  + x_2^2 + \cdots + x_n^2. \label{eq:evaluate_1}
\end{align}}{
\begin{align}
&\left( \frac{x_2+\cdots+x_n}{n-1} \right)^2 + x_2^2 + \cdots + x_n^2 - \frac{1}{n}\left( \frac{x_2+\cdots+x_n}{n-1} \right)^2 \nonumber \\
&\quad\, - 2 \frac{(x_2+\cdots+x_n)^2}{n(n-1)} -  \frac{(x_2+\cdots+x_n)^2}{n} \nonumber \\
&=  -\frac{1}{n-1}\left( x_2+\cdots+x_n \right)^2  + x_2^2 + \cdots + x_n^2. \label{eq:evaluate_1}
\end{align}}
while evaluating $\tilde{F}(x_1,\ldots,x_n)$   at $x_1=\mu \pm L$ gives
 \ifthenelse{\boolean{THESIS_VERSION}}{
\begin{align}
&\left( \mu \pm L \right)^2 + x_2^2 + \cdots + x_n^2 - \frac{1}{n}\left( \mu \pm L \right)^2 %
- 2 (\mu \pm L)\frac{x_2+\cdots+x_n}{n} - \frac{(x_2+\cdots+x_n)^2}{n}. \label{eq:evaluate_2}
\end{align}}{
\begin{align}
&\left( \mu \pm L \right)^2 + x_2^2 + \cdots + x_n^2 - \frac{1}{n}\left( \mu \pm L \right)^2 \nonumber \\
&- 2 (\mu \pm L)\frac{x_2+\cdots+x_n}{n} - \frac{(x_2+\cdots+x_n)^2}{n}. \label{eq:evaluate_2}
\end{align}}
Now, taking the difference between \eqref{eq:evaluate_2} and \eqref{eq:evaluate_1} gives
\ifthenelse{\boolean{THESIS_VERSION}}{
\begin{align}
&\frac{n-1}{n} (\mu \pm L)^2 - \frac{2}{n}(\mu \pm L)(x_2+\cdots+x_n) %
+ \frac{1}{n(n-1)} (x_2+\cdots+x_n)^2. \label{eq:evaluate_3}
\end{align}}{
\begin{align}
&\frac{n-1}{n} (\mu \pm L)^2 - \frac{2}{n}(\mu \pm L)(x_2+\cdots+x_n) \nonumber \\
&+ \frac{1}{n(n-1)} (x_2+\cdots+x_n)^2. \label{eq:evaluate_3}
\end{align}}
The expression in \eqref{eq:evaluate_3} is convex in $x_2+\cdots+x_n$. Hence, it attains its maximum value at one of the end points $x_2+\cdots+x_n=(n-1)(\mu \pm L)$,  %
which leads to the following derivations for $\Delta_2$,  %
\ifthenelse{\boolean{THESIS_VERSION}}{
\begin{align*}
\Delta_2 &= \frac{1}{n-1} \cdot  \max_{a,b \in \{0,1\}} \left[ \frac{n-1}{n} (\mu + (-1)^a L)^2 %
- \frac{2(n-1)}{n}(\mu  + (-1)^a L)(\mu + (-1)^b L) \right. \\
&\quad\quad\quad\quad\quad\quad\quad\;\;\; + \left. \frac{n-1}{n} (\mu +(-1)^b L)^2 \right] \\
&= \frac{1}{n} \cdot \max_{a,b \in \{0,1\}}  \left[ \mu^2 + 2(-1)^a \mu L + L^2 %
- 2\mu^2-2 (-1)^{a+b} L^2  - 2((-1)^a+(-1)^b) \mu L \right. \\
&\quad\quad\quad\quad\quad\;\;\;\; \left. + \mu^2 + 2(-1)^b \mu L + L^2 \right] \\
&= \frac{1}{n} \cdot \max_{a,b \in \{0,1\}}  \left[ 2L^2 -2 (-1)^{a+b} L^2  \right] \\
&= \frac{4L^2}{n},
\end{align*}}{
\begin{align*}
\Delta_2 &= \frac{1}{n-1} \cdot \max_{a,b \in \{0,1\}} \left[ \frac{n-1}{n} (\mu + (-1)^a L)^2 \right. \\
&\quad\quad\quad\quad\quad - \left. \frac{2(n-1)}{n}(\mu  + (-1)^a L)(\mu + (-1)^b L) \right. \\
&\quad\quad\quad\quad\quad + \left. \frac{n-1}{n} (\mu +(-1)^b L)^2 \right] \\
&= \frac{1}{n} \cdot \max_{a,b \in \{0,1\}}  \left[ \mu^2 + 2(-1)^a \mu L + L^2 \right. \\
&\quad\quad\quad\quad\quad \left. - 2\mu^2-2 (-1)^{a+b} L^2  - 2((-1)^a+(-1)^b) \mu L \right. \\
&\quad\quad\quad\quad\quad \left. + \mu^2 + 2(-1)^b \mu L + L^2 \right] \\
&= \frac{1}{n} \cdot \max_{a,b \in \{0,1\}}  \left[ 2L^2 -2 (-1)^{a+b} L^2  \right] \\
&= \frac{4L^2}{n},
\end{align*}}
from which Lemma~\ref{lem:DP_squared} follows from \cite[Thm.~A.1]{DworkRoth14_1}.

\subsection{Proof of Lemma~\ref{lem:keeping_last_optimal_schemeII}} \label{sec:proof_keeping_last_optimal_schemeII}

We provide the proof for the case of RR scheduling. 
 In this case, $t_i = i (M-1)$. %
 By combining \eqref{eq:var_Tab} and \eqref{eq:var_noise_term_schemeI} (in Appendix~\ref{app:Tba_variances}) it follows that 
\begin{align*}
\Var{T_{a \to b}} &= \left(\frac{\sigmaDP^2}{(M-1)^2} + \frac{\sigma_b^2}{M-1} \right)\sum_{i=1}^{\kappa_{b \to a}} \left(\sum_{j=i}^{\kappa_{b \to a}} \frac{w_j}{j} \right)^2.
\end{align*}
Next, by using Lagrange multipliers and taking the derivative with respect to $w_j$ results  in 
\begin{align} \label{eq:lemma5}
\left( \frac{\sigmaDP^2}{(M-1)^2} +  \frac{\sigma_b^2}{M-1} \right) \cdot \frac{2}{j} \sum_{i=1}^{j} \sum_{l=i}^{\kappa_{b \to a}} \frac{w_l}{l}-\lambda = 0,
\end{align}
for $j=1,\dotsc,\kappa_{b \to a}$, 
where the Lagrange multiplier $\lambda$ is due to the constraint $\sum_{i=1}^{\kappa_{b \to a}} w_i = 1$. Combining the expression in \eqref{eq:lemma5} for $j$ and $j-1$, results in
\begin{align} \label{eq:ai_rec}
\frac{(j-1)\lambda'}{j} + \frac{2}{j} \sum_{i=j}^{\kappa_{b \to a}} \frac{w_i}{i} = \lambda',
\end{align}
where $\lambda' \triangleq   \frac{\lambda}{\nicefrac{\sigmaDP^2}{(M-1)^2} +  \nicefrac{\sigma_b^2}{(M-1)}}$.
Setting $j=\kappa_{b \to a}$ in \eqref{eq:ai_rec} results in $w_{\kappa_{b \to a}} = \nicefrac{\kappa_{b \to a} \lambda'}{2}$. Then, setting $j=\kappa_{b \to a}-1$ in \eqref{eq:ai_rec} and substituting $w_{\kappa_{b \to a}} = \nicefrac{\kappa_{b \to a} \lambda'}{2}$ results in $w_{\kappa_{b \to a}-1}=0$, and by the recursion in \eqref{eq:ai_rec}  all remaining $w_i$'s become zero, which concludes the proof. %
The same proof holds for rRR scheduling by replacing  $M$ by $|\mathcal{C}_a|$.

\subsection{Proof of Proposition~\ref{prop:2}} \label{app:prop2}
The result follows by mapping the current time $t$ to the agent $b$ (which may or may not be in the class of agent $a$) being queried at this specific time and also the exact number of times this particular agent has been queried by agent $a$ up to (and including) time $t$, i.e., the current value of $\kappa_{b \to a}$, and then combining this with \eqref{eq:var_muat} for the variance of the estimate $\mu_a^{(t)}$ at time $t$ and \eqref{eq:var_Tab}. By numbering all agents,  excluding agent $a$, from $1$ to $M-1$, it can readily be seen that the agent queried at time $t$ is $b = 
(t-\thres) \bmod (M-1)+1$ with  RR scheduling. Agent $b$ may or may not be in the class of agent $a$.  Moreover, 
\[
\kappa_{b \to a} =  1+ \left\lfloor \frac{t-\thres}{M-1} \right\rfloor,
\]
i.e., $\kappa_{b \to a}=1$ for $1 \leq t \leq M-1$, $\kappa_{b \to a}=2$ for $M \leq t \leq 2(M-1)$, etc. For the agents $j \in [b]$ processed before (and including) agent $b$ in the same \emph{round}, there have been the same number of  $\kappa_{j \to a}= \kappa_{b \to a}$ updates to the linear statistic $T_{j \to a}$ as for $T_{b \to a}$, while for the agents $j \in [(t-\thres) \bmod (M-1)+2: M-1]$ the last update was in the previous \emph{round}. Hence, for these agents the corresponding number of updates to the linear statistic $T_{j \to a}$ is $\kappa_{j \to a}= \kappa_{b \to a}-1$.  Next, the sequence of times an agent $j \in [M-1]$ is queried is given by $t^{(j)}_i = \thres+(i-1)(M-1)+j-1$, i.e., for agent $1$ it is $1,1+(M-1),1+2(M-1)$, etc., for agent $2$ it is $2,2+(M-1),2+2(M-1)$, etc., and so on. 
Now, number the agents in the class of agent $a$,  excluding agent $a$ itself, from $\ell_1$ to $\ell_{n-1}$, where $1 \leq \ell_1 < \cdots < \ell_{n-1} < M$ and $n$ is the number of agents in the class of agent $a$. 
The result now follows from \eqref{eq:var_muat} and \eqref{eq:var_Tab} and by summing over all agents $\ell_j$, $j \in [n-1]$, in the class of agent $a$, multiplied with the probability $\binom{M-1}{n-1} p^{n-1} (1-p)^{\numa-n}$ of having  $n$ agents in the class of agent $a$ (the number of agents in the class of agent $a$ follows a binomial distribution), and then finally averaging over the number of ways of selecting $\ell_1$ to $\ell_{n-1}$ such that $1 \leq \ell_1 < \cdots < \ell_{n-1} < M$ and summing over the number of agents in the class of agent $a$ from $1$ to $M-1$. Note that the binomial coefficient from the probability mass function of the binomial RV giving the class size $n$, $\binom{M-1}{n-1}$, cancels with  the number of ways of selecting $\ell_1$ to $\ell_{n-1}$, since this number is equal to the same binomial coefficient $\binom{M-1}{n-1}$.

\subsection{Proof of Proposition~\ref{prop:iid_bernstein}}  \label{app:proof_prop:4}
This proof is based on discussions on \cite{479502}.

Assume that $\delta_1=\cdots=\delta_n=1$ (and, thus, $\sigma^2 = \sigma_1^2 + \dotsb + \sigma_n^2$). The result for arbitrary constants follows directly as  $\delta_i X_i \sim \BC{\delta_i \mu_i}{\delta_i^2 \sigma_i^2}{|\delta_i|\beta_i}$ when $X_i \sim \BC{\mu_i}{\sigma_i^2}{\beta_i}$. Also, further in the proof we assume for simplicity that all $\mu_i=0$, as the Bernstein's condition is only concerned with central moments. We use a shorthand $\beta_{\mathrm{max}}  \triangleq \max \{\beta_1,\ldots,\beta_n, \sigma_1,\ldots,\sigma_n\}$.

For $k=3$ and arbitrary $n$, we have
\ifthenelse{\boolean{THESIS_VERSION}}{
	\begin{align*}
		\left|\EE{ \left(X_1+\cdots+X_n \right)^3}\right|  %
		&\leq \sum_{i=1}^n \left| \EE{X_i^3} \right|
		+ 3 \sum_{i_1 \neq i_2} \left| \EE{X_{i_1}^2} \right| \cdot \left| \EE{X_{i_2}} \right| \\
		&\quad + 6 \sum_{1 \le i_1<i_2<i_3 \le n} \left| \EE{X_{i_1}} \right| \cdot \left| \EE{X_{i_2}} \right| \cdot \left| \EE{X_{i_3}} \right| \\
		&\overset{(a)}{=}\sum_{i=1}^n \left| \EE{X_i^3} \right|
		\overset{(b)}{\le} \frac 12 k! \sigma^2 \beta_{\max}^{k-2}
		\leq \frac 12 k! \sigma^2 (\sqrt n \beta_{\max})^{k-2}.
	\end{align*}}{
	\begin{align*}
		&\left|\EE{ \left(X_1+\cdots+X_n \right)^3}\right|  \\
		&\leq \sum_{i=1}^n \left| \EE{X_i^3} \right|
		+ 3 \sum_{i_1 \neq i_2} \left| \EE{X_{i_1}^2} \right| \cdot \left| \EE{X_{i_2}} \right| \\
		&\quad + 6 \sum_{1 \le i_1<i_2<i_3 \le n} \left| \EE{X_{i_1}} \right| \cdot \left| \EE{X_{i_2}} \right| \cdot \left| \EE{X_{i_3}} \right| \\
		&\overset{(a)}{=}\sum_{i=1}^n \left| \EE{X_i^3} \right|
		\overset{(b)}{\le} \frac 12 k! \sigma^2 \beta_{\max}^{k-2}
		\leq \frac 12 k! \sigma^2 (\sqrt n \beta_{\max})^{k-2}.
	\end{align*}}
	Here, %
	$(a)$ is because $\EE{X_i} = 0$ for all $i$ and $(b)$ follows from the individual Bernstein's conditions.

Further in the proof, we assume $k \ge 4$. We have
\ifthenelse{\boolean{THESIS_VERSION}}{
\begin{align*}
	\left| \EE{\left(X_1+\dotsb+X_n\right)^k} \right| %
	&= \left| \sum_{\substack{i_1+\dotsb+i_n=k \\ i_1,\dotsc,i_n \ge 0}} \frac{k!}{i_1!\cdot \dots \cdot i_n!} \EE{X_1^{i_1}} \cdot \dots \cdot \EE{X_n^{i_n}}\right| \\
	&\le \sum_{\substack{i_1+\dotsb+i_n=k \\ i_1,\dotsc,i_n \ge 0}} \frac{k!}{i_1!\cdot \dots \cdot i_n!} \prod_{j=1}^n \left|\EE{X_j^{i_j}}\right|.
\end{align*}}{
\begin{align*}
	&\left| \EE{\left(X_1+\dotsb+X_n\right)^k} \right| \\
	&= \left| \sum_{\substack{i_1+\dotsb+i_n=k \\ i_1,\dotsc,i_n \ge 0}} \frac{k!}{i_1!\cdot \dots \cdot i_n!} \EE{X_1^{i_1}} \cdot \dots \cdot \EE{X_n^{i_n}}\right| \\
	&\le \sum_{\substack{i_1+\dotsb+i_n=k \\ i_1,\dotsc,i_n \ge 0}} \frac{k!}{i_1!\cdot \dots \cdot i_n!} \prod_{j=1}^n \left|\EE{X_j^{i_j}}\right|.
\end{align*}}
Note that $\EE{X_j^0} = 1$ and $\EE{X_j}=\mu_j=0$. We continue as follows:
\ifthenelse{\boolean{THESIS_VERSION}}{
\begin{align*}
	\sum_{\substack{i_1+\dotsb+i_n=k \\ i_1,\dotsc,i_n \ge 0}} \frac{k!}{i_1!\cdot \dots \cdot i_n!} \prod_{j=1}^n \left|\EE{X_j^{i_j}}\right| %
	&= \sum_{j=1}^n \left|\EE{X_j^k}\right| + \sum_{\substack{i_1+\dotsb+i_n=k \\ i_1,\dotsc,i_n \neq 1,k-1,k}} \frac{k!}{i_1!\cdot \dots \cdot i_n!} \prod_{i_j \neq 0} \left|\EE{X_j^{i_j}}\right| \\
	&\le \sum_{j=1}^n \frac 12 k! \sigma_j^2 \beta_j^{k-2} %
	+ \sum_{\substack{i_1+\dotsb+i_n=k \\ i_1,\dotsc,i_n \neq 1,k-1,k}} \frac{k!}{i_1!\cdot \dots \cdot i_n!} \prod_{i_j \neq 0} \frac 12 i_j! \sigma_j^2 \beta_j^{i_j-2} \\
	&\le \frac 12 k! \sigma^2 \beta_{\mathrm{max}}^{k-2} + k! \sum_{\substack{i_1+\dotsb+i_n=k \\ i_1,\dotsc,i_n \neq 1,k-1,k \\ w=\supp(i_1,\dotsc,i_n)}} \frac{\beta_{\mathrm{max}}^{k-w}}{2^w}  \prod_{i_j \neq 0} \sigma_j,
\end{align*}}{
\begin{align*}
	&\sum_{\substack{i_1+\dotsb+i_n=k \\ i_1,\dotsc,i_n \ge 0}} \frac{k!}{i_1!\cdot \dots \cdot i_n!} \prod_{j=1}^n \left|\EE{X_j^{i_j}}\right| \\
	&= \sum_{j=1}^n \left|\EE{X_j^k}\right| + \sum_{\substack{i_1+\dotsb+i_n=k \\ i_1,\dotsc,i_n \neq 1,k-1,k}} \frac{k!}{i_1!\cdot \dots \cdot i_n!} \prod_{i_j \neq 0} \left|\EE{X_j^{i_j}}\right| \\
	&\le \sum_{j=1}^n \frac 12 k! \sigma_j^2 \beta_j^{k-2} \\
	&\qquad + \sum_{\substack{i_1+\dotsb+i_n=k \\ i_1,\dotsc,i_n \neq 1,k-1,k}} \frac{k!}{i_1!\cdot \dots \cdot i_n!} \prod_{i_j \neq 0} \frac 12 i_j! \sigma_j^2 \beta_j^{i_j-2} \\
	&\le \frac 12 k! \sigma^2 \beta_{\mathrm{max}}^{k-2} + k! \sum_{\substack{i_1+\dotsb+i_n=k \\ i_1,\dotsc,i_n \neq 1,k-1,k \\ w=\supp(i_1,\dotsc,i_n)}} \frac{\beta_{\mathrm{max}}^{k-w}}{2^w}  \prod_{i_j \neq 0} \sigma_j,
\end{align*}}
where $\supp(i_1,\dotsc,i_n)$ denotes the number of nonzeros among $i_1,\dotsc,i_n$.

Observe that if $J = \{j \in [n] \mid i_j \neq 0\}$, where $|J| = w \ge 2$, then for any $j_1,j_2 \in J$, $j_1 < j_2$, it is true that
\[
	\prod_{j \in J} \sigma_j \le \sigma_{j_1} \sigma_{j_2} \beta_{\mathrm{max}}^{w-2} \le \frac{\sigma_{j_1}^2 + \sigma_{j_2}^2}2 \beta_{\max}^{w-2}.
\]
Summing up over the choice of ordered pairs $(j_1,j_2) \in J^2$, $j_1 < j_2$, we get
\[
	\prod_{j \in J} \sigma_j \le \frac{\beta_{\mathrm{max}}^{w-2}}{\binom w2} \sum_{\substack{j_1,j_2 \in J \\ j_1 < j_2}} \sigma_{j_1} \sigma_{j_2} \le \frac{\beta_{\max}^{w-2}}w \sum_{j \in J} \sigma_j^2.
\]
We continue as follows:
\ifthenelse{\boolean{THESIS_VERSION}}{
\begin{align*}
	&\frac 12 k! \sigma^2 \beta_{\mathrm{max}}^{k-2} 
	+ k! \sum_{\substack{i_1+\dotsb+i_n=k \\ i_1,\dotsc,i_n \neq 1,k-1,k \\ w=\supp(i_1,\dotsc,i_n)}} \frac{\beta_{\mathrm{max}}^{k-w}}{2^w}  \prod_{i_j \neq 0} \sigma_j %
	\le \frac 12 k! \sigma^2 \beta_{\mathrm{max}}^{k-2}  + \frac 12 k! \beta_{\mathrm{max}}^{k-2} \sum_{\substack{i_1+\dotsb+i_n=k \\ i_1,\dotsc,i_n \neq 1,k-1,k \\ w=\supp(i_1,\dotsc,i_n)}} \frac 1{w 2^{w-1}} \sum_{i_j \neq 0} \sigma_j^2.
\end{align*}}{
\begin{align*}
	&\frac 12 k! \sigma^2 \beta_{\mathrm{max}}^{k-2} 
	+ k! \sum_{\substack{i_1+\dotsb+i_n=k \\ i_1,\dotsc,i_n \neq 1,k-1,k \\ w=\supp(i_1,\dotsc,i_n)}} \frac{\beta_{\mathrm{max}}^{k-w}}{2^w}  \prod_{i_j \neq 0} \sigma_j \\
	&\le \frac 12 k! \sigma^2 \beta_{\mathrm{max}}^{k-2}  + \frac 12 k! \beta_{\mathrm{max}}^{k-2} \sum_{\substack{i_1+\dotsb+i_n=k \\ i_1,\dotsc,i_n \neq 1,k-1,k \\ w=\supp(i_1,\dotsc,i_n)}} \frac 1{w 2^{w-1}} \sum_{i_j \neq 0} \sigma_j^2.
\end{align*}}

Consider the following polynomial in $n$ variables $s_1,\dotsc,s_n$:
\begin{equation}\label{eq:sym-poly1}
	f(s_1,\dotsc,s_n)   \triangleq \sum_{\substack{i_1+\dotsb+i_n=k \\ i_1,\dotsc,i_n \neq 1,k-1,k \\ w=\supp(i_1,\dotsc,i_n)}} \frac 1{w 2^{w-1}} \sum_{i_j \neq 0} s_j.
\end{equation}
It has degree $1$ (linear) and is symmetric in its variables. Any degree-$1$ symmetric polynomial in $n$ variables has the form
\begin{equation}\label{eq:sym-poly2}
	f(s_1,\dotsc,s_n) = a \frac{s_1 + \dotsb + s_n}n.
\end{equation}
Setting all $s_j = 1$ and comparing \cref{eq:sym-poly1} with \cref{eq:sym-poly2}, we obtain that
\[
	a = \sum_{\substack{i_1+\dotsb+i_n=k \\ i_1,\dotsc,i_n \neq 1,k-1,k \\ w=\supp(i_1,\dotsc,i_n)}} \frac 1{2^{w-1}}.
\]
Therefore,
\ifthenelse{\boolean{THESIS_VERSION}}{
\begin{align*}
	&\frac 12 k! \sigma^2 \beta_{\mathrm{max}}^{k-2}  + \frac 12 k! \beta_{\mathrm{max}}^{k-2} \sum_{\substack{i_1+\dotsb+i_n=k \\ i_1,\dotsc,i_n \neq 1,k-1,k \\ w=\supp(i_1,\dotsc,i_n)}} \frac 1{w 2^{w-1}} \sum_{i_j \neq 0} \sigma_j^2 %
	=\frac 12 k! \sigma^2 \beta_{\mathrm{max}}^{k-2} 
	+ \frac 12 k! \frac{\sigma^2}n \beta_{\mathrm{max}}^{k-2} \sum_{\substack{i_1+\dotsb+i_n=k \\ i_1,\dotsc,i_n \neq 1,k-1,k \\ w=\supp(i_1,\dotsc,i_n)}} \frac 1{2^{w-1}}.
\end{align*}}{
\begin{align*}
	&\frac 12 k! \sigma^2 \beta_{\mathrm{max}}^{k-2}  + \frac 12 k! \beta_{\mathrm{max}}^{k-2} \sum_{\substack{i_1+\dotsb+i_n=k \\ i_1,\dotsc,i_n \neq 1,k-1,k \\ w=\supp(i_1,\dotsc,i_n)}} \frac 1{w 2^{w-1}} \sum_{i_j \neq 0} \sigma_j^2 \\
	&=\frac 12 k! \sigma^2 \beta_{\mathrm{max}}^{k-2} 
	+ \frac 12 k! \frac{\sigma^2}n \beta_{\mathrm{max}}^{k-2} \sum_{\substack{i_1+\dotsb+i_n=k \\ i_1,\dotsc,i_n \neq 1,k-1,k \\ w=\supp(i_1,\dotsc,i_n)}} \frac 1{2^{w-1}}.
\end{align*}}

For a fixed $2 \le w \le \lfloor \nicefrac k2 \rfloor$, the number of integer solutions of
\begin{gather*}
	i_1+\dotsb+i_n=k, \\ 
	i_1,\dotsc,i_n \neq 1,k-1,k, \\
	w=\supp(i_1,\dotsc,i_n),
\end{gather*}
is $\binom nw$ times the number of integer solutions of\footnote{Everywhere we assume that $\binom xy = 0$ if $x < y$.}
\begin{gather*}
	x_1 + \dotsb + x_w = k, \\
	2 \le x_1, \dotsc, x_w \le k-2.
\end{gather*}
Since $w \ge 2$, the constraint $\le k-2$ is redundant and can be dropped. Next, by substitution $x_i = y_i+1$, the equation is equivalent to %
\begin{gather*}
	y_1 + \dotsb + y_w = k-w, \\
	y_1, \dotsc, y_w \ge 1,
\end{gather*}
which has $\binom{k-w-1}{w-1}$ integer solutions (the number of $w$-compositions).

Altogether, we continue as follows:
\ifthenelse{\boolean{THESIS_VERSION}}{
\begin{align*}
	&\frac 12 k! \sigma^2 \beta_{\mathrm{max}}^{k-2} 
	+ \frac 12 k! \frac{\sigma^2}n \beta_{\mathrm{max}}^{k-2} \sum_{\substack{i_1+\dotsb+i_n=k \\ i_1,\dotsc,i_n \neq 1,k-1,k \\ w=\supp(i_1,\dotsc,i_n)}} \frac 1{2^{w-1}} %
	= \frac 12 k! \sigma^2 (1+A_{nk}) \beta_{\mathrm{max}}^{k-2},
\end{align*}}{
\begin{align*}
	&\frac 12 k! \sigma^2 \beta_{\mathrm{max}}^{k-2} 
	+ \frac 12 k! \frac{\sigma^2}n \beta_{\mathrm{max}}^{k-2} \sum_{\substack{i_1+\dotsb+i_n=k \\ i_1,\dotsc,i_n \neq 1,k-1,k \\ w=\supp(i_1,\dotsc,i_n)}} \frac 1{2^{w-1}} \\
	&\quad= \frac 12 k! \sigma^2 (1+A_{nk}) \beta_{\mathrm{max}}^{k-2},
\end{align*}}
where
\[
	A_{nk} \triangleq \frac 1n \sum_{w=2}^{\lfloor \nicefrac k2 \rfloor} \frac 1{2^{w-1}} \binom nw \binom{k-w-1}{w-1}.
\]
Now, we will require that
\begin{equation*}
	\frac 12 k! \sigma^2 (1+A_{nk}) \beta_{\mathrm{max}}^{k-2} 
	\le 
	\frac 12 k! \sigma^2 \left(\alpha_{nk} \sqrt n \beta_{\mathrm{max}}\right)^{k-2},
\end{equation*}
where $\alpha_{nk} > 0$ is some expression we want to bound. The last inequality is equivalent to
\begin{equation}\label{eq:sqr_ineq_1}
\alpha_{nk}^{k-2} \ge \frac{1+A_{nk}}{n^{\nicefrac k2-1}}.
\end{equation}
Now, we want to find an \emph{upper} bound for all $n \ge 2$ and $k \ge 4$ on
the right-hand side of \cref{eq:sqr_ineq_1}.
We proceed as follows:
\ifthenelse{\boolean{THESIS_VERSION}}{
\begin{align*}
	\frac{1+A_{nk}}{n^{\nicefrac k2-1}} 
	&=\frac{1}{n^{\nicefrac k2 - 1}}
	+ \frac{1}{n^{\nicefrac k2}} \sum_{w=2}^{\lfloor \nicefrac k2 \rfloor} \frac 1{2^{w-1}} \binom nw \binom{k-w-1}{w-1} \\
	&\le\frac{1}{n^{\lfloor\nicefrac k2\rfloor - 1}}
	+ \frac{1}{n^{\lfloor\nicefrac k2\rfloor}} \sum_{w=2}^{\lfloor \nicefrac k2 \rfloor} \frac 1{2^{w-1}} \binom nw \binom{2 \lfloor \nicefrac k2 \rfloor-w}{w-1} \\
	&= \left[ \ell \triangleq \lfloor \nicefrac k2 \rfloor \ge 2 \right] \\
	&= \frac{1}{n^{\ell-1}} + \frac{1}{n^\ell} \sum_{w=2}^{\ell} \frac{1}{2^{w-1}} \binom nw \binom{2\ell-w}{w-1} \\
	&=\left[\text{for $w \le \ell$, we have } \frac{\binom nw}{n^\ell} \le \frac{1}{w!n^{\ell-w}}\right] \\
	&\le \frac{1}{n^{\ell-1}} + \sum_{w=2}^{\ell} \frac{1}{2^{w-1} w! n^{\ell-w}} \binom{2\ell-w}{w-1}.
\end{align*}}{
\begin{align*}
	&\frac{1+A_{nk}}{n^{\nicefrac k2-1}} \\
	&\quad=\frac{1}{n^{\nicefrac k2 - 1}}
	+ \frac{1}{n^{\nicefrac k2}} \sum_{w=2}^{\lfloor \nicefrac k2 \rfloor} \frac 1{2^{w-1}} \binom nw \binom{k-w-1}{w-1} \\
	&\quad\le\frac{1}{n^{\lfloor\nicefrac k2\rfloor - 1}}
	+ \frac{1}{n^{\lfloor\nicefrac k2\rfloor}} \sum_{w=2}^{\lfloor \nicefrac k2 \rfloor} \frac 1{2^{w-1}} \binom nw \binom{2 \lfloor \nicefrac k2 \rfloor-w}{w-1} \\
	&\quad= \left[ \ell \triangleq \lfloor \nicefrac k2 \rfloor \ge 2 \right] \\
	&\quad= \frac{1}{n^{\ell-1}} + \frac{1}{n^\ell} \sum_{w=2}^{\ell} \frac{1}{2^{w-1}} \binom nw \binom{2\ell-w}{w-1} \\
	&\quad=\left[\text{for $w \le \ell$, we have } \frac{\binom nw}{n^\ell} \le \frac{1}{w!n^{\ell-w}}\right] \\
	&\quad\le \frac{1}{n^{\ell-1}} + \sum_{w=2}^{\ell} \frac{1}{2^{w-1} w! n^{\ell-w}} \binom{2\ell-w}{w-1}.
\end{align*}}
The last expression for a fixed $\ell \ge 2$ has the  form
\[
	a_0 + \frac{a_1}{n} + \frac{a_2}{n^2} + \dotsb + \frac{a_{\ell-1}}{n^{\ell-1}},
\]
where $a_0,a_1,\dotsc,a_{\ell-1}$ are positive constants. Thus, for a fixed $\ell \ge 2$, it is decreasing in $n$, and
\ifthenelse{\boolean{THESIS_VERSION}}{
\begin{align*}
	\frac{1}{n^{\ell-1}} + \sum_{w=2}^{\ell} \frac{1}{2^{w-1} w! n^{\ell-w}} \binom{2\ell-w}{w-1} %
	&\le \left.\left( \frac{1}{n^{\ell-1}} + \sum_{w=2}^{\ell} \frac{1}{2^{w-1} w! n^{\ell-w}} \binom{2\ell-w}{w-1} \right)\right|_{n=2} \\
	&= \frac{1}{2^{\ell-1}} + \frac{1}{2^{\ell-1}}\sum_{w=2}^{\ell} \frac{1}{w!} \binom{2\ell-w}{w-1}.
\end{align*}}{
\begin{align*}
	&\frac{1}{n^{\ell-1}} + \sum_{w=2}^{\ell} \frac{1}{2^{w-1} w! n^{\ell-w}} \binom{2\ell-w}{w-1} \\
	&\le \left.\left( \frac{1}{n^{\ell-1}} + \sum_{w=2}^{\ell} \frac{1}{2^{w-1} w! n^{\ell-w}} \binom{2\ell-w}{w-1} \right)\right|_{n=2} \\
	&= \frac{1}{2^{\ell-1}} + \frac{1}{2^{\ell-1}}\sum_{w=2}^{\ell} \frac{1}{w!} \binom{2\ell-w}{w-1}.
\end{align*}}
Since $\frac{1}{w!} \binom{2\ell-w}{w-1} \le 2^{\ell-w}$ (details omitted for brevity), we have
\ifthenelse{\boolean{THESIS_VERSION}}{
\begin{align*}
	&\frac{1}{2^{\ell-1}} + \frac{1}{2^{\ell-1}}\sum_{w=2}^{\ell} \frac{1}{w!} \binom{2\ell-w}{w-1} %
	\le \frac{1}{2^{\ell-1}} + \frac{1}{2^{\ell-1}}\sum_{w=2}^{\ell} 2^{\ell-w} = 1.
\end{align*}}{
\begin{align*}
	&\frac{1}{2^{\ell-1}} + \frac{1}{2^{\ell-1}}\sum_{w=2}^{\ell} \frac{1}{w!} \binom{2\ell-w}{w-1} \\
	&\qquad\le \frac{1}{2^{\ell-1}} + \frac{1}{2^{\ell-1}}\sum_{w=2}^{\ell} 2^{\ell-w} = 1.
\end{align*}}

Therefore, $\alpha_{nk}^{k-2} \le 1$, and thus, $\alpha_{nk} \le 1$. We can take $\alpha_{nk} = 1$ and thus obtain that for any $n \ge 2$ and $k \ge 4$,
\begin{align*}
	\left| \EE{\left(X_1+\dotsb+X_n\right)^k} \right| \le \frac 12 k! (\sqrt n \beta_{\mathrm{max}})^{k-2},
\end{align*}
and $X_1 + \dotsb + X_n$ satisfies the Bernstein's condition with  $\beta$-parameter $\sqrt n \beta_{\mathrm{max}}$.

\subsection{Proof of Lemma~\ref{lem:Tba_BC}} \label{app:proof_Tba_BC}
	Fix two agents $a$ and $b \neq a$. For brevity, we will write in this proof $\kappa$ instead of $\kappa_{b \to a}$.
	
	From Appendix~\ref{app:Tba_variances}, it is clear that under RR scheduling in both PM-I and PM-II, with MoM and non-MoM weights, $\Var{T_{b \to a}} = O \left( \nicefrac 1t\right)$ and thus, there exists $\sigma_T^2$ independent of $t$ such that $\Var{T_{b \to a}} \le \nicefrac{\sigma_T^2}t$ for all $t$.
	
	Now, let us show that $T_{b \to a}$ satisfies Bernstein's condition. As a shorthand, we define $\beta_b^{\max} \triangleq \max(\beta_b, \sigma_b)$ and $\beta_{\mathrm{DP}}^{\max} \triangleq \max(\beta_{\mathrm{DP}}, \sigmaDP)$. From \cref{eq:Tab}, we get
	\ifthenelse{\boolean{THESIS_VERSION}}{
	\begin{align}
		T_{b \to a} &= \sum_{j=1}^\kappa w_j \left( \bar X_b^{(t_j)} + Z_{b \to a}^{(t_j)} \right) %
		= \sum_{j=1}^\kappa w_j \left( \frac{1}{t_j} \sum_{i=1}^{t_j} X_b^{(i)} + Z_{b \to a}^{(t_j)} \right) \notag \\
		&= \sum_{j=1}^\kappa \frac{w_j}{t_j} \sum_{i=1}^{t_j} X_b^{(i)} + \sum_{j=1}^\kappa w_j Z_{b \to a}^{(t_j)} %
		= \sum_{i=1}^\kappa \left( \sum_{j=i}^\kappa \frac{w_j}{t_j} \right) \sum_{\tau=t_{i-1}+1}^{t_i} X_b^{(\tau)} + \sum_{j=1}^\kappa w_j Z_{b \to a}^{(t_j)}. \label{eq:Tba_expanded}
	\end{align}}{
	\begin{align}
		T_{b \to a} &= \sum_{j=1}^\kappa w_j \left( \bar X_b^{(t_j)} + Z_{b \to a}^{(t_j)} \right) \notag \\
		&= \sum_{j=1}^\kappa w_j \left( \frac{1}{t_j} \sum_{i=1}^{t_j} X_b^{(i)} + Z_{b \to a}^{(t_j)} \right) \notag \\
		&= \sum_{j=1}^\kappa \frac{w_j}{t_j} \sum_{i=1}^{t_j} X_b^{(i)} + \sum_{j=1}^\kappa w_j Z_{b \to a}^{(t_j)} \notag \\
		&= \sum_{i=1}^\kappa \left( \sum_{j=i}^\kappa \frac{w_j}{t_j} \right) \sum_{\tau=t_{i-1}+1}^{t_i} X_b^{(\tau)} + \sum_{j=1}^\kappa w_j Z_{b \to a}^{(t_j)}. \label{eq:Tba_expanded}
	\end{align}}

	\subsubsection{The First Term of \eqref{eq:Tba_expanded}}

	Note that all $\nicefrac{w_j}{t_j} \ge 0$ and, thus, 
	\[
		\max_{i \in [\kappa]} \left\{ \sum_{j=i}^\kappa \frac{w_j}{t_j} \right\} =  \sum_{j=1}^\kappa \frac{w_j}{t_j}.
	\]
	Since all $X_b^{(i)} \sim \BC{\mu_b}{\sigma_b^2}{\beta_b}$ and are mutually independent, we have (using Proposition~\ref{prop:iid_bernstein}) for the first term in \eqref{eq:Tba_expanded}:
	\ifthenelse{\boolean{THESIS_VERSION}}{
	\begin{align*}
		&\sum_{i=1}^\kappa \left( \sum_{j=i}^\kappa \frac{w_j}{t_j} \right) \sum_{\tau=t_{i-1}+1}^{t_i} X_b^{(\tau)} %
		\sim \BC{\mu_b}{\sigma_b^2 \sum_{i=1}^\kappa (t_i-t_{i-1})\left( \sum_{j=1}^\kappa \frac{w_j}{t_j} \right)^2}{ \beta_b^{\max} t_{\kappa}^{1/2} \sum_{j=1}^\kappa \frac{w_j}{t_j}}.
	\end{align*}}{
	\begin{align*}
		&\sum_{i=1}^\kappa \left( \sum_{j=i}^\kappa \frac{w_j}{t_j} \right) \sum_{\tau=t_{i-1}+1}^{t_i} X_b^{(\tau)} \\
		& \hspace{-0.12cm} \sim \BC{\mu_b}{\sigma_b^2 \sum_{i=1}^\kappa (t_i-t_{i-1})\left( \sum_{j=1}^\kappa \frac{w_j}{t_j} \right)^2}{ \beta_b^{\max} t_{\kappa}^{1/2} \sum_{j=1}^\kappa \frac{w_j}{t_j}}.
	\end{align*}}
	For MoM (and both privacy schemes), we have 
	\ifthenelse{\boolean{THESIS_VERSION}}{
	\begin{gather}\label{eq:term-1}
	\begin{aligned}
		\beta_b^{\max} t_{\kappa}^{1/2} \sum_{j=1}^\kappa \frac{w_j}{t_j} &= \frac{\beta_b^{\max} t_{\kappa}^{1/2}}{\kappa} \sum_{j=1}^\kappa \frac{1}{t_j}
		\le \frac{\beta_b^{\max} t_{\kappa}^{1/2}}{\kappa} \sum_{j=1}^\kappa \frac{1}{j} %
		= \beta_b^{\max} t_{\kappa}^{1/2} \frac{O(\log \kappa)}{\kappa} = O\left( \frac{\log t}{\sqrt{t}} \right),
	\end{aligned}
	\end{gather}}{
	\begin{gather}\label{eq:term-1}
	\begin{aligned}
		\beta_b^{\max} t_{\kappa}^{1/2} \sum_{j=1}^\kappa \frac{w_j}{t_j} &= \frac{\beta_b^{\max} t_{\kappa}^{1/2}}{\kappa} \sum_{j=1}^\kappa \frac{1}{t_j}
		\le \frac{\beta_b^{\max} t_{\kappa}^{1/2}}{\kappa} \sum_{j=1}^\kappa \frac{1}{j} \\
		&= \beta_b^{\max} t_{\kappa}^{1/2} \frac{O(\log \kappa)}{\kappa} = O\left( \frac{\log t}{\sqrt{t}} \right),
	\end{aligned}
	\end{gather}}
	and for non-MoM (and both privacy schemes):
	\begin{align}
		\beta_b^{\max} t_{\kappa}^{1/2} \sum_{j=1}^\kappa \frac{w_j}{t_j} = \frac{\beta_b^{\max} t_{\kappa}^{1/2}}{t_\kappa} = O \left( \frac 1{\sqrt{t}} \right) = O \left( \frac{\log t}{\sqrt{t}} \right). \label{eq:term-2}
	\end{align}

	\subsubsection{The Second Term of \eqref{eq:Tba_expanded}}
	Now consider the second term in \eqref{eq:Tba_expanded}. For PM-I, we have
	\begin{gather*}
		\sum_{j=1}^\kappa w_j Z_{b \to a}^{(t_j)} = \sum_{i=1}^\kappa \left( \sum_{j=i}^\kappa \frac{w_j}{t_j} \right) Z_{b \to a}^{(t_{i-1}+1:t_i)},
	\end{gather*}
	which (similarly to \eqref{eq:term-1} and \eqref{eq:term-2}) satisfies Bernstein's condition with parameter $O \left( \frac{\log t}{\sqrt{t}} \right)$.
	
	For PM-II and non-MoM, we have 
	\[
		\sum_{j=1}^\kappa w_j Z_{b \to a}^{(t_j)} = Z_{b \to a}^{(t_\kappa)},
	\]
	and $Z_{b \to a}^{(t_\kappa)}$ is the scaled by $\nicefrac{1}{t_\kappa}$ sum of $w_{\mathrm H}(\kappa) \le t_\kappa$ independent noises, each satisfying Bernstein's condition with the parameter $\beta_{\mathrm{DP}}$. Therefore, it satisfies Bernstein's condition with  $\beta$-parameter
	\begin{align}
		\frac{\beta_{\mathrm{DP}}^{\max} t_\kappa^{1/2}}{t_k} = O\left( \frac 1{\sqrt{t}} \right) = O\left( \frac{\log t}{\sqrt{t}}\right).
		\label{eq:term-3}
	\end{align}

	Now, consider the case of PM-II and MoM. The total noise term is constructed from individual noises added to the subsums as described in Section~\ref{sec:PM-II}, and there are exactly $\kappa$ such noises altogether. Each of those individual noises has variance $\sigmaDP^2$ and satisfies Bernstein's condition with the parameter $\beta_{\mathrm{DP}}$. Due to the MoM construction (average of all the noisy means received from the agent $b$ so far), many such subsums are used multiple times in the final noise term of the statistic $T_{b \to a}$.
	
	There are two types of the subsums (for our purposes here). First, a subsum of the form $[1:t_{2^s}]$ for any $s \ge 0$ has the corresponding noise term $Z_{b \to a}^{(1:t_{2^s})}$, which will be included only in $Z_{b \to a}^{(t_\kappa)}$ for $\kappa=2^s, 2^s+1, \dotsc, 2^{s+1}-1$ (for larger values of $\kappa$, it will be covered by larger subsums of the form $[1:t_{2^{s'}}]$ for $s' > s$). Also, the subsum $[1:t_{2^s}]$ (together with $Z_{b \to a}^{(1:t_{2^s})}$) will be included in $T_{b \to a}$ for $t \ge t_{2^s}$, where the coefficient in front of $Z_{b \to a}^{(1:t_{2^s})}$ will be at most
	\ifthenelse{\boolean{THESIS_VERSION}}{
	\begin{align*}
		\frac{1}{\kappa}\left( \frac{1}{t_{2^s}} + \frac{1}{t_{2^s+1}} + \dotsb + \frac{1}{t_{2^{s+1}-1}} \right) %
		&\le \frac{1}{\kappa}\left( \frac{1}{2^s} + \frac{1}{2^s+1} + \dotsb + \frac{1}{2^{s+1}-1} \right) \\
		&\le \frac{1}{\kappa}\left( \frac{1}{2^s} + \frac{1}{2^s} + \dotsb + \frac{1}{2^s} \right) = \frac{1}{\kappa}.
	\end{align*}}{
	\begin{align*}
		&\frac{1}{\kappa}\left( \frac{1}{t_{2^s}} + \frac{1}{t_{2^s+1}} + \dotsb + \frac{1}{t_{2^{s+1}-1}} \right) \\
		&\quad\le \frac{1}{\kappa}\left( \frac{1}{2^s} + \frac{1}{2^s+1} + \dotsb + \frac{1}{2^{s+1}-1} \right) \\
		&\quad\le \frac{1}{\kappa}\left( \frac{1}{2^s} + \frac{1}{2^s} + \dotsb + \frac{1}{2^s} \right) = \frac{1}{\kappa}.
	\end{align*}}

	Second, all the subsums $[t_i:t_j]$ for $2^s +1 \le t_i \le 2^{s+1}-1$ will have $i \le j \le 2^{s+1}-1$. Moreover, these subsums will be present in $Z_{b \to a}^{(t)}$ \emph{only if } $[1:t_{2^s}]$ is also present. Consequently, the coefficients in $T_{b \to a}$ in front of the noises corresponding to these subsums will be less than $\nicefrac{1}{\kappa}$.
	
	Altogether, for PM-II and MoM, the noise term in $T_{b \to a}$ satisfies Bernstein's condition with $\beta$-parameter not more than
	\begin{equation}
	\frac{\beta_{\mathrm{DP}}^{\max} \kappa^{1/2}}{\kappa} = O\left( \frac 1{\sqrt{t}} \right) = O \left( \frac{\log t}{\sqrt{t}} \right).
	\label{eq:term-4}
	\end{equation}

	Combining \eqref{eq:Tba_expanded} with \cref{eq:term-1,eq:term-2,eq:term-3,eq:term-4}, we have that $T_{b \to a}$ satisfies Bernstein's condition with parameter $O \left( \frac{\log t}{\sqrt{t}} \right)$. Therefore, there exists a constant $\beta_T > 0$ independent of $t$, so that 
	\[
	T_{b \to a} \sim \BC{\mu_b}{\Var{T_{b \to a}}}{\frac{\beta_T \ln t}{\sqrt{t}}}.
	\]

\balance

	\subsection{Proof of Lemma~\ref{lem:prob-type-II}} \label{app:proof_lem:prob-type-II}
	\ifthenelse{\boolean{THESIS_VERSION}}{
\begin{align*}
	\Prv{\chi_a^{(t)}(b; \theta_t) = 1} 
	&= \Prv{\left| \bar{X}_a^{(t)} - T_{b \to a}  \right| \le  \Phi^{-1}\left( 1 - \frac{\theta_t}{2}\right) \sqrt{\frac{\sigma_a^2}{t} + \Var{T_{b \to a}}}} \\
	&\le \Prv{ \bar{X}_a^{(t)} - T_{b \to a} \le  \Phi^{-1}\left( 1 - \frac{\theta_t}{2}\right) \sqrt{\frac{\sigma_a^2}{t} + \Var{T_{b \to a}}}} \\
	&= \Prv{\bar{X}_a^{(t)} - T_{b \to a} \le  \Phi^{-1}\left( 1 - \frac{\theta_t}{2}\right) \sqrt{\frac{s_t^2}{t}}},
\end{align*}}{
\begin{align*}
	&\Prv{\chi_a^{(t)}(b; \theta_t) = 1} \\
	&= \Prv{\left| \bar{X}_a^{(t)} - T_{b \to a}  \right| \le  \Phi^{-1}\left( 1 - \frac{\theta_t}{2}\right) \sqrt{\frac{\sigma_a^2}{t} + \Var{T_{b \to a}}}} \\
	&\le \Prv{ \bar{X}_a^{(t)} - T_{b \to a} \le  \Phi^{-1}\left( 1 - \frac{\theta_t}{2}\right) \sqrt{\frac{\sigma_a^2}{t} + \Var{T_{b \to a}}}} \\
	&= \Prv{\bar{X}_a^{(t)} - T_{b \to a} \le  \Phi^{-1}\left( 1 - \frac{\theta_t}{2}\right) \sqrt{\frac{s_t^2}{t}}},
\end{align*}}
where $s_t^2 \triangleq t \Var{\bar{X}_a^{(t)} - T_{b \to a}} = \sigma_a^2 + t \Var{T_{b \to a}} \le \sigma_a^2 + \sigma_T^2$. Note that $s_t = \sqrt{\sigma_a^2 + t \Var{T_{b \to a}}} \ge \sigma_a$.

To simplify notation, we define 
\begin{align*}
	z_t &\triangleq \Phi^{-1}\left(1 - \frac{\theta_t}2\right) = o(\sqrt t), \\
	\Delta &\triangleq \mu_a - \mu_b > 0 \quad \text{(w.l.o.g)}, \\
	\beta_U &\triangleq \sqrt 2 \max\left\{ \beta_a, \beta_T, \sigma_a, \sigma_T  \right\}, \\
	U_t &\triangleq \bar X_a^{(t)} - T_{b \to a} \sim \BC{\Delta}{\frac{s_t^2}t}{\frac{\beta_U \ln t}{\sqrt{t}}}, \\
	c_t &\triangleq \frac{\Delta}{s_t} > 0,
\end{align*}
where $U_t$ satisfies Bernstein's condition with $\beta$-parameter $\nicefrac{\beta_U \ln t}{\sqrt{t}}$ by Lemma~\ref{lem:Tba_BC} and Proposition~\ref{prop:iid_bernstein}.

We continue as follows:
\ifthenelse{\boolean{THESIS_VERSION}}{
\begin{align*}
	\Prv{\chi_a^{(t)}(b; \theta_t) = 1} 
	&\le \Prv{U_t \le z_t \frac{s_t}{\sqrt t}} %
	= \Prv{\frac{U_t - \Delta}{s_t/\sqrt t} \le z_t - \frac{\Delta}{s_t}\sqrt t} \\
	&\le \Prv{\frac{U_t - \Delta}{s_t/\sqrt t} \le z_t - c_t\sqrt t} \\
	&\stackrel{(a)}{\le} \Prv{\left| \frac{U_t - \Delta}{s_t/\sqrt t} \right| \ge c_t\sqrt t - z_t} \\
	&\stackrel{(b)}{\le} 2 {\mathrm{e}}^{-\frac{(c_t \sqrt t - z_t)^2}{2 + \nicefrac{2 \beta_U}{\sigma_a}\ln t (c_t \sqrt t - z_t)}} = 2{\mathrm{e}}^{-O \left( \frac{\sqrt t}{\log t} \right)} %
	= o \left( \frac{1}{t^n} \right),
\end{align*}}{
\begin{align*}
	&\Prv{\chi_a^{(t)}(b; \theta_t) = 1} 
	\le \Prv{U_t \le z_t \frac{s_t}{\sqrt t}} \\
	&= \Prv{\frac{U_t - \Delta}{s_t/\sqrt t} \le z_t - \frac{\Delta}{s_t}\sqrt t} \\
	&\le \Prv{\frac{U_t - \Delta}{s_t/\sqrt t} \le z_t - c_t\sqrt t} \\
	&\stackrel{(a)}{\le} \Prv{\left| \frac{U_t - \Delta}{s_t/\sqrt t} \right| \ge c_t\sqrt t - z_t} \\
	&\stackrel{(b)}{\le} 2 {\mathrm{e}}^{-\frac{(c_t \sqrt t - z_t)^2}{2 + \nicefrac{2 \beta_U}{\sigma_a}\ln t (c_t \sqrt t - z_t)}} = 2{\mathrm{e}}^{-O \left( \frac{\sqrt t}{\log t} \right)} \\
	&= o \left( \frac{1}{t^n} \right),
\end{align*}}
for any positive integer $n$. Here, $(a)$ is because for all large enough $t$, $z_t - c_t \sqrt t = o(\sqrt t) - c_t \sqrt t < 0$. And $(b)$ follows from Proposition~\ref{prop:BC-tails}, since $$\frac{U_t - \Delta}{s_t/\sqrt t} \sim \BC{0}{1}{\frac{\beta_U \ln t}{s_t }}$$ and thus, by monotonicity of the Bernstein parameter, $$\frac{U_t - \Delta}{s_t/\sqrt t} \sim \BC{0}{1}{\frac{\beta_U}{\sigma_a}\ln t}.$$
\ifCLASSOPTIONcaptionsoff
  \newpage
\fi

\putbib[\bibfiles/defshort1,\bibfiles/biblioHY,\bibfiles/refs]

\end{bibunit}

\end{document}